\documentclass{article}

\usepackage[main, final]{neurips_2026}

\usepackage[utf8]{inputenc} 
\usepackage[T1]{fontenc}    
\usepackage{hyperref}       
\usepackage{url}            
\usepackage{booktabs}       
\usepackage{amsfonts}       
\usepackage{nicefrac}       
\usepackage{microtype}      
\usepackage{xcolor}         

\usepackage{multirow}
\usepackage{makecell}
\usepackage{graphicx}
\usepackage[table]{xcolor} 
\usepackage{enumitem}
\definecolor{ForestGreen}{RGB}{34,139,34}
\definecolor{BrickRed}{RGB}{203,65,84}
\setlist[enumerate]{leftmargin=*, nosep}
\usepackage{subcaption} 
\usepackage{amsthm}
\usepackage{algorithm}
\usepackage{algpseudocode}
\usepackage{amsmath, amssymb}
\usepackage{tcolorbox}
\usepackage{tikz} 

\newtcolorbox{promptbox}[1][]{
  colback=gray!5,
  colframe=gray!60,
  coltitle=black,
  fonttitle=\bfseries,
  title={Prompt},
  boxrule=0.8pt,
  left=5pt, right=5pt, top=5pt, bottom=5pt,
  fontupper=\small\ttfamily,
  before upper={\setlength{\parindent}{0pt}\setlength{\parskip}{0.8em}},
  width=\textwidth, 
  #1
}

\newcommand{\EBC}{\mathrm{EBC}}
\newcommand{\Prij}{\widehat{p}^{\,G}_{ij}}

\newtheorem{proposition}{Proposition}

\theoremstyle{remark}
\newtheorem*{remark}{Remark}
\newtheorem*{theorem*}{Theorem}
\newtheorem*{proposition*}{Proposition}

\title{From Perception to Punchline: Empowering VLM with the Art of In-the-wild Memes}

\author{Xueyan Li\thanks{Equal Contribution.} \\
Shanghai Artificial Intelligence Laboratory \\
\texttt{karroyan13@gmail.com} \\
\And
Yingyi Xue$^*$ \\
Fudan University \\
Shanghai Artificial Intelligence Laboratory \\
\texttt{yingyixue7@gmail.com} \\
\And
Mengjie Jiang \\
Columbia Engineering\\
Columbia University \\
\texttt{mj3290@columbia.edu}
\And
Qingzi Zhu \\
Shanghai Artificial Intelligence Laboratory \\
\texttt{zhuqingzi@pjlab.org.cn}
\And
Yazhe Niu\thanks{Corresponding author.}\\
The Chinese University of Hong Kong MMLab \\
\texttt{niuyazhe314@outlook.com}
}

\begin{document}

\maketitle
\vspace{-16pt}
\begin{abstract}
Generating humorous memes is a challenging multimodal task that moves beyond direct image-to-caption supervision.
It requires a nuanced reasoning over visual content, contextual cues, and subjective humor.
To bridge this gap between visual perception and humorous punchline creation, we propose \textit{HUMOR}, a novel framework that guides VLMs through hierarchical reasoning and aligns them with group-wise human preferences.
First, HUMOR employs a hierarchical, multi-path Chain-of-Thought (CoT): the model begins by identifying a template-level intent, then explores diverse reasoning paths under different contexts, and finally anchors onto a high-quality, context-specific path.
This CoT supervision, which traces back from ground-truth captions, enhances reasoning diversity.
We further analyze that this multi-path exploration with anchoring maintains a high expected humor quality, under the practical condition that high-quality paths retain significant probability mass.
Second, to capture subjective humor, we train a pairwise reward model that operates within groups of memes sharing the same template.
Following established theory, this approach ensures a consistent and robust proxy for human preference, even with subjective and noisy labels.
\textcolor{black}{The reward model then enables group-wise RL with KL-controlled updates, yielding a bounded stability guarantee.}
Extensive experiments show that HUMOR empowers various VLMs with superior reasoning diversity, more reliable preference alignment, and higher overall meme quality.
\textcolor{black}{Beyond memes, our work suggests a framework for subjective multimodal generation where coherent groups make relative judgments more reliable than global scores.}
\end{abstract}

\section{Introduction}
\label{sec:intro}
Creativity in multimodal understanding and generation is progressively shifting from literal description and visual perception to subjective and context-dependent outcomes, such as humor~\citep{hessel2022androids, hwang2023memecap}, aesthetics~\citep{rombach2022high}, and social alignment~\citep{bai2022constitutional}.
In these domains, quality is not defined by a single ground-truth but is instead guided by diverse—and often noisy—distributions of human preference~\citep{yadav2025beyond, burn2018multimodality, song2025large}.
While recent vision–language models (VLMs) have scaled to achieve strong results on captioning and visual question answering~\citep{kuang2025natural, ghandi2023deep}, which admit relatively objective targets~\citep{yan2023achieving}, the question of how to align models with such open-ended values~\citep{bhatia2024local, feng2024modular} remains an open challenge.

As a salient instance and testbed of this open-ended challenge, meme punchline generation illuminates these limitations.
Current approaches often treat it as a straightforward image-to-caption task, optimized with fixed supervised losses on collected data~\citep{peirson2018dank, hwang2023memecap}.
This formulation collapses the inherently diverse reasoning required for humor into the language model, suppresses intermediate interpretability, and tends to yield outputs that are fluent yet shallow or unfunny~\citep{yadav2025beyond}.
Direct generation from a template forfeits control over the interpretive process and makes targeted steering difficult.
Here a template denotes the shared visual basis or theme image for a group of memes.
While recent evidence shows chain-of-thought (CoT) reasoning improves performance in VLMs~\citep{zhang2023multimodal, hu2024visual, lad2025}, we argue that meme generation necessitates a \textbf{hierarchical, multi-path reasoning process.}
Observations from real-world meme data suggest that multiple reasoning paths can lead to distinct metaphorical bindings or punchlines~\citep{xu2022met}.
Concretely, we decompose the process into two stages: a template-level stage that infers a canonical intent, followed by a context-level stage that grounds this intent in specific visual details.
By first exploring multiple reasoning paths and then explicitly anchoring one with supervision, \textcolor{black}{this design encourages diversity while maintaining sufficient humor quality.}
As our theoretical analysis (Section~\ref{sec:method-rl}) shows, this approach preserves a conditional lower bound on expected humor, provided that high-quality paths retain sufficient probability mass and remaining paths are not significantly worse.


\begin{figure*}[t]
	\centerline{\includegraphics[width=0.85\linewidth]{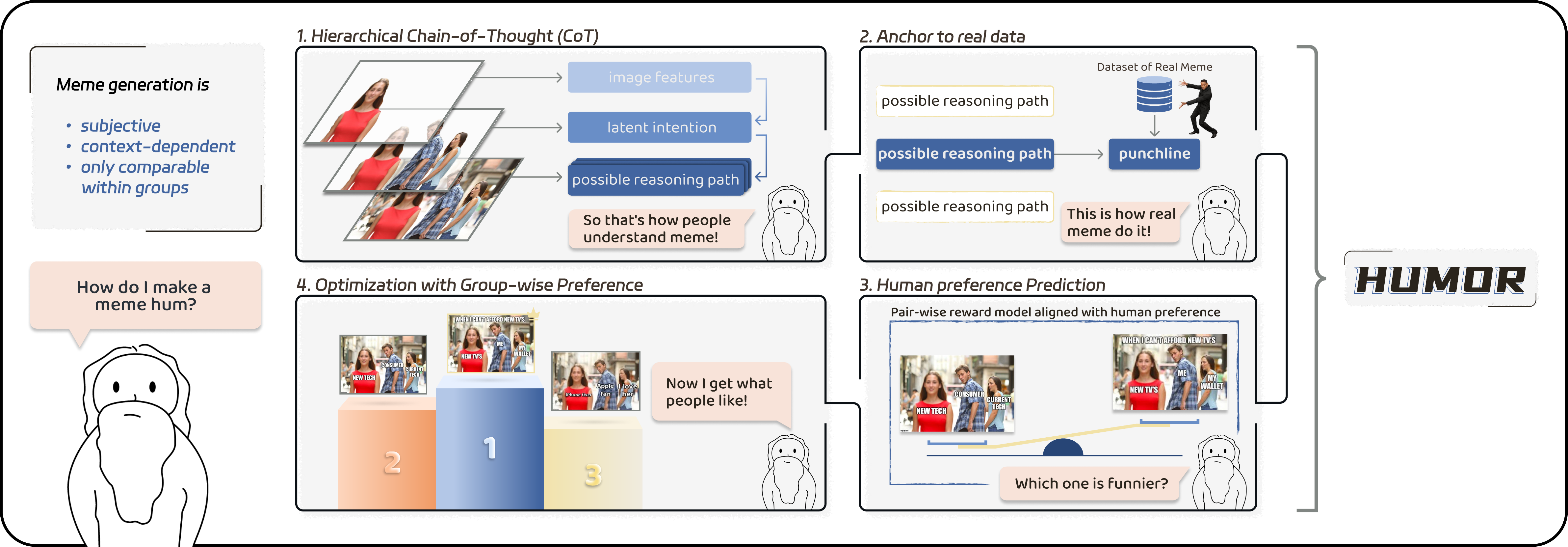}}
\caption{Given a template image, HUMOR first performs hierarchical reasoning with multi-path CoT: a template-level stage infers latent intent, while a context-level stage explores multiple visually grounded reasoning paths. One high-quality path is anchored by tracing back from ground-truth captions, promoting diversity while preserving a conditional humor lower bound. A reward model then compares memes only within groups sharing the same template, preserving rank consistency and providing a proxy for human preference. \textcolor{black}{This reward further enables stable group-wise RL with KL-controlled updates, so potential degradation is explicitly bounded.} Together, these components combine structured reasoning, group-wise preference modeling, and stable optimization for meme generation.
}
\vspace{-18pt}
\label{fig:overview}
\end{figure*}

Furthermore, optimizing generation toward human-preferred humor is essential to meet these theoretical conditions. 
Prior work on computational humor often relies on text-only cues or global funniness scores ~\citep{baluja2024text, kalloniatis2024computational, zhu2025commonality}, implicitly assuming humor is directly comparable across different meme templates. 
In practice, however, human judgments are more reliable and consistent within a coherent template group than across groups with differing conventions.
Ignoring this group structure introduces noise, harms generalization, and encourages models to exploit reward superficial overlaps rather than genuine humorous fit(see empirical analyses in Sec.~\ref{sec:vlm_reliability} and Sec.~\ref{sec:rm_analysis}).
Although humor cannot be reduced to a deterministic rule, we can design a \textbf{pairwise reward model} that enforces rank consistency within each template group.
This model provides a stable, human-aligned preference signal, and further enables group-wise RL to ensure that expected humor cannot degrade beyond a bounded amount.


Motivated by these above analyses, we propose \textit{HUMOR}: a Hierarchical Understanding and Meme Optimization framework via reinforcement learning (RL).
As illustrated in Figure~\ref{fig:overview} (high-level overview), our framework addresses the core challenges in meme punchline through several key components: hierarchical multi-path CoT reasoning, group-wise preference modeling, and stable policy optimization via RL.
\textit{HUMOR} decouples metaphorical reasoning from concrete punchline generation, respects within-group comparability, and translates preference signals into stable policy updates. In summary, our contributions are as follows:


\begin{enumerate} 
    \item \textbf{Formal modeling of meme punchline generation within semantically coherent groups.} We formulate the task as an open-ended, group-wise reasoning problem, supported by hierarchical multi-path CoT supervision and preference reward optimization.
    This approach decouples template-level intent from context-level grounding, exposing explicit reasoning traces while laying the foundation for structured preference optimization.
    \item \textbf{\textcolor{black}{Theoretical guarantees for humor measurement and stable optimization.}} \textcolor{black}{We rigorously show that multi-path CoT supervision preserves a conditional humor lower bound, and that our preference learning ensures consistent within-group ordering with KL-controlled stability.}
    \item \textbf{\textcolor{black}{Comprehensive experiments}} \textcolor{black}{across multiple models and data demonstrate that \textit{HUMOR} improves reasoning diversity, preference alignment, and overall meme quality, with additional analyses on unseen templates and a small cross-domain pilot supporting its broader applicability. To support reproducibility, we have released our anonymized dataset}\footnote{\url{https://huggingface.co/datasets/AnonySubmission/HUMOR-Datasets}}.
\end{enumerate}

\vspace{-6pt}
\section{Problem Formulation}
\vspace{-4pt}
\label{sec:formulation}

This section formulates the core assumptions and components used throughout the paper.
\textcolor{black}{We define the structured meme space and group-wise comparability, then characterize local pairwise preference data and assume a latent humor functional within each group. We further introduce an observation model linking latent humor to pairwise comparisons, and finally formulate the objective of a meme generator together with the key evaluation quantities. This results in a self-contained problem formulation that emphasizes group-wise comparability while remaining agnostic to training methods.}

\paragraph{Meme Space and Group-wise Comparability:}
\label{subsec:meme-space}
Let \(\mathcal{M}\) denote the set of all memes under consideration.
Each meme is represented as a multimodal pair $m \;=\; (I, c)$, where \(I\in \mathcal{I}\) is a base image and \(c\) is a textual punchline rendered at designated positions. 
Many memes are created from widely shared \emph{templates} and are interpreted through context-dependent associations.
Since humor is highly subjective and context-sensitive, absolute comparisons of humor across different templates are often ill-posed.
Therefore, we assume and partition the meme space into $K$ disjoint groups:
\[
\begin{aligned}
\mathcal{G} = \{G_1, \dots, G_K\},
G_k \subset \mathcal{M}, \quad G_k \cap G_\ell = \emptyset \ (k \neq \ell)
\end{aligned}
\]
Memes within the same group share a comparable structure (e.g., the same template, punchline schema, \textcolor{black}{or more generally a shared semantic context and communicative intent}), which enables meaningful humor comparison.
We posit that human judgments of humor are reliable \emph{within} each group \(G_k\in\mathcal{G}\), but do not assume comparability \emph{across} different groups.

\textcolor{black}{More generally, this formulation is intended for subjective generation settings that satisfy three applicability conditions: (i) output quality is highly subjective and absolute cross-context scoring is unreliable; (ii) group boundaries are natural and semantically coherent rather than artificially constructed; and (iii) within-group pairwise preferences are available or can be collected at low cost. In such cases, local relative preference provides a better learning signal than a single globally calibrated score.}

\paragraph{Local Preference Data:}
\label{subsec:local-preference}
For a given group \(G\), 
we collect human annotations indicating which of two memes is considered funnier.
Formally, for \(m_i,m_j\in G\), define $y_{ij}^G \;=\; \mathbb{I}[\, m_i \succ m_j \,] \in \{0,1\}$ where \(m_i \succ m_j\) denotes a local preference that \(m_i\) is judged to be funnier than \(m_j\).
The dataset consists of triples \((G,(m_i,m_j),y_{ij}^G)\) sampled from a pairing distribution over \(G\).
We allow for incompleteness (not all pairs are labeled) and noise (due to inter-annotator disagreement).
We adopt two mild yet standard assumptions from preference learning\textcolor{black}{~\cite{christiano2023deepreinforcementlearninghuman}}:  
(i) \textit{local comparability}: preferences are elicited and interpreted only within a fixed group \(G\);  
(ii) \textit{weak transitivity}: in expectation, if \(m_i \succ m_j\) and \(m_j \succ m_\ell\), then \(m_i \succ m_\ell\) is more likely than its reversal, without requiring a strict total order.

\paragraph{Latent Humor within A Group:}
\label{subsec:latent-humor}
Within each group \(G\), we posit the existence of a latent humor functional $h_G: G \to [0,1],$
This functional maps each meme \(m\in G\) to a scalar reflecting its relative likelihood of being judged as funny by humans in that group.
We do not assume that \(h_G\) is calibrated across different groups, nor that \(h_{G}\) and \(h_{G'}\) are directly comparable when \(G\neq G'\).

\paragraph{Observation Model for Pairwise Labels:}
\label{subsec:observation-model}
Pairwise comparison labels are modeled as noisy observations of underlying differences in latent humor.
Formally, we assume:
\begin{equation}
\label{eq:generic-link}
\Pr\!\big[\, m_i \succ m_j \mid G \,\big]
\;=\;
\Lambda\!\big( \, h_G(m_i) - h_G(m_j) \, \big),
\end{equation}
where \(\Lambda:\mathbb{R}\!\to\!(0,1)\) is a strictly increasing link function\textcolor{black}{~\citep{sun2025rethinking}}.
Eq.~\ref{eq:generic-link} captures the intuition that the probability of preferring \(m_i\) to \(m_j\) depends \emph{only} on their latent humor gap within the same group: when \(h_G(m_i)\!\approx\! h_G(m_j)\), the choice is nearly ambiguous (probability \(\approx\!1/2\));
as the gap increases, the probability moves smoothly toward \(1\) (if \(h_G(m_i)\!>\!h_G(m_j)\)) or \(0\) (otherwise), capturing that larger humor gaps lead to more consistent comparisons. 

\paragraph{Generation Goal and Evaluation Quantities:}
\label{subsec:goal-eval}
\textcolor{black}{A meme generation model is defined as a conditional probability distribution over punchlines (or called captions) given an image: $\pi_\theta(\cdot \mid I):
\quad I \mapsto\; \Delta(\mathcal{C})$, where $\Delta(\mathcal{C})$ denotes the set of probability distributions over the caption space $\mathcal{C}$.}
A meme sample \(m=(I,c)\) is instantiated by sampling a caption \(c \sim \pi_\theta(\cdot\mid I)\).
For any target group \(G\) containing meme candidates derived from the base image $I$, the expected within-group humor of \(\pi_\theta\) is defined as $
\label{eq:expected-humor}
\mathcal{H}_G(\theta)
\;=\;
\mathbb{E}_{\, c \sim \pi_\theta(\cdot\mid I)}\big[\, h_G\big((I,c)\big) \,\big]
$.
The overall population objective is then obtained by aggregating over groups according to a task-specific distribution over \((I,G)\):
\begin{equation}
\label{eq:population-objective}
\mathcal{H}(\theta)
\;=\;
\mathbb{E}_{\, (I,G)}\big[\, \mathcal{H}_G(\theta) \,\big].
\end{equation}

\vspace{-8pt}
\section{HUMOR Framework}
\label{sec:framework}
\vspace{-4pt}
\label{sec:method}

\begin{figure*}[!htbp]
    \centering
    \includegraphics[width=0.85\linewidth]{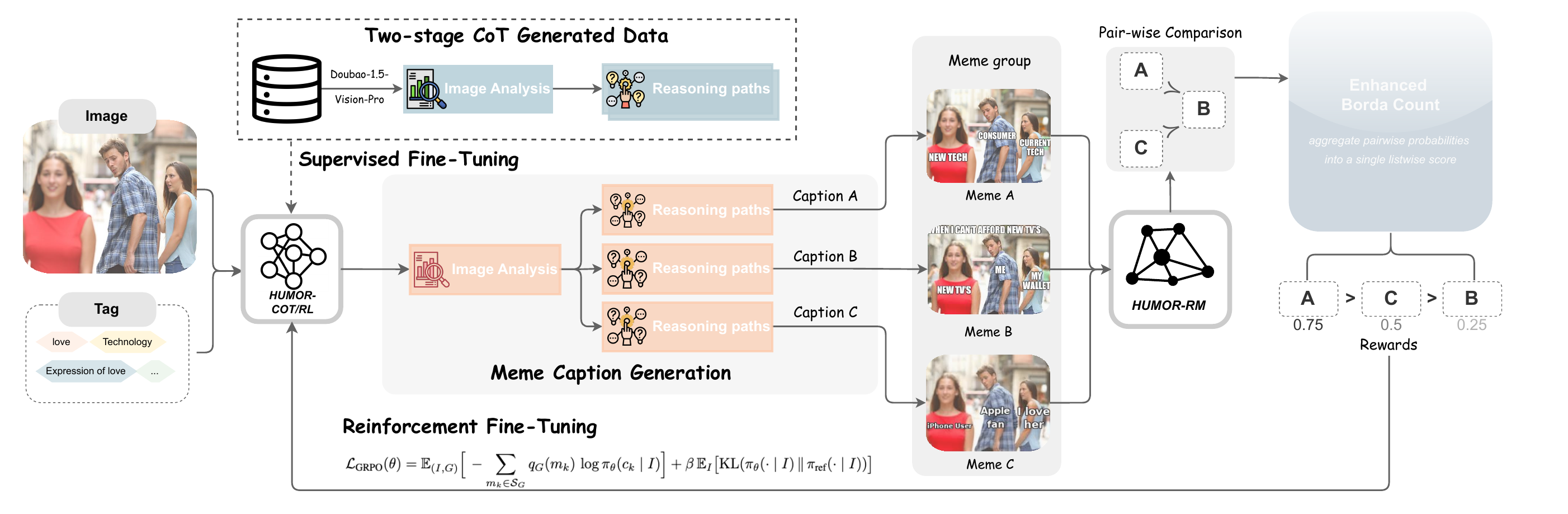}
    \caption{Training Pipeline of HUMOR. Multi-path CoT expands reasoning coverage and anchors a canonical path; the reward model translates pair data into a rank-consistent group-level signal (via EBC); GRPO then updates the generator toward higher-ranked captions.}
    \label{fig:pipeline}
    \vspace{-12pt}
\end{figure*}

We propose \textbf{HUMOR}: \textbf{H}ierarchical \textbf{U}nderstanding and \textbf{M}eme \textbf{O}ptimization, a framework that guides VLMs through hierarchical reasoning and aligns them with group-wise humor preferences. \textcolor{black}{The overview of the framework is shown in Fig.~\ref{fig:pipeline}.
HUMOR consists of three components: hierarchical CoT supervision, pairwise reward modeling, and group-wise policy optimization.
These components collectively \textcolor{black}{support diverse reasoning, consistent preference learning, and stable preference-aligned optimization.}
Propositions~\ref{prop:cot-lb}, \ref{prop:rank-consistency}, \ref{prop:noise-robustness-appendix}, and~\ref{prop:grpo-bound} formally establish the coherence and controllability of our framework.}

\vspace{-6pt}
\subsection{Hierarchical Chain-of-Thought Supervision}
\label{sec:cot}
\vspace{-4pt}
\textcolor{black}{Meme creation mirrors a hierarchical cognitive routine: humans first parse what a visual template affords, and then realize a chosen intent with text that fits the surrounding context~\cite{flamson2008encryption}. We therefore model meme generation as a two-stage reasoning process, separating (i) intent inference from the image and (ii) context-sensitive textual realization of that intent. In practice, however, training trajectories are often single-path because they are derived from a single gold caption: back-deriving a rationale from one answer yields only one route.}

\textcolor{black}{As shown in Fig.~\ref{fig:single-path}, when trained with the final answer as a single path, the model collapses the reasoning process into a single decoding step, failing to develop true association and in-depth understanding.
It only establishes a superficial mapping from user input to the answer, leading to superficial captions and inability to adapt to the template nature of memes.
Therefore, it is necessary to first explore the template's latent intent, and deliberately generate multiple semantic association possibilities under this template to support the flexible use of the template's high-level meaning.}
To address this, we conceptualize the meme understanding and reasoning process as a hierarchical chain-of-thought $r=(r_{\text{tmpl}}, r_{\text{cont}})$, which explicitly decouples template-level interpretation from context-level grounding.
Captions are then realized by sampling from $P_\phi(c \mid r, I)$.

\begin{figure*}[!htbp]
    \centering
    \includegraphics[width=0.85\linewidth]{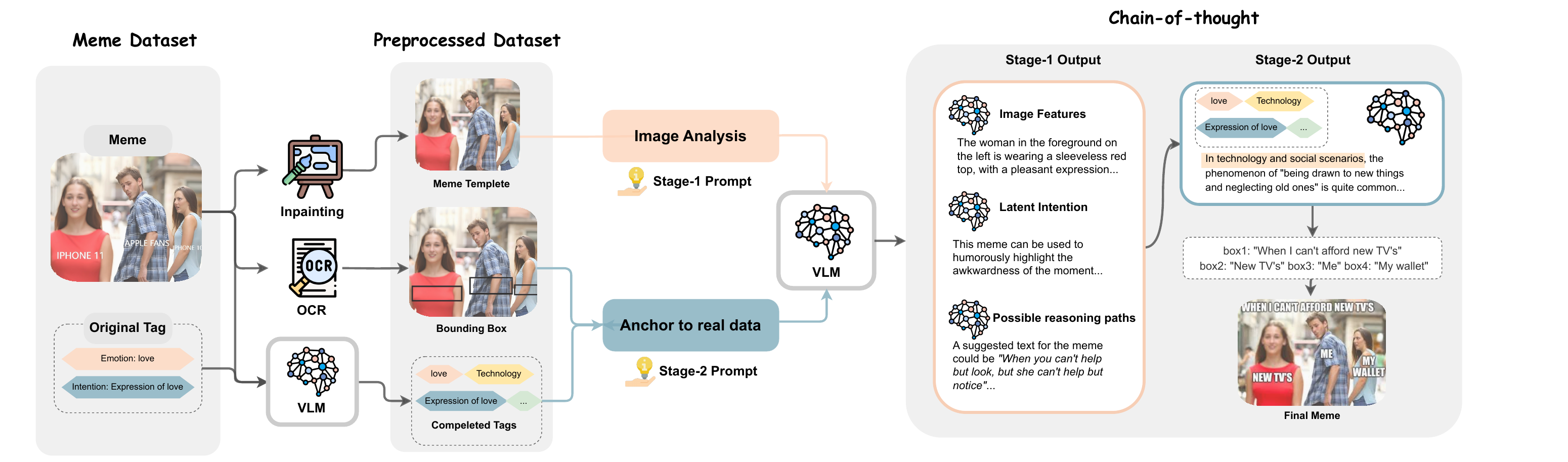}
    \caption{This diagram shows the dataflow for constructing hierarchical CoT supervisions. 
    Stage~1 explores multiple reasoning paths that bind a template to different context-specific details. 
    Stage~2 anchors one high-quality path from ground-truth, preserving diversity while preventing collapse.}
    \label{fig:cot_data}
    \vspace{-10pt}
\end{figure*}

To approximate human authorship, we supervise CoT in two stages. \textcolor{black}{The process is shown in Fig.~\ref{fig:cot_data}.
In Stage~1, we first train the model \(P_{\phi}(r \mid I,\hat U)\) with \emph{multi-path} reasoning traces synthesized by auxiliary LLM “teachers” under \((I,\hat U)\), where \(\hat U\) is a candidate set of \emph{potential user contexts} (e.g., emotions, intentions, scenarios) suggested by the template’s affordances (Appendix~\ref{app:cot_details}). At inference, the model explores multiple reasoning paths conditioned only on $I$, while hypothesizing a candidate set of \emph{potential user contexts} $\hat U$ (e.g., emotions, intentions, or scenarios a user might want to express). Concretely, the model generates reasoning candidates (multiple associative scenarios) $\{ r^{(i)} \}_{i=1}^N \sim P_\phi(r \mid I, \hat U)$, encouraging broad coverage of diverse interpretations.
This part is similar to how humans brainstorm several possible jokes before finalizing one. }

\textcolor{black}{In Stage~2, when groundtruth captions are available, we anchor one path $\tilde r$ to be consistent with the punchlines of real memes (i.e., ground-truth captions) by incorporating the \emph{actual user context} $U$, which is inferred from ground-truth captions.
\textcolor{black}{Formally,} we select $\tilde r = \arg\max_{r} P_\phi(c \mid r, I, U)$, which ensures trajectory consistency while preserving the diversity acquired in Stage~1. At inference time (no gold caption), the generator ranks and selects among Stage~1 paths using its internal scoring/decoding policy (see Appendix~\ref{app:cot_details} for construction details and examples).}

The benefit can be formalized as follows.
Let $\tilde h_G:\mathcal{R}\to[0,1]$ denote group-relative humor measure defined over reasoning paths.
Suppose there exists a set of “star” paths (i.e., better paths) $R^\star$ with probability mass $\alpha>0$ under the reasoning distribution, and the average humor gap between non-star paths and the best paths is bounded by $\delta \geq 0$.
Then, we have the following guarantee:

\begin{proposition}[Conditional humor lower bound]
\label{prop:cot-lb}
Normalizing $\max \tilde h_G=1$, the expected humor after two stages CoT supervision satisfies:
\[
\mathbb{E}_{r \sim P_\theta}[\tilde h_G(r)] \;\ge\; 1-(1-\alpha)\delta.
\]
\end{proposition}


Intuitively, as long as promising reasoning paths retain non-negligible probability ($\alpha$ is not too small) and the remaining paths are only moderately worse (small $\delta$), exploration and anchoring preserve a nontrivial lower bound on expected humor.
\textcolor{black}{Meanwhile, Stage~1 exploration maintains multi-hypothesis diversity, preventing entropy from collapsing in the no-exploration limit, while Stage~2 anchoring ensures that a non-negligible probability mass remains concentrated on promising paths.}
Thus, our CoT framework broadens the range of interpretations without sacrificing quality.
However, although anchoring toward ground-truth paths naturally supports a sufficiently large $\alpha$, the humor gap $\delta$ remains uncontrolled, as some generated paths may still be much less funny than others.
Reducing $\delta$ therefore requires an additional mechanism that captures human humor preferences and guides optimization beyond imitation.

\subsection{Reward Modeling from Pairwise Preferences}
\label{sec:reward}

The ideal learning objective would be to recover the latent humor function $h_G(m)$ for each meme $m$.
Since humor is inherently subjective and lacks a global scale, this is infeasible in practice.
We therefore adopt an \emph{order-consistent} view of reward modeling (following established theory~\citep{sun2025rethinking}) and instantiate it in our \emph{group-wise} meme setting.
Under this formulation, the reward serves as a \emph{within-group surrogate} of $h_G$, trained only from relative judgments, avoiding ill-posed cross-group calibration.
Intuitively, hierarchical CoT ensures that high-quality paths retain a meaningful probability mass (the $\alpha$ condition via Stage~2 anchoring), while the reward model supplies the preference signal necessary to \emph{shrink the average gap among plausible paths} (addressing the $\delta$ condition).
This transforms open-ended exploration into a tractable selection problem.

Each meme $m=(I,c)$ is encoded to a feature vector $\Psi(m)\in\mathbb{R}^d$ using a VLM as the encoder. Let a scoring head $f_\phi:\mathbb{R}^d\to\mathbb{R}$ \textcolor{black}{map this feature vector $\Psi(m)$ to a scalar score.we denote this score as $s_\phi(m) = f_\phi\bigl(\Psi(m)\bigr)$.}
For any pair of memes $(m_i,m_j)$ from the same group $G$, we define the predicted preference probability as:
\begin{equation}
\label{eq:rm-prob}
\widehat{p}^{\,G}_{ij} = \sigma\big(s_\phi(m_i)-s_\phi(m_j)\big),
\end{equation}
where $\sigma(\cdot)$ denotes the logistic function; The model is trained by minimizing the binary cross-entropy over human-annotated or auto-generated preference pairs.

Building upon the reward modeling formulation in Eq.~\ref{eq:rm-prob}, we now formalize two key theoretical properties (order consistency and stability) that justify its use in our within-group meme setting. 

\begin{proposition}[Rank consistency]
\label{prop:rank-consistency}
Under the observation model of Eq.~\ref{eq:generic-link} with any strictly increasing link function, minimizing the pairwise preference loss recovers the same within-group ordering as the latent humor function $h_G$.
Complete proofs are provided in Appendix~\ref{app:reward_proof}.
\end{proposition}

The empirical reliability of this rank consistency across different backbones is further analyzed in Appendix~\ref{app:content_reward}.

\begin{proposition}[Robustness to label noise (margin-aware)]
\label{prop:noise-robustness}
Let $\Delta_{ij}^G = h_G(m_i)-h_G(m_j)$ denote the true humor gap, and assume the annotation process has pairwise error rate $\varepsilon$.
For pairs satisfying $|\Delta_{ij}^G|\ge \delta$, the probability of order reversal is bounded above by a function decreasing in $\delta$ and increasing in $\varepsilon$; large humor gaps are therefore preserved even under noisy labels.
\end{proposition}

These propositions, while grounded in the order-consistent analysis of~\citet{sun2025rethinking}, are specifically instantiated under our group-wise comparability.
They serve as the theoretical \emph{drivers} to reduce the humor gap $\delta$ after CoT has secured $\alpha$.
Since pairwise data can be sparse, we aggregate $\widehat{p}^{\,G}_{ij}$ into a coherent within-group ranking via \emph{Expected Borda Count (EBC)} (see Appendix~\ref{app:reward_ebc} for more explanations and implementations).
For a candidate set $\mathcal{S}_G$, each meme’s EBC score equals its expected number of wins against others under the model in Eq.~\ref{eq:rm-prob}.
This provides a stable training target, and inherits expected order consistency when the pairwise model is consistent (Appendix~\ref{app:reward_proof}).
\textcolor{black}{To complement this outcome-oriented supervision, we further incorporate auxiliary rewards to regulate the intermediate reasoning path (Appendix~\ref{app:aux_rewards}), comprising a structural format reward (Appendix~\ref{app:format_reward}) and \textcolor{black}{a monotonicity-validated content reward (Appendix~\ref{app:content_reward}).}}
Detailed constructs and examples of pairwise data are provided in Appendix~\ref{app:reward_data}.

\subsection{Group-wise Policy Optimization}
\label{sec:method-rl}

Following the CoT supervision stage and reward model training, we further fine-tune the meme generator to \emph{increase} the probability of higher-ranked captions.
Concretely, we leverage the trained reward model and adopt a \textbf{Group-wise Relative Policy Optimization (GRPO)} objective~\citet{shao2024deepseekmathpushinglimitsmathematical}.
For a candidate set of memes $\mathcal{S}_G$ with ranking $q_G$ from EBC, the fine-tuning loss is:
\begin{equation}
\label{eq:grpo-loss}
\begin{aligned}
\mathcal{L}_{\text{GRPO}}(\theta) = 
\mathbb{E}_{(I,G)} \left[ -\sum_{m_k \in \mathcal{S}_G} q_G(m_k) \log \pi_\theta(c_k \mid I) \right]
+ \beta \, \mathbb{E}_{I} \left[ \mathrm{KL}(\pi_\theta(\cdot \mid I) \, \Vert \, \pi_{\text{ref}}(\cdot \mid I)) \right]
\end{aligned}
\end{equation}
where $\pi_{\text{ref}}$ denotes the policy obtained after CoT training.
The first term aligns $\pi_\theta$ with the group-local preference distribution $q_G$ (rank-consistent with $h_G$).
Since prior \textcolor{black}{preference} optimization analyses ~\citet{christiano2023deepreinforcementlearninghuman,neu2012apprenticeshiplearningusinginverse,haarnoja2018softactorcriticoffpolicymaximum} often propose optimistic lower bounds, we \textcolor{black}{adopt a corrected, KL-controlled stability guarantee} that holds under our setting and noise model.
Specifically, the original upper bound on the humor-score deviation (induced by preference noise) can be refined under GRPO into a bound that scales with the KL between the trained policy and the reference policy (proof in Appendix~\ref{app:grpo_proof}).

\begin{proposition}[Bounded change of expected humor under GRPO]
\label{prop:grpo-bound}
Assume Proposition~\ref{prop:rank-consistency} holds and $h_G\in[0,1]$. Let $\Delta_{\mathrm{KL}}=\mathbb{E}_I[\mathrm{KL}(\pi_\theta(\cdot\mid I)\,\|\,\pi_{\text{ref}}(\cdot\mid I))]$.
Then
\begin{equation}
\begin{aligned}
\mathbb{E}_{(I,G)} \Big[ \mathbb{E}_{c\sim \pi_\theta} h_G(I,c) \Big] 
\ge \mathbb{E}_{(I,G)} \Big[ \mathbb{E}_{c\sim \pi_{\text{ref}}} h_G(I,c) \Big] 
\quad - \sqrt{\tfrac{1}{2} \Delta_{\mathrm{KL}}}
\end{aligned}
\end{equation}
Hence, if GRPO enforces $\Delta_{\mathrm{KL}}\le \tau$, the expected humor cannot drop by more than $\sqrt{\tau/2}$; \textcolor{black}{the first term encourages movement toward the group-local preference distribution $q_G$, while the KL term makes any possible degradation explicit and bounded.}
\end{proposition}

This bound, derived via Pinsker’s inequality, formalizes the stability underlying our approach in practice: CoT supervision supplies sufficient support ($\alpha$), the reward model and EBC induce a group-local order that reduces $\delta$, and GRPO turns this order into controlled policy updates. 
In sum, our use of order-consistent surrogates aligns with established theory, but the \emph{group-wise instantiation}, the \emph{corrected KL-based bound}, and the \emph{integration with multi-path CoT for open-ended generation} are key ingredients that make the approach effective and verifiable for meme generation.

\vspace{-6pt}
\section{Experiment}
\vspace{-4pt}
\label{sec:experiments}

\subsection{Meme Quality and Diversity with HUMOR}
\label{sec:meme_quality_diversity}

\paragraph{Settings:} We evaluate HUMOR against competitive baselines and model variants, including multiple open-source and closed-source VLMs and our \textit{HUMOR-CoT}, which is fine-tuned with hierarchical CoT. To study the effectiveness of alternative CoT strategies for meme generation, we also include several advanced CoT frameworks\textcolor{black}{~\citep{kim2023cot,chen2024huatuogpt}}, all trained under the same data and protocol. Detailed training settings and dataset statistics are provided in Appendix~\ref{app:training_setting} and ~\ref{app:dataset_stats}.

Given the open-ended, human-aligned nature of meme generation, we prioritize human evaluation, where annotators score generated memes along four predefined quality axes. We additionally report the conventional text-level similarity between generated captions and their reference texts. To measure diversity, we introduce \textbf{Distance under Context Swap}, which replaces the original training context with a randomly selected one (kept fixed across models) and computes the textual divergence between the resulting caption and the original; a larger distance indicates less overfitting to SFT labels and better adaptation to new contexts.

Because VLM-based rubric scores are unstable for meme evaluation (Sec.~\ref{sec:vlm_reliability}), we report \textit{Human Rate} as a complementary VLM-based metric.
\textcolor{black}{Human Rate is a Gemini-based binary human/AI discriminability metric that estimates whether a rendered meme appears human-generated; we use it as a human-likeness proxy rather than an absolute humor score. Full metric definition, evaluator selection, and human-alignment validation are provided in Appendix~\ref{app:VLM-evaluator-details}.}
\textcolor{black}{Here we use Gemini-2.5-pro for this metric, as it shows the most stable and consistent behavior among candidate VLM evaluators in our reliability analysis (Appendix~\ref{app:VLM-evaluator-details}). Appendix~\ref{app:evaluation_setting} offers definitions of all metrics and evaluation criteria.}

\begin{table*}[ht]
\centering
\normalsize
\renewcommand{\arraystretch}{1.2}
\setlength{\tabcolsep}{5pt}
\caption{\textbf{Evaluation results across open-source models, closed-source models, and Qwen2.5-7B-Instruct fine-tuned with our proposed and different CoT methods.} 
Metrics include context-swap distance (diversity criterion), text-level similarity (sim. to original meme text), human evaluation (Humor, Readability, Relevance, Originality), and Human Rate.}
\resizebox{\textwidth}{!}{
\begin{tabular}{llcccccccccccc}
\toprule[1.5pt]
\multicolumn{2}{l}{\multirow{2}{*}{\large Category / Model}} 
& \multicolumn{4}{c}{Human Evaluation (0-5) $\uparrow$} 
& \multirow{2}{*}{\shortstack[c]{Text-level\\Similarity\ $\uparrow$}}
& \multirow{2}{*}{\shortstack[c]{Context-swap\\Distance $\uparrow$}}
& \multirow{2}{*}{Human Rate (\%)$\uparrow$} \\
\cmidrule(lr){3-6}
&
& Humor & Readability & Relevance & Originality \\ 
\midrule[1pt]
\rowcolor{gray!15}
\multicolumn{9}{l}{\textbf{Open-source Models}} \\
& Qwen2.5-7B-Instruct~\citep{bai2025qwen2}  & 2.39  & 3.35  &2.91  &2.57   & \textcolor{black}{0.576}  &  0.564 &75.7   \\
& Qwen2.5-32B-Instruct~\citep{bai2025qwen2}  & 2.54  & 3.52  & 3.09   &2.76   &\textcolor{black}{0.564}   & 0.566  & 82.2  \\
& InternVL3-8B~\citep{zhu2025internvl3}    &2.39   &  2.79 & 3.04  & 2.79  & \textcolor{black}{0.545}  & 0.564  & 62.7  \\
& GLM-4.1V-9B-Thinking~\citep{hong2025glm}   & 1.73  & 2.62  & 2.75  & 2.71  &  \textcolor{black}{0.602} & 0.572  & 45.1  \\
& Keye-VL-8B-preview~\citep{team2025kwai}   &  2.35 & 3.19  & 2.99  & 2.71  & \textcolor{black}{0.585}  &0.580   & 69.0  \\
\midrule[1pt]
\rowcolor{gray!15}
\multicolumn{9}{l}{\textbf{Closed-source Models}} \\
& GPT-4o~\citep{openai_gpt4o_blog_2024}   &  2.70  & 2.99  &  3.21 & \textbf{2.97}   & \textcolor{black}{0.603}  & 0.552  & 91.3  \\
& Gemini-2.5-flash~\citep{comanici2025gemini}  &  \underline{2.81}  & 3.29  &3.25   &2.88  & \textcolor{black}{0.600}& 0.561  &  -  \\
\midrule[1pt]
\rowcolor{gray!15}
\multicolumn{9}{l}{\textbf{Fine-tuned Model}} \\
& HUMOR-CoT    & 2.68  & \textbf{3.70}   & \underline{3.50}   &  \underline{2.90}  & \textcolor{black}{\textbf{0.640}}  &    \underline{0.590} &  \underline{91.5}  \\
& CoT with Single Path~\citep{kim2023cot}   & 1.87  &2.79   & 2.68  &2.45   & \textcolor{black}{\underline{0.637}}   & 0.570  &  86.0 \\
& CoT with Self-Improve~\citep{chen2024huatuogpt}   & 2.38  &  \underline{3.68}  & 3.00  &2.65   &  \textcolor{black}{0.629} & 0.578 & 89.1  \\
& CoT with Subquestion~\citep{wei2022chain}  & 1.85  & 3.32  & 2.58  &  2.47 &\textcolor{black}{0.639}   & \textbf{0.597}  &  87.2 \\
& HUMOR-RL (preview)  & \textbf{2.83}  & 3.67   & \textbf{3.55}   &  2.79  & \textcolor{black}{0.631}  &   0.588 &  \textbf{92.3}   \\

\bottomrule[1.5pt]
\end{tabular}
}
\label{tab:main}
\end{table*}

\paragraph{Results:}
As summarized in Table~\ref{tab:main}, the proposed \textit{HUMOR} framework achieves substantial improvements across multiple evaluation dimensions, validating its efficacy for humor-oriented meme generation.
Specifically, in terms of \textit{Humor}, \textit{HUMOR-CoT} attains a score of 2.68, surpassing the base model Qwen2.5-7B-Instruct (2.39). 
Qualitative analysis suggests that \textit{HUMOR}-improved models better capture nuanced humor mechanisms such as sarcasm and self-mockery.
For \textbf{Readability}, \textit{HUMOR-CoT} achieves a score of 3.70, outperforming all compared variants—including powerful closed-source models.
It can generate meme texts with appropriate length and engaging structure, avoiding the verbosity common in many VLMs while maintaining humor expressivity, thereby better aligning with human writing conventions.
It also excels in \textbf{theme relevance} and \textbf{originality}, demonstrating an ability to interpret deeper user intent rather than superficially referencing visual content.
Although semantic similarity is less indicative for meme captions—which often consist of short phrases, \textit{HUMOR-CoT} still achieves the closest alignment to reference captions among all models.
Our proposed \textbf{Context-Swap Distance} metric further reveals that HUMOR-CoT (0.590) exceeds the baseline (0.564), indicating stronger generalization and context adaptability when user inputs are altered. 
This supports the hypothesis that hierarchical CoT reduces overfitting to concrete training labels.
Finally, \textit{HUMOR-CoT} achieves a \textcolor{black}{Human Rate} of over 91\%, significantly outperforming the base model (75.7\%) and even surpassing the closed-source GPT-4o (91.3\%).

\textcolor{black}{Ablations on alternative CoT variants further illustrates the superiority of HUMOR:
while \textit{Single Path} lacks bottom-up visual grounding and produces narrow reasoning chains; \textit{Self-Improve} attains high readability, it yields conservative, ``safe but dull'' outputs; \textit{Subquestion} mitigates overfitting but suffers from over-decomposition, impairing humor and relevance.
In contrast, HUMOR-CoT’s two-stage reasoning more closely emulates human cognition process for better meme generation.
Beyond human and text-level evaluations, we further validate model alignment through a VLM-based reclassification test (Appendix~\ref{app:VLM Classification Result}).
As summarized in Table~\ref{tab:reclassify}, HUMOR-CoT consistently surpasses both the Qwen2.5-7B and Qwen2.5-32B base models across all four semantic dimensions—emotion, intention, theme, and style.
Notably, despite being trained on the smaller 7B backbone, HUMOR-CoT even outperforms the 32B variant, demonstrating that the hierarchical CoT design contributes more effectively to user-intent preservation than scaling model size alone.}

\paragraph{\textcolor{black}{Retrieval-augmented emerging memes.}}
\textcolor{black}{We further examine whether HUMOR-style reasoning can be combined with retrieval for evolving meme usages. In Appendix~\ref{app:retrieval_emerging_memes}, we provide a small qualitative case study where recent meme-trend descriptions are retrieved and inserted into the prompt. Retrieval supplies external and time-sensitive context, while HUMOR improves how the VLM interprets the template, explores plausible communicative intents, and generates captions under local preference structure. This supports our claim that retrieval-based adaptation and HUMOR are complementary rather than competing approaches.}

\paragraph{\textcolor{black}{Cross-domain pilot.}}
\textcolor{black}{To probe whether the same reasoning and group-wise selection procedure can be instantiated beyond meme generation, we additionally conduct a small pilot on persona-conditioned visual captioning (Appendix~\ref{app:cross_domain_pilot}). This experiment is not intended as a full transfer benchmark, but as directional evidence: under coherent image-persona groups, HUMOR-style multi-path reasoning improves semantic candidate diversity and local pairwise preference outcomes compared with direct generation.}

\subsection{VLM Reliability Evaluation}
\label{sec:vlm_reliability}

\begin{figure*}[!htbp]
    \centering
    \includegraphics[width=1\linewidth]{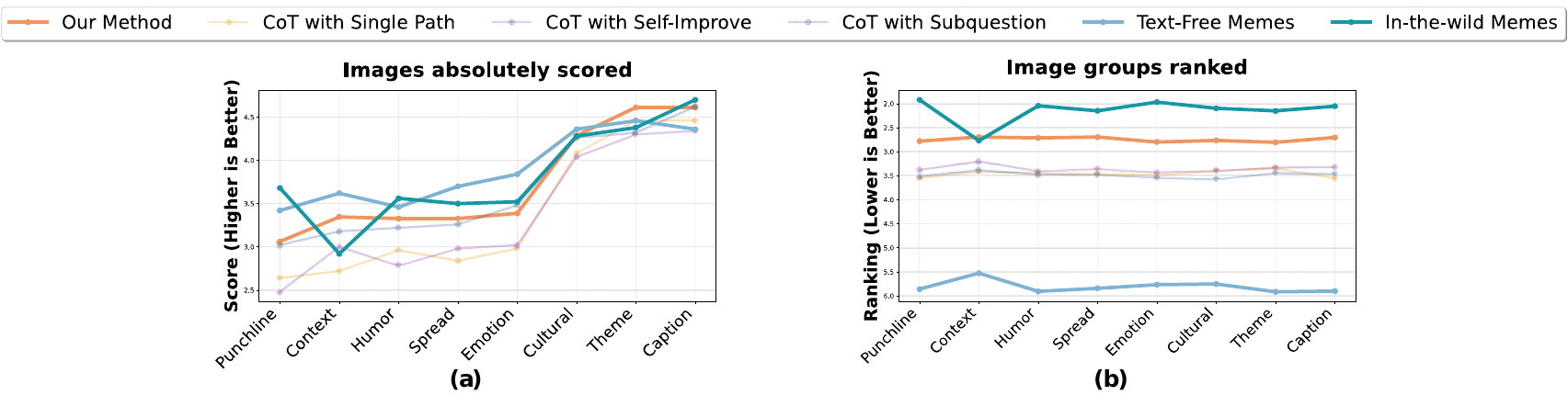}
    \vspace{-14pt}
    \caption{(a) VLM-based \textbf{absolute scoring} fails to distinguish meme quality. (b) \textbf{Group-wise ranking} produces more reliable distinctions, better aligned with human.}
    \vspace{-10pt}
    \label{fig:abs-vs-rank}
\end{figure*}
\begin{figure*}[!htbp]
    \vspace{-12pt}
    \centering
    \begin{minipage}[b]{0.53\textwidth}
        \centering
        \includegraphics[width=\linewidth]{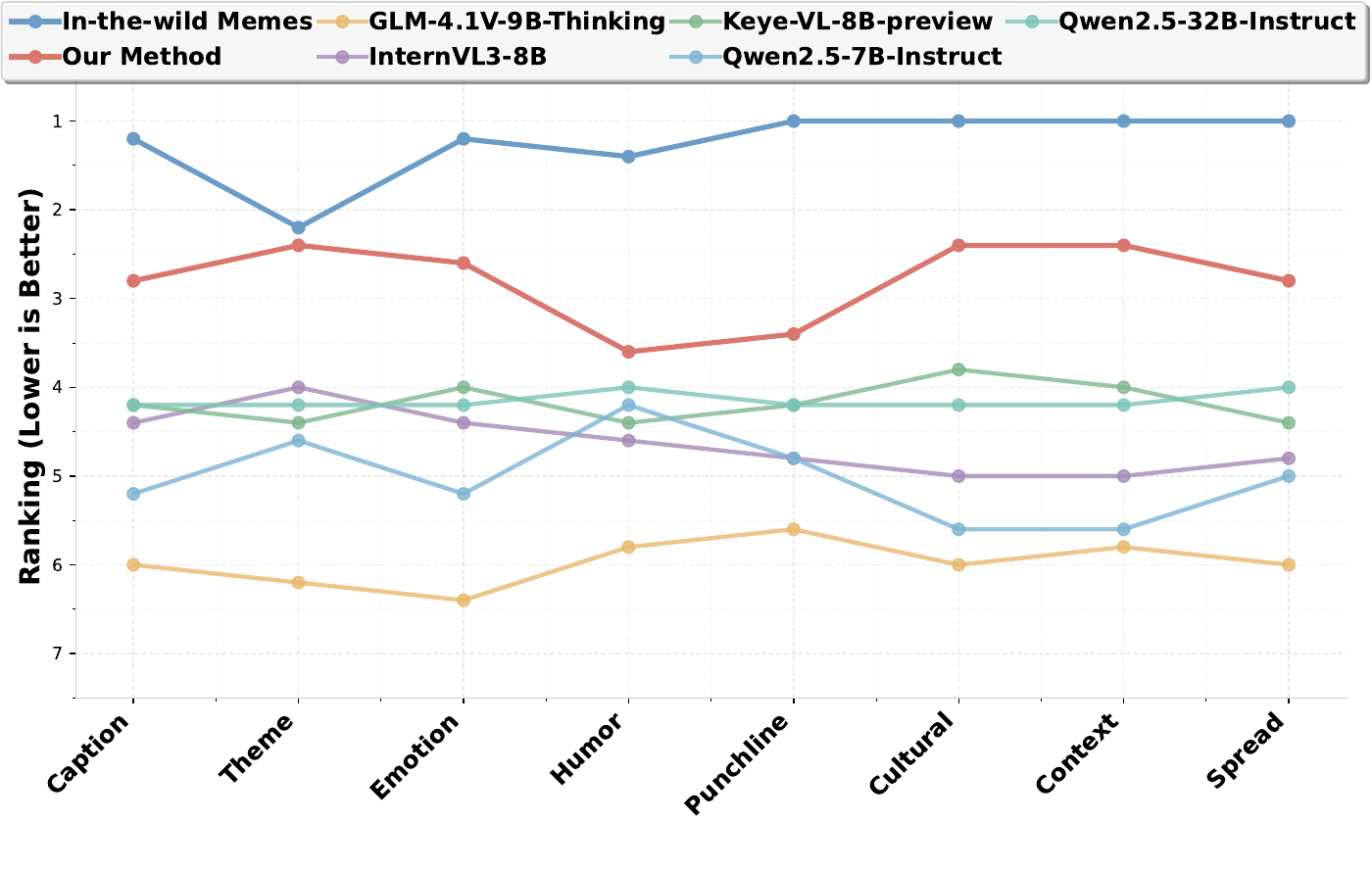}
        \caption{\textcolor{black}{Group-wise ranking results on 20 unseen meme templates. 
Lower ranking is better. HUMOR-CoT generalizes well and remains competitive with human-created memes (in-the-wild memes).}}
        \label{fig:unseen_rank}
    \end{minipage}
    \hfill
    \begin{minipage}[b]{0.45\textwidth}
        \centering
        \includegraphics[width=\linewidth]{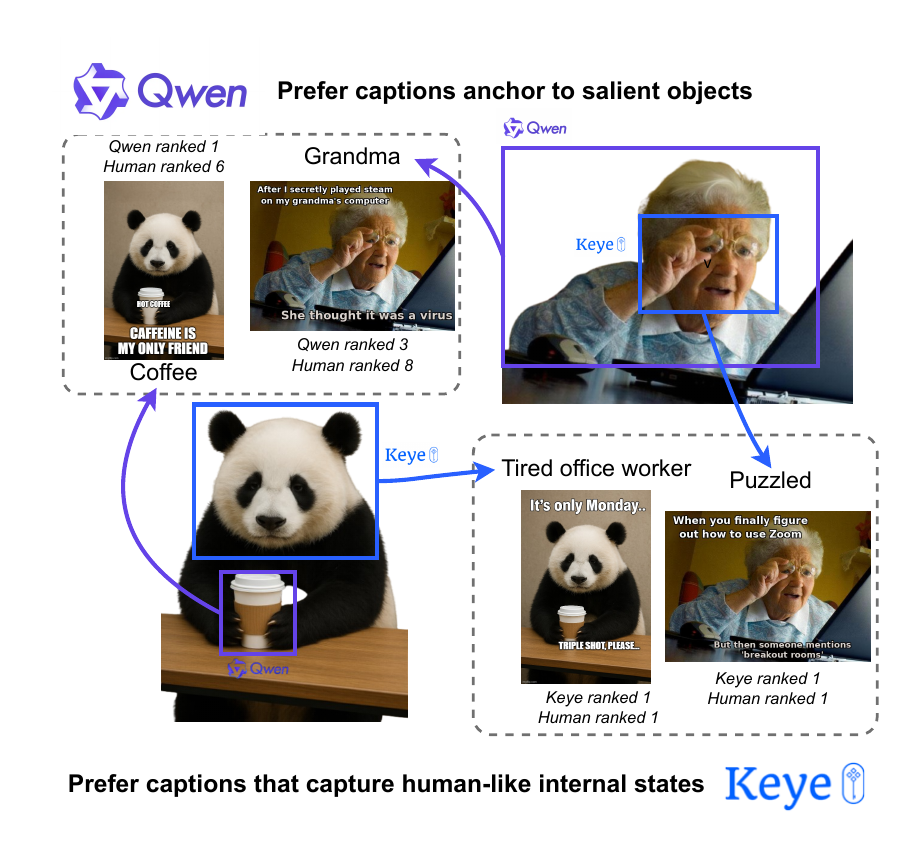}
        \caption{\textit{Qwen2.5-VL-7B} prefers captions that mention direct objects, whereas \textit{Keye-VL-8B} prefers captions reflecting the human-like perception and understanding.}

        \label{fig:rm_base}
    \end{minipage}
\end{figure*}

After the CoT-based experiments, we further examined the reliability of VLM-based scoring for meme evaluation. In practice, existing VLMs often fail to align with human judgment: even for clearly distinct examples such as \textit{In-the-wild Memes} (human-created and high-quality) versus \textit{Text-Free Memes} (text removed), their absolute scores remain nearly identical, revealing that absolute scoring is inadequate for assessing humor or cultural nuance.
\textcolor{black}{As shown in Fig.~\ref{fig:abs-vs-rank}(b), the group-wise relative ranking protocol produces much clearer distinctions between high- and low-quality memes and aligns well with human perception. A human study further validates that these rankings capture genuine preference structures, showing strong agreement with \textit{Gemini-2.5-pro} (Spearman~\textbf{0.72}, Kendall’s~$\tau$~\textbf{0.63}); full details are provided in Appendix~\ref{app:groupwise_alignment}. Under this reliable evaluation protocol, HUMOR-CoT ranks second only to human-created memes and consistently surpasses all CoT-based training baselines.}
\textcolor{black}{Building on this reliable ranking framework, we further assess HUMOR’s ability to generalize to meme templates entirely unseen during training. We evaluate 20 novel templates with no image–text overlap with the training corpus.
Gemini-2.5-pro jointly ranks outputs from different variants.
As shown in Fig.~\ref{fig:unseen_rank}, HUMOR-CoT again ranks second only to human-created memes, mirroring the in-distribution trend. This demonstrates strong zero-shot robustness: the hierarchical CoT effectively transfers its learned humor construction to unfamiliar formats rather than overfitting to template-specific patterns.}
\textcolor{black}{For completeness, the full evaluation prompts are provided in Appendix~\ref{app:ranking_prompt}, detailed experimental settings in Appendix~\ref{app:VLM evaluates experimental setup}, and representative outputs comparing different CoT reasoning schemes in Appendix~\ref{app:Generated Samples}. Additional analyses—including risk-case identification (Appendix~\ref{app:risk_case}), failure-case diagnostics (Appendix~\ref{app:failure_case}), real-world application (Appendix~\ref{app:workplace_application}) and generalization to Unseen templates (Appendix~\ref{app:unseen_templates})—offer further qualitative and quantitative evidence supporting the robustness and interpretability of HUMOR-CoT.}

\subsection{Reward Model Rank Consistency and RL Training}
\label{sec:rm_rank_and_rl}
\vspace{-4pt}

Table~\ref{tab:rm} evaluates reward models trained with the group-wise ranking strategy.
These models are fine-tuned from different base models to align with human preference rankings (see Appendix~\ref{app:rm_training_setting}).
For evaluation, we use five meme templates, \textit{Image1–Image5}, each containing 10--15 candidate memes (Figure~\ref{fig:rank_dataset}). For each template, we obtain a \emph{group-level} human ranking via MaxDiff (Appendix~\ref{app:max_diff}), and the resulting rankings for these example templates are shown in Appendix~\ref{app:Meme Ranking Result}.
Model rankings are obtained by (i) collecting in-group pairwise comparisons from either the base model or the fine-tuned reward model (\textit{HUMOR-RM}), and (ii) aggregating them with Expected Borda Count (EBC) to produce a more reasonable sequence ranking. We report Kendall's $\tau$ and its $p$-value to evaluate the rank consistency objective (Section~\ref{sec:reward}). \textit{HUMOR-RM} built on \textit{Keye-VL-8B} achieves consistently high $\tau$ with significant $p$-values, often $p\!\le\!10^{-3}$, across Image1--Image5, indicating strong within-group agreement with human preferences. Results on \textit{Qwen2.5-VL-7B} are mixed, showing moderate alignment on some templates but near-chance behavior on others, with inconsistent significance. \textit{Qwen2.5-VL-32B} and other backbones show limited gains. \textcolor{black}{Overall, the group-wise fine-tuning usually improves or preserves ranking consistency under the same training and ranking supervision, with the strongest and most statistically reliable gains appearing on the semantically stronger Keye-VL backbone.}


\begin{table*}[t]
\centering
\small
\setlength{\tabcolsep}{2pt}   
\renewcommand{\arraystretch}{1.08}
\caption{(Reward Model) Ranking results of different baselines among distinct template images. It indicates the change after fine-tuning relative to the baseline: an increase in Kendall tau $\tau$ and a decrease in p-value $p$ represent improvements (highlighted in green), while the opposite indicates deterioration (shown in red). Significance levels: * $p<0.05$; ** $p<0.01$; *** $p<0.001$.}
\begin{tabular}{l*{5}{cc}}
\toprule
\multirow{2}{*}{Model} &
\multicolumn{2}{c}{\textbf{Template 1}} &
\multicolumn{2}{c}{\textbf{Template 2}} &
\multicolumn{2}{c}{\textbf{Template 3}} &
\multicolumn{2}{c}{\textbf{Template 4}} &
\multicolumn{2}{c}{\textbf{Template 5}} \\
\cmidrule(lr){2-3}\cmidrule(lr){4-5}\cmidrule(lr){6-7}\cmidrule(lr){8-9}\cmidrule(lr){10-11}
 & $\tau\uparrow$ & $\text{$p$}\downarrow$ & $\tau\uparrow$ & $\text{$p$}\downarrow$ & $\tau\uparrow$ & $\text{$p$}\downarrow$ & $\tau\uparrow$ & $\text{$p$}\downarrow$ & $\tau\uparrow$ & $\text{$p$}\downarrow$ \\
\midrule
\textbf{Qwen2.5-VL-7B (Base)}     & 0.16 & 0.60 & 0.28 & 0.17 & 0.47 & 0.07 & -0.10 & 0.63 & 0.29 & 0.29 \\
\textbf{Qwen2.5-VL-7B (Finetuned)}& 0.47 & 0.07 & 0.56 & 0.03* & 0.42 & 0.11 & 0.14 & 0.50 & 0.47 & 0.07 \\
{\scriptsize$\Delta$ vs Base}   & {\scriptsize\textcolor{ForestGreen}{\(+0.31\)}} & {\scriptsize\textcolor{ForestGreen}{\(-0.53\)}} &
                                 {\scriptsize\textcolor{ForestGreen}{\(+0.28\)}} & {\scriptsize\textcolor{ForestGreen}{\(-0.14\)}} &
                                 {\scriptsize\textcolor{BrickRed}{\(-0.04\)}} & {\scriptsize\textcolor{BrickRed}{\(+0.04\)}} &
                                 {\scriptsize\textcolor{ForestGreen}{\(+0.25\)}} & {\scriptsize\textcolor{ForestGreen}{\(-0.13\)}} &
                                 {\scriptsize\textcolor{ForestGreen}{\(+0.18\)}} & {\scriptsize\textcolor{ForestGreen}{\(-0.22\)}} \\
\midrule
\textbf{Qwen2.5-VL-32B (Base)}       & 0.16 & 0.61 & 0.16 & 0.44 & -0.02 & 1.00 & 0.14 & 0.50 & 0.29 & 0.29 \\
\textbf{Qwen2.5-VL-32B (Finetuned)}  & 0.29 & 0.29 & 0.47 & 0.02* & 0.07 & 0.86 & 0.30 & 0.14 & 0.42 & 0.11 \\
{\scriptsize$\Delta$ vs Base}   & {\scriptsize\textcolor{ForestGreen}{\(+0.13\)}} & {\scriptsize\textcolor{ForestGreen}{\(-0.32\)}} &
                                 {\scriptsize\textcolor{ForestGreen}{\(+0.30\)}} & {\scriptsize\textcolor{ForestGreen}{\(-0.42\)}} &
                                 {\scriptsize\textcolor{ForestGreen}{\(+0.09\)}} & {\scriptsize\textcolor{ForestGreen}{\(-0.14\)}} &
                                 {\scriptsize\textcolor{ForestGreen}{\(+0.15\)}} & {\scriptsize\textcolor{ForestGreen}{\(-0.36\)}} &
                                 {\scriptsize\textcolor{ForestGreen}{\(+0.13\)}} & {\scriptsize\textcolor{ForestGreen}{\(-0.18\)}} \\
\midrule
\textbf{Keye-VL-8B (Base)}           & 0.05 & 0.85 & 0.09 & 0.70 & 0.16 & 0.60 & 0.29 & 0.29 & 0.16 & 0.60 \\
\textbf{Keye-VL-8B (Finetuned)}      & 0.78 & 0.00*** & 0.77 & 0.00*** & 0.78 & 0.00*** & 0.78 & 0.00*** & 0.78 & 0.00*** \\
{\scriptsize$\Delta$ vs Base}   & {\scriptsize\textcolor{ForestGreen}{\(+0.73\)}} & {\scriptsize\textcolor{ForestGreen}{\(-0.84\)}} &
                                 {\scriptsize\textcolor{ForestGreen}{\(+0.69\)}} & {\scriptsize\textcolor{ForestGreen}{\(-0.70\)}} &
                                 {\scriptsize\textcolor{ForestGreen}{\(+0.62\)}} & {\scriptsize\textcolor{ForestGreen}{\(-0.60\)}} &
                                 {\scriptsize\textcolor{ForestGreen}{\(+0.49\)}} & {\scriptsize\textcolor{ForestGreen}{\(-0.29\)}} &
                                 {\scriptsize\textcolor{ForestGreen}{\(+0.62\)}} & {\scriptsize\textcolor{ForestGreen}{\(-0.60\)}} \\
\bottomrule
\end{tabular}

\label{tab:rm}
\end{table*}


\subsection{Reward Model Analysis on Different Base Model}
\label{sec:rm_analysis}
Across all evaluated templates (Image 1–5), the \textit{Keye-VL-8B} base model achieves higher in-group ranking consistency with human preferences than \textit{Qwen2.5-VL} variants.
We next examine why the post-training trajectories differ across base models and whether our training scheme induces model-specific preferences.
Here, we present the differences among the top-ranked images preferred by reward models fine-tuned on different base models. 
In Fig.~\ref{fig:rm_base}, 
\textit{Qwen2.5-VL-7B} tends to anchor caption preferences on salient visual objects. 
For instance, when Image 5 depicts a panda holding a coffee cup, it favors captions containing the word "coffee";
Similarly, for Image 2, which shows an older woman looking at a laptop, it prefers references for "grandma" or computer-related terms.
In contrast, \textit{Keye-VL-8B} more consistently captures implied internal states or situational cues within the scene and aligns them with the template’s communicative intent.
In the same examples, it interprets the panda as resembling a "tired office worker" and the woman as appearing "puzzled", which aligns better with human rankings under our evaluation protocol.
These findings aligns with our theoretical expectation: while the reward model supplies only a preference ordering, effective alignment ultimately depends on the base model's capacity to represent the nuanced cues underlying human humor perception.


\vspace{-8pt}
\section{Conclusion}
In this work, we tackled the complex challenge of teaching VLMs the art of in-the-wild meme generation, a task that requires nuanced reasoning beyond standard image captioning.
Our proposed framework, \textit{HUMOR}, successfully bridges the gap from visual perception to humorous punchline by instituting a two-stage process of hierarchical reasoning and preference alignment.
Through a novel hierarchical CoT, the model learns to explore diverse creative paths while anchoring on high-quality outcomes.
Furthermore, by leveraging group-wise preference modeling and RL, we ensure the generated humor aligns with human judgment in a stable and consistent manner.
\textcolor{black}{More broadly, HUMOR points to a conditional framework for subjective open-ended multimodal generation, especially when outputs can be organized into semantically coherent groups and local comparative judgments are more reliable than global scores.}

\section*{Acknowledgement}
We gratefully acknowledge the support from Shanghai Artificial Intelligence Laboratory. The resources and funding provided by the lab significantly contributed to this work.

\bibliographystyle{plainnat}
\bibliography{custom}


\appendix

\clearpage

\section{Related Works}
\subsection{Evolution of Vision-Language Models for Multi-modal Process}
The pursuit of unified vision-language modeling has progressed through three distinct phases of architectural innovation. Early foundational work established bidirectional frameworks for cross-modal understanding: ERNIE-ViLG~\citep{Zhang2021ERNIEViLGUG} and the Unifying Multi-modal Transformer~\citep{10.1145/3474085.3481540} pioneered transformer-based architectures that jointly optimized text-to-image and image-to-text generation through multi-modal tokenization and autoregressive objectives. Concurrently, \textcolor{black}{Zero-Shot Text-to-Image Generation~\citep{Ramesh2021ZeroShotTG}} demonstrated the scalability potential of such approaches through their zero-shot text-to-image generation framework, establishing critical baselines for large-scale multi-modal pretraining.


Contemporary breakthroughs have redefined architectural paradigms through multimodal unification. Models like Show-o~\citep{xie2024show} and MonoFormer~\citep{zhao2024monoformer} successfully fused autoregressive and diffusion mechanisms within singular architectures via shared attention layers. 
\textcolor{black}{Beyond architectural fusion, recent research highlights the critical role of reasoning strategies. Chain-of-Thought (CoT) prompting has been empirically shown to enhance the complex reasoning capabilities of VLMs by eliciting intermediate rationales~\citep{zhang2023multimodal, hu2024visual}.}
Building upon these advancements, our work leverages multi-modal comprehension capabilities to address the unique challenges of meme generation—particularly its requirement for hierarchical reasoning and understanding subjective humor.

\subsection{Meme Analysis, Generation, and Alignment}
\paragraph{Humor Analysis and Generation.} 
\textcolor{black}{Computational humor draws from established linguistic and anthropological theories~\citep{apte1985humor, binsted2006computational} to formally model incongruity and semantic shifts.} 
Internet memes have emerged as a vital component of digital culture, prompting substantial scholarly attention to their multi-modal communications. Extensive research has focused on analyzing topics~\citep{du2020understanding}, semantics~\citep{xu2022met}, and emotions~\citep{sharma2020semeval} conveyed in memes. The evolution of meme generation techniques has progressed through distinct technological phases. Initial systems employed rule-based architectures, exemplified by ~\citet{oliveira2016one}'s template-driven approach using standardized structures like \textit{"One does not simply X"}, and ~\citet{wang2015can}'s dual-channel model integrating textual and visual features. The advent of deep learning catalyzed more sophisticated paradigms. Peirson and Tolunay pioneered this transition with Dank Learning~\citep{peirson2018dank}, combining Inception V3 image encoders with attention-enhanced LSTM decoders. Subsequent innovations introduced transformer architectures: Sadasivam et al.'s MemeBot~\citep{sadasivam2020memebot} and Vyalla et al.'s Memeify~\citep{vyalla2020memeify} demonstrated enhanced text-image alignment through multi-modal fusion.

Recent breakthroughs leverage large language models (LLMs) and VLMs to achieve unprecedented scale. Memecraft~\citep{wang2024memecraft} enables targeted meme creation for social advocacy. Addressing multi-image complexity, Chen et al. proposed XMeCap~\citep{chen2024xmecap}, introducing a two-stage framework with supervised fine-tuning guided by novel similarity metrics. 
Concurrently, benchmark datasets have emerged to evaluate capabilities. MemeCap~\citep{hwang2023memecap} provides metaphor annotations, while the New Yorker benchmarks series~\citep{hessel2022androids} assess humor comprehension through caption matching and explanation tasks. The MCC dataset~\citep{sharma2023memex} further incorporates external knowledge for abstraction analysis.

\paragraph{\textcolor{black}{Retrieval, Preference Optimization, and Our Scope.}}
\textcolor{black}{Retrieval-based meme adaptation, template mining, and trend tracking are useful components for practical meme-generation systems because they supply external context, such as similar meme usages, recent cultural references, or concrete exemplars. HUMOR is complementary to these system-level mechanisms: retrieval provides additional context, while HUMOR improves how the underlying VLM interprets a template, explores multiple plausible contexts, and selects outputs under group-local preference signals. This distinction is important because simply exposing a VLM to meme-related information does not fully solve metaphor, humor, or multimodal-intent reasoning, as prior meme-understanding benchmarks continue to show substantial model failures~\citep{hessel2022androids,hwang2023memecap}. Standard preference-optimization methods such as RLHF/DPO~\citep{ouyang2022training,rafailov2023direct} motivate our alignment stage, but they typically assume a more globally comparable preference signal. HUMOR instead makes local comparability explicit and operationalizes it through within-group pairwise reward modeling. Similar local-preference structures also appear beyond memes, for example in prompt-conditioned image preference datasets such as Pick-a-Pic~\citep{kirstain2023pickapic} and ImageReward~\citep{xu2023imagereward}, and in persona-conditioned captioning~\citep{shuster2019personalitycaptions}.}

\textcolor{black}{While capability has scaled, aligning models with \textit{subjective} human preferences remains a critical frontier.
Unlike objective tasks with ground-truth, humor and creativity require modeling diverse and often noisy judgments.
Recent works have begun to address this by aligning models with diverse human values~\citep{zhou2024beyond} and exploring personalized or pluralistic strategies~\citep{feng2024modular}. 
Specifically in the domain of humor, \citet{song2025large} highlight the challenges of modeling subjective humor preferences using LLMs. Our work advances this direction by proposing a group-wise preference formulation, mitigating the noise inherent in cross-context humor comparison.}

\section{Hierarchical Chain-of-Thoughts of Metaphor}
\label{app:cot_details}

To enhance our model's understanding of humor, we replicated the human meme creation process.
Through extensive analysis of human meme creation, we extracted a paradigm for hierarchical meme feature analysis.

Take the "Distracted Boyfriend" meme as an example. 
Humans first capture: the delighted expression of the woman on the left, the action of the man in the center looking back and his subtle flirtatious gaze, the annoyed posture of the woman on the right, and the triangular compositional relationship and explicit emotional direction formed by the three individuals. 
Humans further abstract this scene and discover that it can be applied to any scenario of infatuation with something new and abandonment of the old, establishing entity mapping relationships. 
Thus, when the user's request is workplace culture, this template can be adapted to depict a leader being attracted by a new employee during a meeting, with a senior employee showing an expression of helplessness, vividly illustrating the workplace "new vs. old" relationship and generating humor.

How would humans fill in the text? Through statistical analysis of 5,000 classic memes, we found that the text positions in common meme templates are fixed, and the text content is highly correlated with its position. For instance, in the "Distracted Boyfriend" template, the position corresponding to the woman on the right is often used to represent the neglected object, the position corresponding to the man in the center represents the subject of attention shift, and the position corresponding to the woman on the left is the newly focused entity. Therefore, we integrate "text content generation" and "text position allocation" in the meme generation process. By annotating text box positions in the image, the model only needs to use its inherent visual localization ability to find the boxes, understand that text needs to be written in specific areas, and then combine spatial semantic mapping relationships to generate text with greater humorous effects in these positions.

We aim to imitate this thought process to construct Chain-of-Thought (CoT) data:

\paragraph{Data Collection and Preprocessing}

\paragraph{Meme Images} We collected over 4,000 meme images from public dataset~\citep{xu2024generating}, and established a multi-dimensional labeling system:

\begin{enumerate}
    \item \textbf{Emotion Classification}: Covers 7 basic emotions and intensity levels.
    \item \textbf{Intent Detection}: Differentiates between 10 creation intents such as offense and entertainment.
    \item \textbf{Metaphor Analysis}: Records metaphorical entities and cross-domain mapping relationships.
\end{enumerate}

\textcolor{black}{\paragraph{Safety-Driven Dataset Cleaning.} 
To mitigate potential risks within the raw dataset—such as political bias, sexually explicit content, and sensitive themes like discrimination—we implemented an automated filtering protocol leveraging the intrinsic safety guardrails of the VLM API (e.g. \texttt{doubao-1.5-vision-pro}). Specifically, during the image understanding phase, we prompted the API to interpret each meme. We adopted a "refusal-based" criterion: instances where the API triggered a safety warning or refused to generate a response were flagged as containing harmful or negative content. These samples were systematically excluded from our training corpus to ensure compliance with ethical safety standards.}

\paragraph{Base Images and Text Content/Position Information} The FLUX.1-dev-Controlnet-Inpainting-Beta model is used to erase and restore the text areas in original memes, obtaining text-free base images. Meanwhile, OCR technology precisely records the (position, content) pairs of text, providing spatial semantic data for subsequent training.

\paragraph{User Requirements} We reconstructed user requirements in reverse using APIs. Taking the meme's labels and final text as inputs, we utilized prompts to reverse-engineer the user's initial request. We analyzed the following dimensions of user requirements: emotion category, emotion intensity, intention, Scene or theme , style preference, and keywords.

\paragraph{CoT Data Generation}

\paragraph{Stage One} Using the base image as input, we extract high-level semantics of the meme.

First, we perform visual element decomposition. Our framework systematically deconstructs meme templates from four key visual dimensions:

\begin{enumerate}
    \item \textbf{Main Subject Characteristics}: Analyze facial expressions, poses, clothing, and dynamic relationships between characters.
    \item \textbf{Composition Logic}: Identify visual focal points, color contrasts, and spatial relationships.
    \item \textbf{Cultural Markers}: Recognize identifiable meme formats and pop culture references.
    \item \textbf{Narrative Threads}: Interpret body language implications and prop symbolism.
\end{enumerate}

Then, we conduct scenario association and humor construction based on visual analysis:

\begin{enumerate}
    \item \textbf{Social Contexts}: Identify scenarios suitable for group chats, comment sections, and private conversations.
    \item \textbf{Topic Relevance}: Establish connections with workplace culture, life dilemmas, and internet hotspots.
    \item \textbf{Emotional Mapping}: Determine appropriate humor techniques, including satire, self-deprecation, exaggeration, and contrast.
\end{enumerate}

\paragraph{Stage Two} Using the base image analysis from Stage One, user requirements, and final text as inputs, we infer the customized creation process for specific requests.

We provide few-shot examples of this parsing process. For instance, for the "Distracted Boyfriend" meme, when Stage One yields the semantic pattern of infatuation with something new and abandonment of the old, and identifies three entity positions: A [attention-shifting subject], B [newly focused entity], and C [neglected object], the user's request is a technology theme with the keyword "Apple fanatic." We consider how to align the expression of infatuation with something new and abandonment of the old with the context of technology product updates to reflect being an Apple fanatic. We infer that the semantic mapping of new and old phones is similar. Therefore, combining this image, we deduce that the text should be filled as: "A: APPLE FANS, B: IPHONE 11, C: IPHONE 10," humorously expressing enthusiasm for Apple's new technological products.

\paragraph{Training Rationale and Process}

We conduct instruction-tuning training using CoT data as supervisory signals. Since our training data contains numerous instances of the same base image, the two-stage CoT process essentially learns metaphorical semantic relationships across different scenarios. It is a divergent associative thinking training where one base image corresponds to multiple scenarios. This CoT approach not only enables the model to understand the high-level semantics of the image itself but also establishes multi-scenario associative capabilities.

\paragraph{Determination and Extraction of Generated Text Format}
Text boxes in the image are marked using a top-to-bottom, left-to-right coordinate sorting rule, and text content is recorded in the labels in order and in box format.
The prompt explicitly requires the model to output in the format "box1:text1, box2:text2."

\begin{figure*}[!htbp]
    \centering
    \includegraphics[width=1\linewidth]{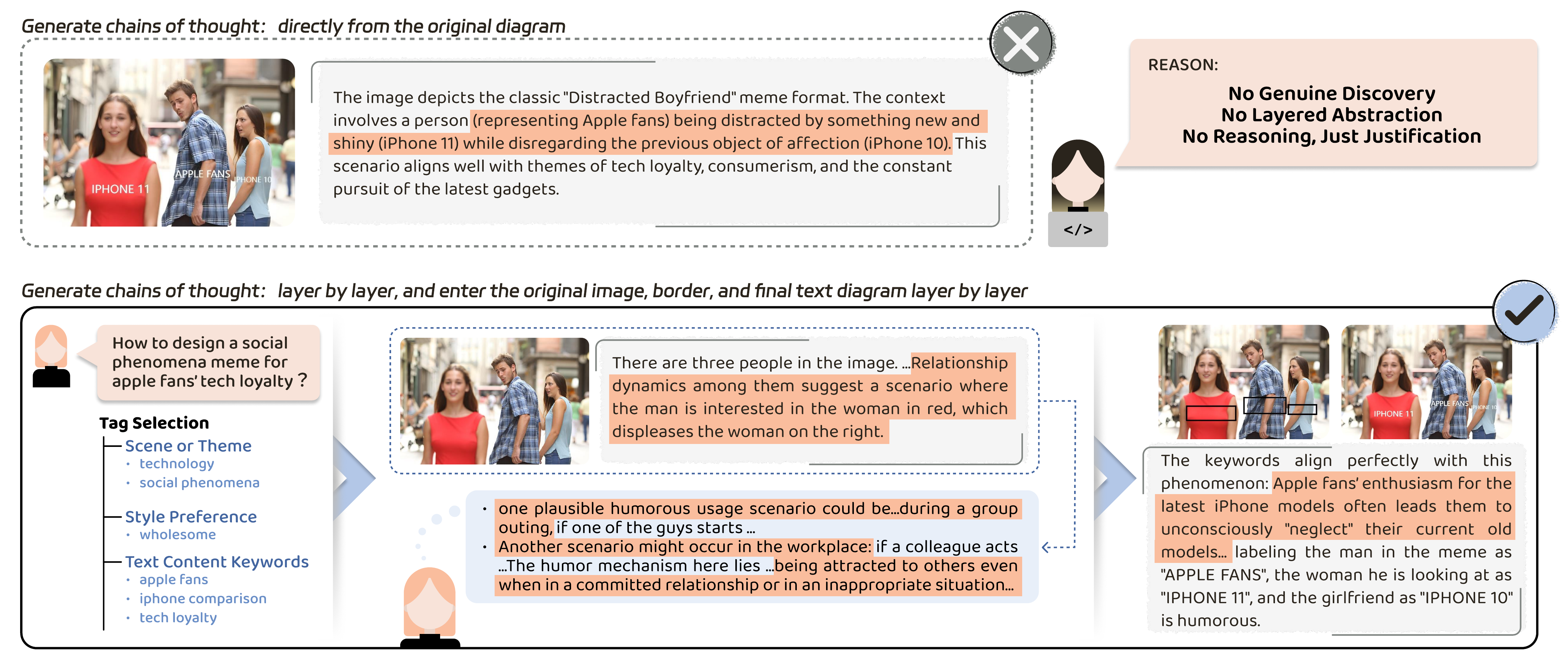}
    \caption{Comparison between direct CoT generation from the original image and our hierarchical CoT generation approach.}
    \vspace{-10pt}
    \label{fig:cot_compare}
\end{figure*}

\paragraph{Critical Comparison: Direct vs. Hierarchical CoT}
The direct approach of generating chains of thought from the original image is essentially reverse engineering rather than genuine reasoning. It suffers from four critical flaws: 1) \textbf{No Genuine Discovery}: it skips the exploratory stage where humor emerges from active associative search, jumping straight to a fixed answer; 2) \textbf{No Layered Abstraction}: it leaps from raw visual details to a specific conclusion without building transferable intermediate metaphors; 3) \textbf{No Reasoning, Just Justification}: instead of true inference, it merely defends a predetermined conclusion.

In contrast, our layered CoT framework mirrors human reasoning by progressively abstracting from visual description to general metaphorical patterns and then to domain-specific humor instantiations, thereby enabling genuine creativity and robust generalization.

\begin{figure*}[!htbp]
    \centering
    \includegraphics[width=1\linewidth]{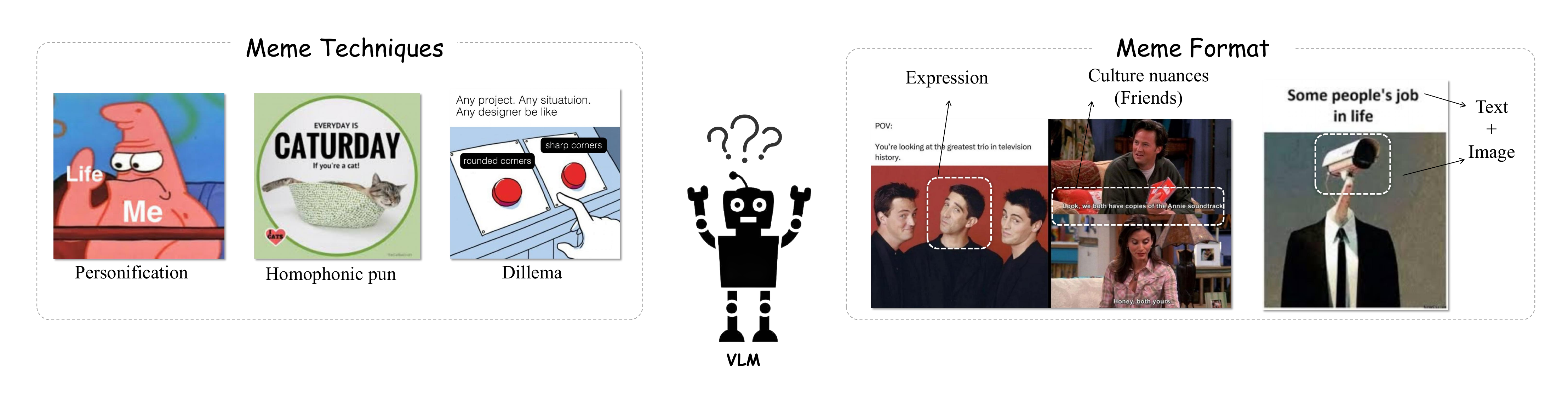}
    \caption{\textcolor{black}{Examples of memes common on the internet}}
    \vspace{-10pt}
    \label{fig:toy_example}
\end{figure*}
\section{Reward Modeling: Assumptions and Proofs}
\label{app:reward_proof}

\subsection{Setup and Assumptions}
For a fixed group $G$, the latent humor functional is $h_G:G\to[0,1]$.
Pairwise labels follow the observation model of Eq.~(1):
\[
\Pr[m_i \succ m_j \mid G] \;=\; \Lambda\big(h_G(m_i)-h_G(m_j)\big),
\]
where $\Lambda:\mathbb{R}\to(0,1)$ is strictly increasing. 
A reward model maps a meme $m=(I,c)$ to a score $s_\phi(m)$; the pairwise probability is 
\[
\hat p^{\,G}_{ij} \;=\; \sigma\!\big(s_\phi(m_i)-s_\phi(m_j)\big),
\]
and $\phi$ is learned by minimizing the empirical pairwise cross-entropy $\mathcal{L}_{\text{pair}}$.
We assume (A1) the data contains i.i.d.\ pairs drawn within $G$ with non-degenerate coverage; 
(A2) the model class for $s_\phi$ is rich enough to fit the Bayes-optimal decision boundary; 
(A3) identifiability is up to an additive constant per group (sufficient for ranking).  

\subsection{Rank Consistency (Proposition~1) — Proof}
\begin{proposition*}[Rank consistency (main text Proposition~1)]
\label{thm:rank-consistency-appendix}
Under Eq.~(1) with strictly increasing $\Lambda$, any risk minimizer of the logistic pairwise loss
recovers the same within-group ordering as $h_G$.
\end{proposition*}

\paragraph{Proof.}
Let $\eta_{ij}=\Pr[m_i \succ m_j\mid G]=\Lambda(\Delta_{ij})$ with $\Delta_{ij}=h_G(m_i)-h_G(m_j)$.
The Bayes-optimal pairwise classifier for logistic loss satisfies $\sigma(s^\star_i-s^\star_j)=\eta_{ij}$, hence
\[
s^\star_i - s^\star_j \;=\; \sigma^{-1}(\eta_{ij}) \;=\; \sigma^{-1}\!\big(\Lambda(\Delta_{ij})\big) \;=:\; \psi(\Delta_{ij}),
\]
where $\psi$ is strictly increasing as a composition of strictly increasing functions. Therefore
\[
s^\star_i - s^\star_j > 0 
\iff \Delta_{ij}>0 
\iff h_G(m_i)>h_G(m_j).
\]
Thus any minimizer (up to additive constants) induces the same strict order as $h_G$ inside $G$.
\hfill$\square$

\subsection{Noise Robustness (Proposition~2) — Proof}
\begin{proposition*}[Noise robustness (main text Proposition~2)]
\label{prop:noise-robustness-appendix}
Let $\Delta_{ij}^G = |h_G(m_i)-h_G(m_j)|$. Suppose the learned classifier has average pairwise error $\varepsilon$.
If we split pairs into ``small-margin'' ($\Delta_{ij}^G<\delta$) and ``large-margin'' ($\Delta_{ij}^G\ge\delta$), then the reversal probability obeys
\begin{equation}
\begin{aligned}
\Pr[\text{reversal}] &\le \Pr[\Delta_{ij}^G < \delta] + \Pr[\text{reversal} \mid \Delta_{ij}^G \ge \delta] \\
&\le \Pr[\Delta_{ij}^G < \delta] + \varepsilon_\delta
\end{aligned}
\end{equation}
where $\varepsilon_\delta$ decreases as $\delta$ increases and increases with the classifier error $\varepsilon$; in particular, under the observation model Eq.~(1), the conditional flipping probability on large-margin pairs is upper-bounded by a monotonically decreasing function of $\delta$.
\end{proposition*}

\paragraph{Proof.}
Let $K$ be the event ``classifier reverses the true order''. Decompose by a margin threshold $\delta>0$:
\[
\Pr[K] 
= \Pr[K \wedge (\Delta_{ij}^G<\delta)] + \Pr[K \wedge (\Delta_{ij}^G\ge\delta)]
\le \Pr[\Delta_{ij}^G<\delta] + \Pr[K\mid \Delta_{ij}^G\ge\delta].
\]
The second term is at most the classifier’s conditional error on large-margin pairs, denoted $\varepsilon_\delta$.
Under Eq.~(1), the Bayes error on a pair decreases monotonically with $|\Delta_{ij}^G|$, hence $\varepsilon_\delta$ decreases in $\delta$.
If the global average error is $\varepsilon$, then $\varepsilon_\delta\le \varepsilon$ and often much smaller.
Thus large true gaps are stably preserved, while flips concentrate on small-margin pairs.
\hfill$\square$

\subsection{From Pairwise to Group Ranking (EBC)}
Given sparsity, we aggregate pairwise probabilities into a within-group ranking via Expected Borda Count (EBC): each item’s score equals its expected number of wins against others according to $\hat p^{\,G}_{ij}$.
EBC is a monotone transformation of the empirical pairwise preferences and inherits rank consistency in expectation when the pairwise model is consistent, providing a coherent group-wise order for evaluation and optimization.
(Operational details as in Sec.~4.2.)

\section{Group-wise Policy Optimization (GRPO): Guarantees and Proofs}
\label{app:grpo_proof}

\subsection{Objective and Notation}
For a candidate set $\mathcal{S}_G$ with group ranking distribution $q_G$ (from EBC), the GRPO loss is
\begin{equation}
\begin{aligned}
\mathcal{L}_{\text{GRPO}}(\theta) &= \mathbb{E}_{(I,G)}\Big[-\sum_{m_k\in\mathcal{S}_G} q_G(m_k)\,\log \pi_\theta(c_k\mid I)\Big] \\
&\quad + \beta\,\mathbb{E}_I\big[\mathrm{KL}\big(\pi_\theta(\cdot\mid I)\,\Vert\,\pi_{\text{ref}}(\cdot\mid I)\big)\big]
\end{aligned}
\end{equation}
Intuitively, the first term pushes $\pi_\theta$ toward $q_G$ within the group (listwise), and the KL term limits drift from a safe reference policy $\pi_{\text{ref}}$; both are group-local, matching comparability in our formulation (Sec.~3).

\subsection{Bounded Degradation via KL Control}
We formalize the ``cannot degrade beyond a bounded amount'' claim under bounded KL.

\begin{proposition*}[\textcolor{black}{Bounded change under GRPO} (main text Proposition~2)]
\label{thm:grpo-appendix}
Assume the reward model is rank-consistent (Proposition~\ref{thm:rank-consistency-appendix}) and $h_G\in[0,1]$.
Let $\Delta_{\mathrm{KL}}=\mathbb{E}_I\big[\mathrm{KL}\big(\pi_\theta(\cdot\mid I)\,\|\,\pi_{\text{ref}}(\cdot\mid I)\big)\big]$.
Then the expected within-group humor satisfies
\[
\mathbb{E}_{(I,G)}\Big[\mathbb{E}_{c\sim \pi_\theta(\cdot\mid I)}\,h_G\big((I,c)\big)\Big]
\;\ge\;
\mathbb{E}_{(I,G)}\Big[\mathbb{E}_{c\sim \pi_{\text{ref}}(\cdot\mid I)}\,h_G\big((I,c)\big)\Big]
\;-\;\sqrt{\tfrac{1}{2}\,\Delta_{\mathrm{KL}}}.
\]
Consequently, if GRPO enforces $\Delta_{\mathrm{KL}}\le \tau$ (by choosing $\beta$ or an explicit trust region), the expected humor cannot drop by more than $\sqrt{\tau/2}$. \textcolor{black}{A rank-consistent $q_G$ encourages higher-$h_G$ captions, while the KL constraint makes any possible degradation explicit and bounded.}
\end{proposition*}

\paragraph{Proof.}
For any fixed $(I,G)$, Pinsker’s inequality gives
\begin{equation}
\begin{aligned}
\big\|\pi_\theta(\cdot\mid I) &- \pi_{\text{ref}}(\cdot\mid I)\big\|_{\mathrm{TV}} \\
&\le \sqrt{\tfrac{1}{2}\,\mathrm{KL}\big(\pi_\theta(\cdot\mid I) \,\Vert\, \pi_{\text{ref}}(\cdot\mid I)\big)}
\end{aligned}
\end{equation}
Since $h_G\in[0,1]$, by the variational characterization of total variation for bounded functions,
\begin{equation}
\begin{aligned}
\Big| \mathbb{E}_{\pi_\theta}[h_G] - \mathbb{E}_{\pi_{\text{ref}}}[h_G] \Big| 
&\le \big\| \pi_\theta - \pi_{\text{ref}} \big\|_{\mathrm{TV}} \\
&\le \sqrt{\tfrac{1}{2} \, \mathrm{KL}(\pi_\theta \Vert \pi_{\text{ref}})}
\end{aligned}
\end{equation}
Averaging over $(I,G)$ yields the stated bound.
During GRPO, the cross-entropy term $-\sum q_G\log\pi_\theta$ (with rank-consistent $q_G$) increases mass on higher-$h_G$ captions within the group, while the KL term keeps the deviation controlled. Thus expected humor cannot deteriorate beyond the Pinsker bound and, in practice, improves as the listwise alignment progresses.
\hfill$\square$

\subsection{Discussion: Why Listwise \texorpdfstring{$q_G$}{q\_G} Matters}
Because $q_G$ aggregates pairwise signals into a coherent group distribution consistent with $h_G$’s ordering, the CE term directly performs a proximal step toward the better subset of captions \emph{without} inventing any cross-group scale.
This matches our problem scope and the guarantees in Sec.~4.2–4.3 of the main text.

\section{EBC Aggregation}
\label{app:reward_ebc}

\paragraph{Definition (Expected Borda Count).}
Given a group \(G\) and a finite candidate set \(\mathcal{S}_G=\{m_1,\dots,m_n\}\) with pairwise preference probabilities \(\widehat{p}^{\,G}_{ij}=\Pr[m_i\succ m_j]\), the Expected Borda Count of item \(m_i\) is
\[
\mathrm{EBC}_G(m_i)
\;=\;
\sum_{\substack{j=1\\ j\neq i}}^{n} \widehat{p}^{\,G}_{ij}.
\]
Ties or missing edges are handled by omitting terms (equivalently, treating \(\widehat{p}^{\,G}_{ij}\) as undefined); in evaluation we normalize by the number of available opponents for \(m_i\).

\paragraph{Basic properties.}
(i) If all \(\widehat{p}^{\,G}_{ij}\in\{0,1\}\), EBC reduces to the classical Borda score (number of wins).  
(ii) If there exists a latent utility \(u:\mathcal{S}_G\to\mathbb{R}\) such that \(\widehat{p}^{\,G}_{ij}=\sigma(u(m_i)-u(m_j))\) with strictly increasing \(\sigma\), then sorting by EBC is order-equivalent to sorting by \(\sum_{j\neq i}\sigma(u(m_i)-u(m_j))\); in particular, when gaps are consistent across pairs, the EBC order agrees with the order of \(u\).  
(iii) Under independent edge noise and bounded missingness, the variance of \(\mathrm{EBC}_G(m_i)\) decreases with the number of observed pairs, making the aggregate rank more stable than any single comparison.

\paragraph{Listwise normalization (optional).}
For downstream use, one may define a soft distribution over \(\mathcal{S}_G\) via a temperature \(T>0\):
\[
q_G(m_i) \;=\; \frac{\exp\big(\mathrm{EBC}_G(m_i)/T\big)}{\sum_{k=1}^{n}\exp\big(\mathrm{EBC}_G(m_k)/T\big)},
\]
which converts EBC scores into smooth listwise targets for within-group reweighting. This preserves the group-local nature of the signal and avoids inventing cross-group scales.

\paragraph{Notes on implementation.}
We compute \(\widehat{p}^{\,G}_{ij}\) only within groups and on the (usually small) candidate sets used for evaluation or optimization. When the pair graph is sparse, we keep EBC unbiased by summing over observed opponents and normalizing by their count; when required, we add small-degree regularization to avoid over-confident ranks for items with very few edges. The pseudocode is shown in the Algorithm~\ref{alg:ebc-matrix}

\begin{algorithm*}[t]
\caption{\textcolor{black}{Expected Borda Count (matrix form)}}
\label{alg:ebc-matrix}
\begin{algorithmic}[1]
\Require Candidate set $\mathcal{S}_G=\{m_1,\dots,m_n\}$; pairwise estimates $\Prij=\Pr[m_i\succ m_j]$ (may be undefined); temperature $T>0$ (optional); small-degree regularizer $\alpha\ge 0$ (optional).
\Ensure EBC scores $\EBC_G(m_i)$ for all $m_i$; optionally soft listwise $q_G(m_i)$.

\State Initialize $EBC[i]\gets 0$ and $deg[i]\gets 0$ for all $i\in\{1,\dots,n\}$.
\For{$i=1$ to $n$}
  \For{$j=1$ to $n$}
    \If{$i=j$} \State \textbf{continue} \EndIf
    \If{$\Prij$ is defined} \Comment{omit ties/missing edges}
       \State $EBC[i]\gets EBC[i]+\Prij$
       \State $deg[i]\gets deg[i]+1$
    \EndIf
  \EndFor
\EndFor

\For{$i=1$ to $n$} \Comment{unbiased normalization under sparsity}
  \If{$deg[i] > 0$}
    \State $EBC[i]\gets \dfrac{EBC[i]+\alpha}{deg[i]+\alpha}$ \Comment{$\alpha$ prevents overconfidence at tiny degree}
  \Else
    \State $EBC[i]\gets 0$
  \EndIf
\EndFor

\If{$T$ is provided}
   \State compute $q[i]\gets \exp(EBC[i]/T)$ for all $i$, then $Z\gets \sum_k q[k]$
   \State \textbf{return} $(EBC[i],\, q[i]\leftarrow q[i]/Z)$ for all $i$
\Else
   \State \textbf{return} $EBC[i]$ for all $i$
\EndIf
\end{algorithmic}
\end{algorithm*}

\section{Pair-wise Dataset Construction}
\label{app:reward_data}
Our reward model is trained on \emph{pairwise} comparisons. Intuitively, pairs whose ordering is both \emph{reliably correct} and \emph{increasingly challenging} drive the model toward more consistent ranks. We therefore construct a curriculum of \textbf{five difficulty tiers}, guaranteeing correct orderings while progressively raising difficulty (from trivial mismatches to near-ties within the same template/scene). To span both trivial and subtle distinctions, we sample pairs across all tiers and upweight harder tiers during training, yielding a supervision signal that is confident yet discriminative:
\begin{enumerate}
    \item Wrong Text Meme ($\star$): This is the most straightforward case, where the original text is replaced with unrelated content, completely removing the humor. This type of meme is easy for the model to classify as "non-humorous" and acts as a baseline.

    \item Wrong Location Meme ($\star\star$): A slightly more complex case involves shifting the position of the text in the image. While the metaphor may still exist, the humor diminishes due to the misplacement of text. The model must learn that small positional changes can significantly impact the meme's humor, reflecting a higher degree of difficulty.

    \item Boring Meme ($\star\star$): Here, the meme is altered to include a more mundane or less engaging version of the original text. This teaches the model to distinguish between "humorous" and "boring" versions of the same meme. Although the content still aligns with the original, the humor is less impactful, presenting a challenge for classification.

    \item Detailed Boring Meme ($\star\star\star$): This is a more nuanced case where only one or two words are changed to make the meme less funny. Despite the minimal changes, the meme's humor is significantly affected. The classifier must be able to identify these subtle shifts in humor, marking this as a more difficult classification task.

    \item Generated Meme ($\star\sim\star\star\star$): Finally, memes generated by the fine-tuned VLM represent the highest difficulty level. These memes are intended to be as humorous as the original meme, requiring the classifier to discern fine-grained differences in humor between the generated meme and the original. This provides the model with an opportunity to improve its sensitivity to subtle differences in meme quality.
\end{enumerate}

\begin{figure*}[t]
	\centerline{\includegraphics[width=0.95\linewidth]{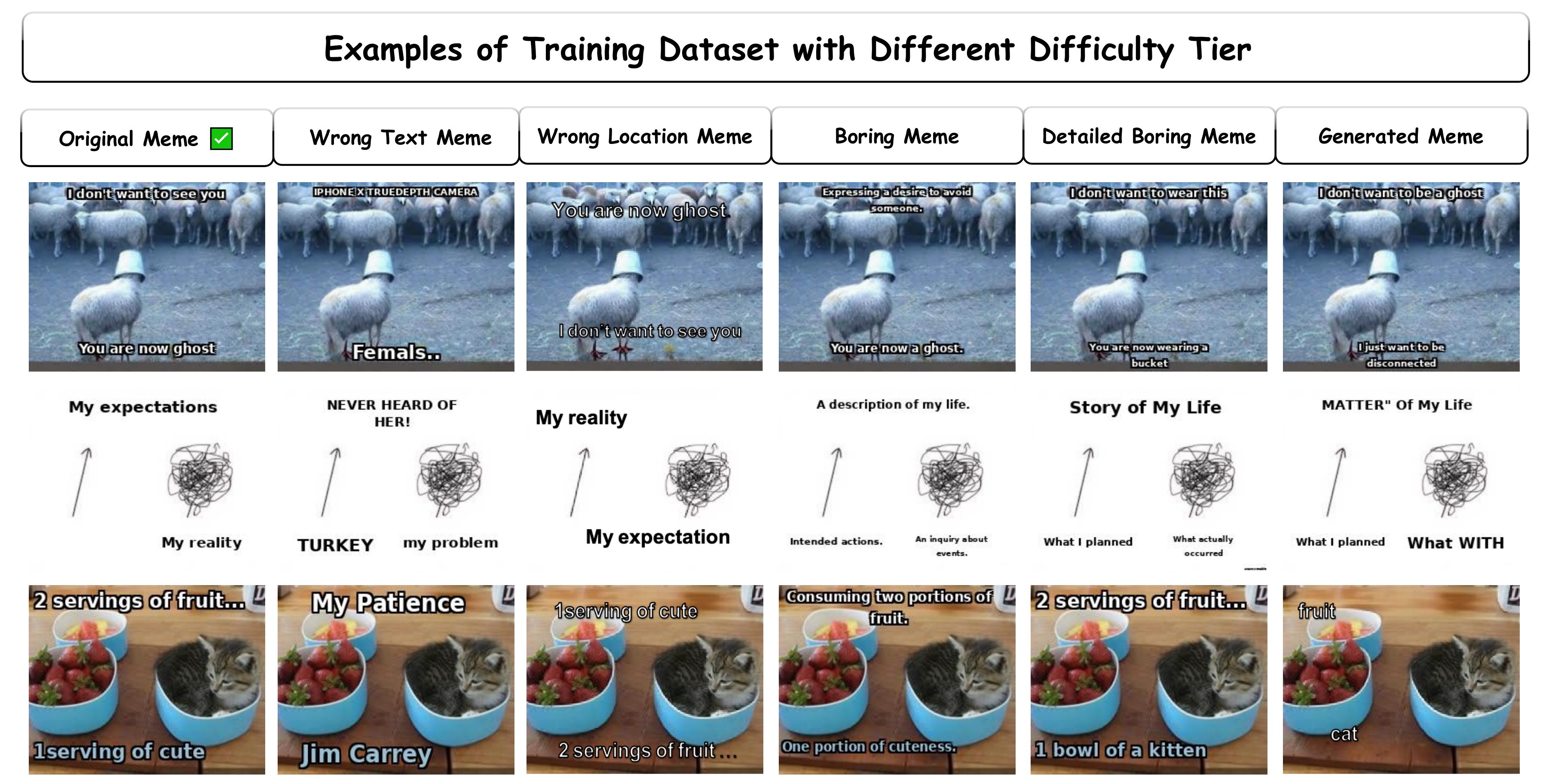}}
\caption{Examples of training datasets with different difficulty tier}
\label{fig:pairwise_dataset}
\end{figure*}

The example of the training dataset is shown in Figure~\ref{fig:pairwise_dataset}. By constructing a dataset with pairs of memes across these varying levels of humor, we enable the classifier to learn not only to distinguish obviously bad memes from good ones but also to understand the nuanced differences that make one meme more humorous than another. This rich dataset plays a crucial role in refining the reward model, allowing it to classify memes based on subtle human preferences.

We stratify training so each mini‐batch contains an equal number from each tier.

\section{Auxiliary Rewards for Reasoning-Path Optimization}
\label{app:aux_rewards}

While optimizing toward the group-wise reward induced by the reward model (Sec.~\ref{sec:reward}) is theoretically sufficient to improve the quality of generated memes, the reinforcement learning stage does not directly supervise the internal reasoning path $r=(r_{\text{tmpl}},r_{\text{scene}})$ because the primary feedback is attached to the realized meme $(I,c)$. 
To explicitly shape the quality of the reasoning process itself, we introduce two auxiliary rewards that operate on $r$: a \emph{format reward} and a \emph{content reward}. 

\subsection{Format Reward}
\label{app:format_reward}
The format reward enforces structural completeness of the CoT to ensure that essential modules appear and are well-formed. 
It is computed by deterministic string/structure matching without using LLM-as-judge. 
Concretely, given a sampled reasoning trace $r$ for $(I,U)$, we check:
\begin{enumerate}
    \item \textbf{Presence of mandatory sections} (e.g., a \texttt{Comprehensive Description} section that summarizes visual content and intended template-level intent).
    \item \textbf{Two-stage structure} (explicit evidence of both template-level intent and context-level grounding consistent with Sec.~\ref{sec:cot}).
    \item \textbf{Text-on-Meme box formatting} (the \texttt{Text on the Meme} block must specify box–text mappings consistent with the bounding boxes $B=\{b_i\}$ so that rendered text $T=\{t_i\}$ aligns with $B$).
\end{enumerate}
The format reward $R_{\mathrm{fmt}}(r)\in[0,1]$ is the normalized sum of satisfied checks. 
It shapes $r$ toward complete and renderable reasoning without requiring any subjective judgment.

\subsection{Content Reward
\label{app:content_reward}}
The content reward evaluates the informativeness and plausibility of the CoT content via an \emph{LLM-as-judge}. 
\textcolor{black}{We prompt an evaluation model to score $r$ along four interpretable dimensions (e.g., visual grounding, template intent clarity, metaphorical mapping, and punchline coherence), each with discrete bands (e.g., 1/4/7 points with band descriptors such as “no object description / coarse description / detailed object attributes”). 
Scores are summed and rescaled to $R_{\mathrm{cnt}}(r)\in[0,1]$.}

\textcolor{black}{However, prior work rarely verifies whether a vision–language reward signal is monotonic with respect to intended semantic quality. To ensure that our RL optimization is grounded on a reliable content metric, we systematically compare several candidate reward options.}

\textcolor{black}{We construct five groups of captions whose intended content quality is controlled at target levels
$\{0,10,30,40,50\}$.
These groups are obtained by prompting two widely used multimodal LLMs—\textit{Qwen2.5-VL-7B} and \textit{Keye-VL-8B}—to generate CoT rationales and captions under progressively stronger quality constraints. For each generated caption, we compute content reward using three scoring strategies: \textit{Qwen2.5-VL-7B} scorer, \textit{Keye-VL-8B} scorer, and \textit{Qwen2.5-VL-7B} output logits, where the final score is calculated from the normalization of logits of each score token.}

\textcolor{black}{In Table~\ref{tab:content_reward}, we then examine whether the final reward values increase along with the intended quality levels. Across both data sources (Qwen-generated and Keye-generated), \textit{Keye-VL-8B} as the judge exhibits the clearest monotonic trend: scores grow consistently as target quality increases. In contrast, \textit{Qwen2.5-VL-7B} scoring shows weaker correlation, and normalized logits are noticeably noisy. Notably, \textit{Keye-VL-8B} remains stable even when scoring content generated by another model, suggesting better cross-distribution generalization.}

\textcolor{black}{These results indicate that \textit{Keye-VL-8B} provides the most rank-consistent, semantically aligned content reward, and we therefore adopt it as the content reward model in our RL stage.}

\textbf{Consistency and Reliability.} We further validate the effectiveness of combining HUMOR-RM with this newly designed content reward for RL training. The degree of rank consistency inherently depends on the chosen base model: semantically stronger and better-aligned backbones (e.g., Keye-VL) yield more reliable results, whereas weaker models align less steadily. For a detailed selection of evaluation models and the consistency tests of this content reward, see Appendix~\ref{app:VLM-evaluator-details} and the validity tests in Appendix~\ref{app:training_setting}.

\begin{table}[t]
\centering
\small
\setlength{\tabcolsep}{5pt}
\caption{\textcolor{black}{Content reward evaluation across different target quality levels. Keye-VL as judge exhibits the clearest monotonic trend, and is therefore adopted as our content reward model in RL.}}
\begin{tabular}{p{1.5cm}p{0.5cm}ccccp{0.6cm}}
\toprule
\textbf{Source Model} & \textbf{Target Score} & \textbf{Qwen-VL} & \textbf{Keye-VL} & \textbf{Logits (Qwen)} \\
\midrule
\multirow{5}{*}{Qwen-VL} 
& 0  & 0.692 & 0.209 & 0.498 \\
& 0.2 & 0.642 & 0.229 & 0.452 \\
& 0.6 & 0.678 & 0.096 & 0.383 \\
& 0.8 & 0.522 & 0.400 & 0.368 \\
& 1 & 0.630 & 0.478 & 0.387 \\
\midrule
\multirow{5}{*}{Keye-VL} 
& 0  & 0.508 & 0.280 & 0.390 \\
& 0.2 & 0.695 & 0.653 & 0.379 \\
& 0.6 & 0.672 & 0.731 & 0.412 \\
& 0.8 & 0.674 & 0.769 & 0.388 \\
& 1 & 0.538 & 0.691 & 0.462 \\
\bottomrule
\end{tabular}
\label{tab:content_reward}
\end{table}

\subsection{Integration with GRPO}
\label{app:integration_grpo}
Let $s_{\mathrm{RM}}(m)$ denote the reward-model score that induces the group-wise ranking distribution $q_G$ via EBC in Sec.~\ref{sec:reward}. 
For a candidate set $\mathcal{S}_G=\{m_k=(I,c_k)\}$ with associated reasoning traces $\{r_k\}$, we construct an \emph{augmented} group-wise target $\tilde q_G$ by combining the primary signal with auxiliary rewards on $r_k$:
\begin{equation}
\label{eq:augmented_q}
\begin{aligned}
\tilde q_G(m_k) &\propto \exp\Big(\tfrac{1}{\tau}\big[s_{\mathrm{RM}}(m_k) + \lambda_{\mathrm{fmt}} R_{\mathrm{fmt}}(r_k) + \lambda_{\mathrm{cnt}} R_{\mathrm{cnt}}(r_k)\big]\Big) \\
\text{s.t.} \quad &\sum_{m_k\in\mathcal{S}_G}\tilde q_G(m_k)=1
\end{aligned}
\end{equation}
where $\tau>0$ is a temperature and $\lambda_{\mathrm{fmt}},\lambda_{\mathrm{cnt}}\!\ge\!0$ are weights. 
The GRPO objective in Eq.~\eqref{eq:grpo-loss} is then used with $q_G$ replaced by $\tilde q_G$.

\begin{remark}[Isotonic shaping and theoretical guarantees]
\label{rem:isotonic}
If $(\lambda_{\mathrm{fmt}},\lambda_{\mathrm{cnt}})$ are chosen such that Eq.~\ref{eq:augmented_q} is an \emph{isotonic} transformation of the reward-model ranking (i.e., it does not invert the order implied by $s_{\mathrm{RM}}$ except to break ties among near-equal items), then the rank consistency guarantees stemming from Proposition~\ref{prop:rank-consistency} are preserved in expectation.
Moreover, the KL-bounded improvement in Proposition~\ref{prop:grpo-bound}
continues to hold because the proof relies on boundedness of $h_G$ and a KL constraint, both unaffected by auxiliary shaping.
In practice we set $\lambda_{\mathrm{fmt}},\lambda_{\mathrm{cnt}}$ small and use them primarily as tie-breakers and regularizers over $r$, which empirically reduces variance and accelerates convergence without altering the main ordering.
\end{remark}

\section{\textcolor{black}{Dataset Statistics and Analysis}}
\label{app:dataset_stats}

\textcolor{black}
{In this section, we provide a comprehensive analysis covering linguistic features, semantic content, and semantic diversity. These statistics validate that the dataset captures the nuanced, punchy, and diverse nature of internet humor required for training robust VLMs.}

\subsection{\textcolor{black}{Linguistic and Semantic Composition}}
\label{app:dataset_linguistic}

\textcolor{black}{
We first analyze the textual properties of the meme captions to ensure they align with the linguistic conventions of internet culture.}

\begin{figure*}[t]
    \centering
    \begin{subfigure}[b]{0.32\textwidth}
        \centering
        \includegraphics[width=\linewidth]{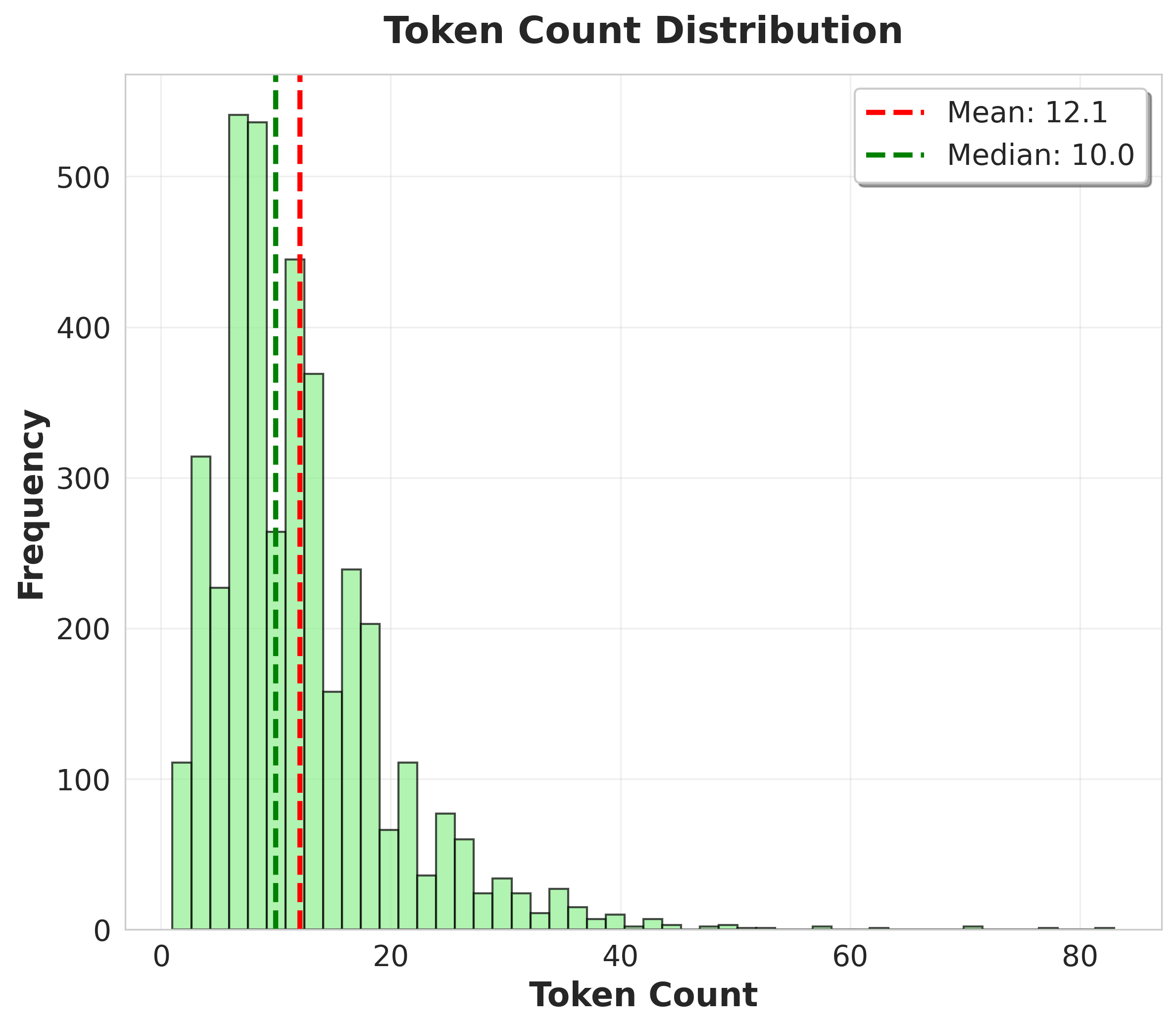}
        \caption{\textcolor{black}{Token count distribution}}
        \label{fig:token_dist}
    \end{subfigure}
    \begin{subfigure}[b]{0.32\textwidth}
        \centering
        \includegraphics[width=\linewidth]{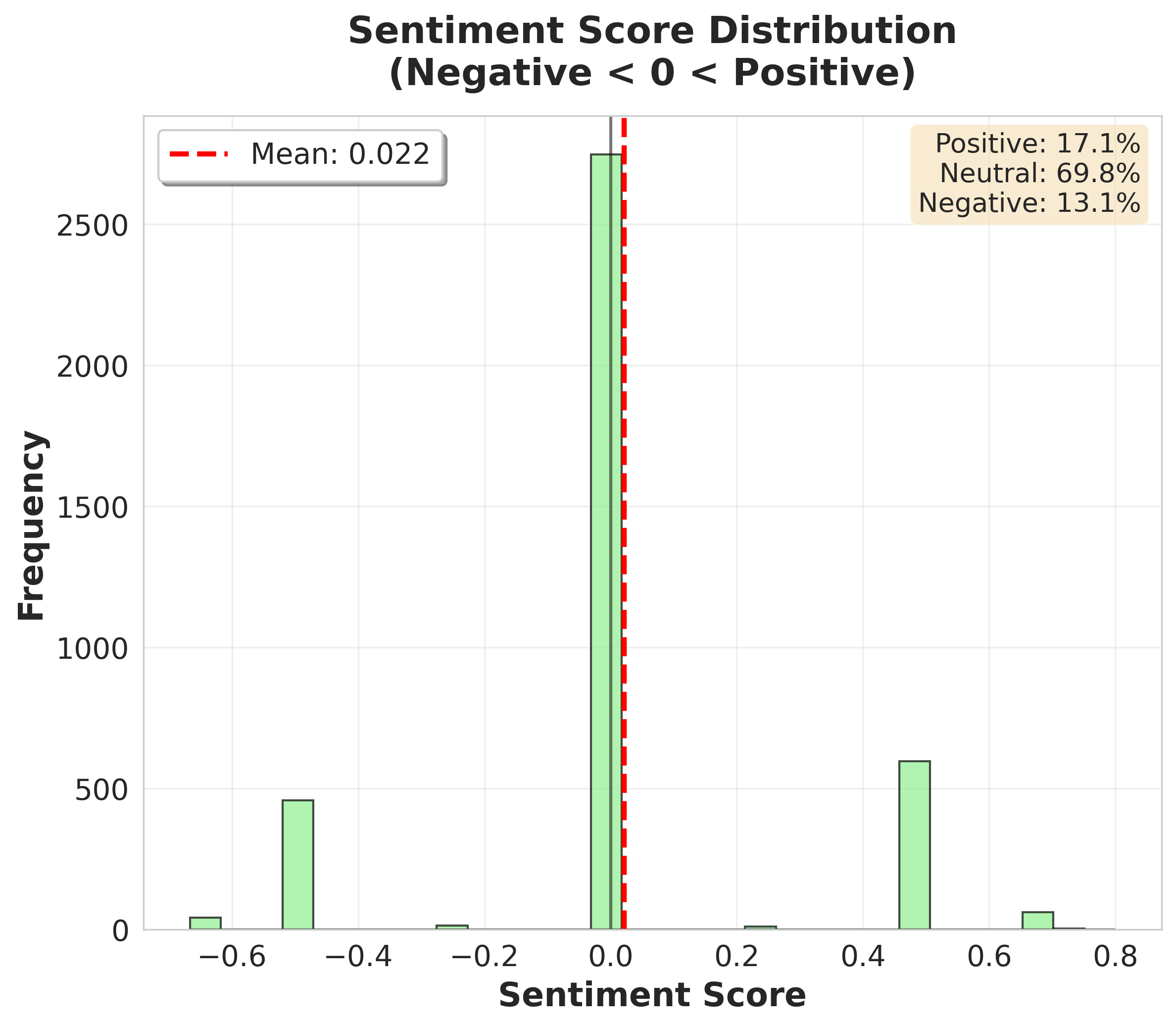}
        \caption{\textcolor{black}{Sentiment score distribution}}
        \label{fig:sentiment_dist}
    \end{subfigure}
    \begin{subfigure}[b]{0.32\textwidth}
        \centering
        \includegraphics[width=\linewidth]{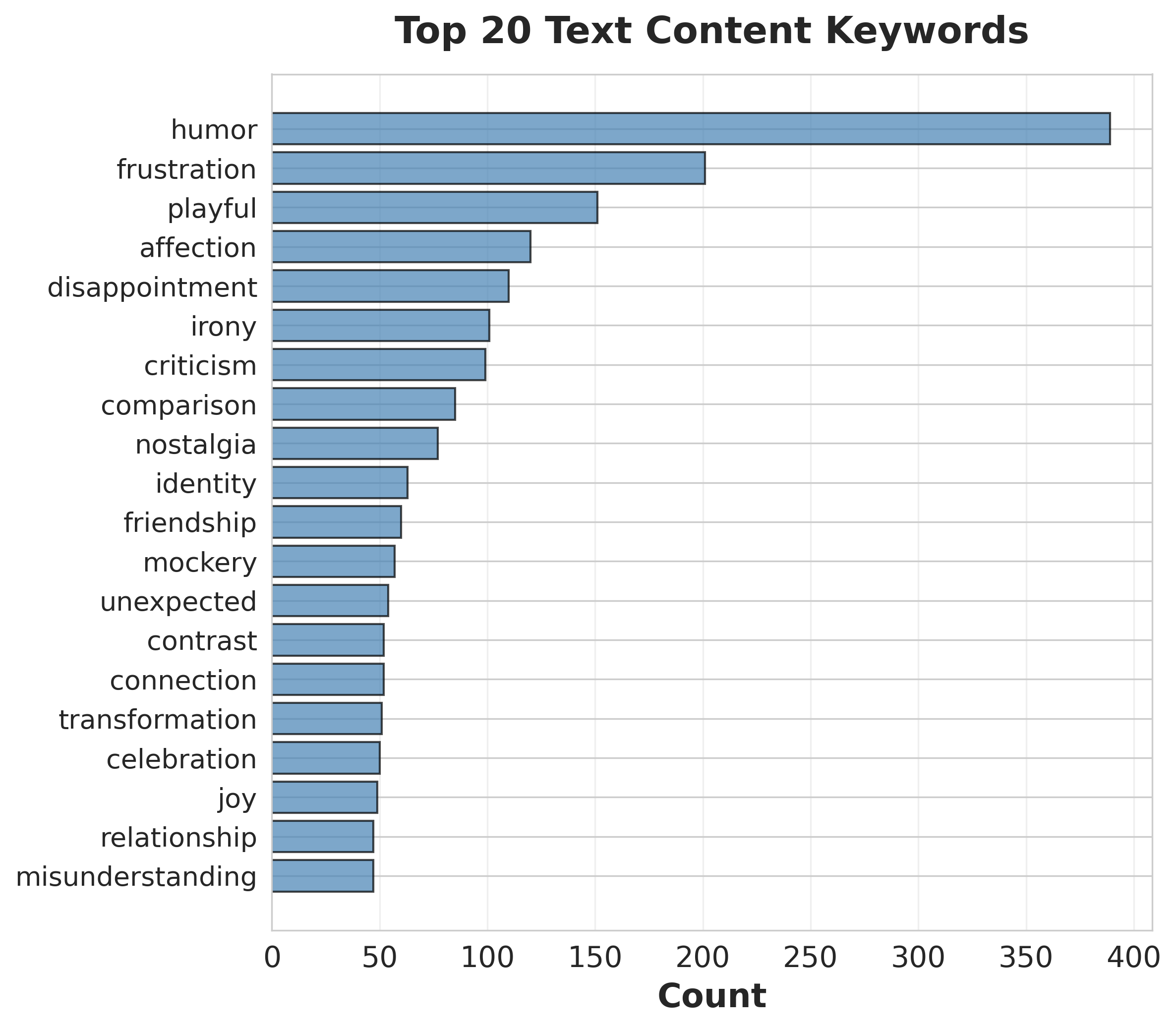}
        \caption{\textcolor{black}{Top 20 keywords}}
        \label{fig:keywords}
    \end{subfigure}
    \caption{\textcolor{black}{Textual properties of meme captions in training dataset}}
    \label{fig:three_subfigs}
\end{figure*}

\textcolor{black}{\textbf{Token Count Distribution:} As illustrated in Figure~\ref{fig:token_dist}, the token count follows a log-normal distribution with a mean of 12.1 and a median of 10.0. This confirms that the dataset consists predominantly of concise, high-impact text, consistent with the ``short and punchy'' nature of memes.}

\textcolor{black}{\textbf{Sentiment Distribution:} The sentiment analysis (Figure~\ref{fig:sentiment_dist}) reveals a dominant Neutral class ($69.8\%$), with balanced Positive ($17.1\%$) and Negative ($13.1\%$) tails. This heavy skew toward neutrality is expected and desirable; meme humor often relies on \textit{deadpan} delivery or irony, where the text itself appears objective or factual, and the humor emerges only through the juxtaposition with visual context.}

\textcolor{black}{\textbf{Semantic Keywords:} The top-30 keyword analysis (Figure~\ref{fig:keywords}) confirms that the dataset is grounded in abstract emotional concepts rather than merely descriptive tags. Dominant keywords such as \textit{Humor}, \textit{Frustration}, \textit{Irony}, and \textit{Disappointment} indicate that the data captures the core thematic essence of relatable internet memes.}

\textcolor{black}{Our data collection is designed to cover diverse in-the-wild meme usage rather than a single hand-crafted template family. The dataset is built from public meme data and contains over 4,000 meme images. In the training setting, we use 3,713 crawled memes, resulting in 3,345 training instances and 368 test instances after filtering and preprocessing. In addition, we analyze 5,000 classic memes to study the relationship between template text positions and caption content. This construction provides broad template and usage variation while keeping the evaluation focused on English-language meme culture.}

\subsection{\textcolor{black}{Semantic Diversity and Rationality of Distance}}
\label{app:dataset_diversity}

\textcolor{black}{A critical quality of a high-quality meme dataset is \textbf{paraphrastic diversity}---the ability to express the same underlying template intent through varied textual realizations. To quantify this, we analyzed the distribution of semantic distances ($1 - \text{Cosine Similarity}$) between captions within the dataset.}

\begin{figure*}[h]
    \centering
    \begin{subfigure}{0.56\textwidth}
        \includegraphics[width=\linewidth]{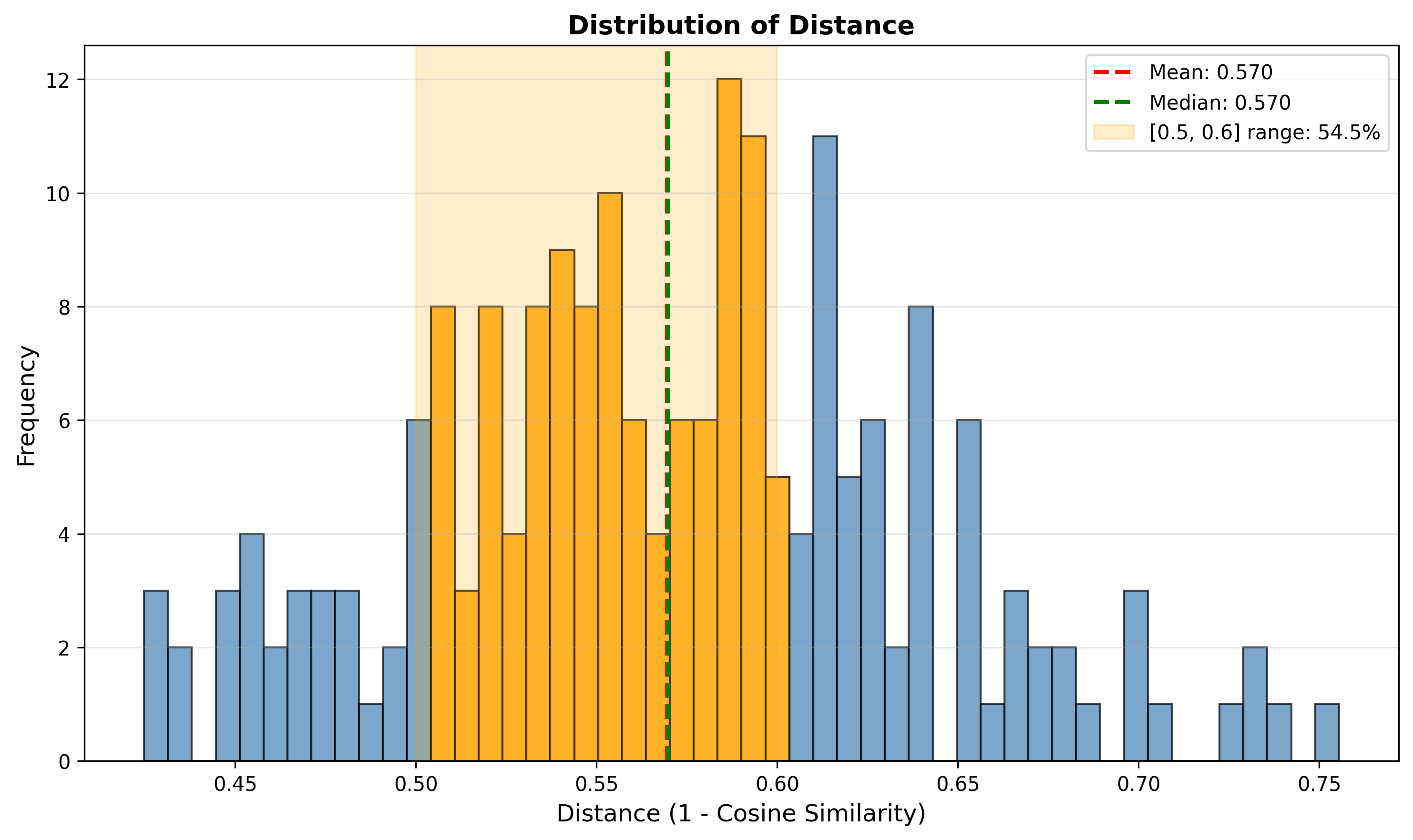}
        \caption{\textcolor{black}{Histogram of Semantic Distances}}
        \label{fig:dist_hist}
    \end{subfigure}
    \hfill
    \begin{subfigure}{0.42\textwidth}
        \includegraphics[width=\linewidth]{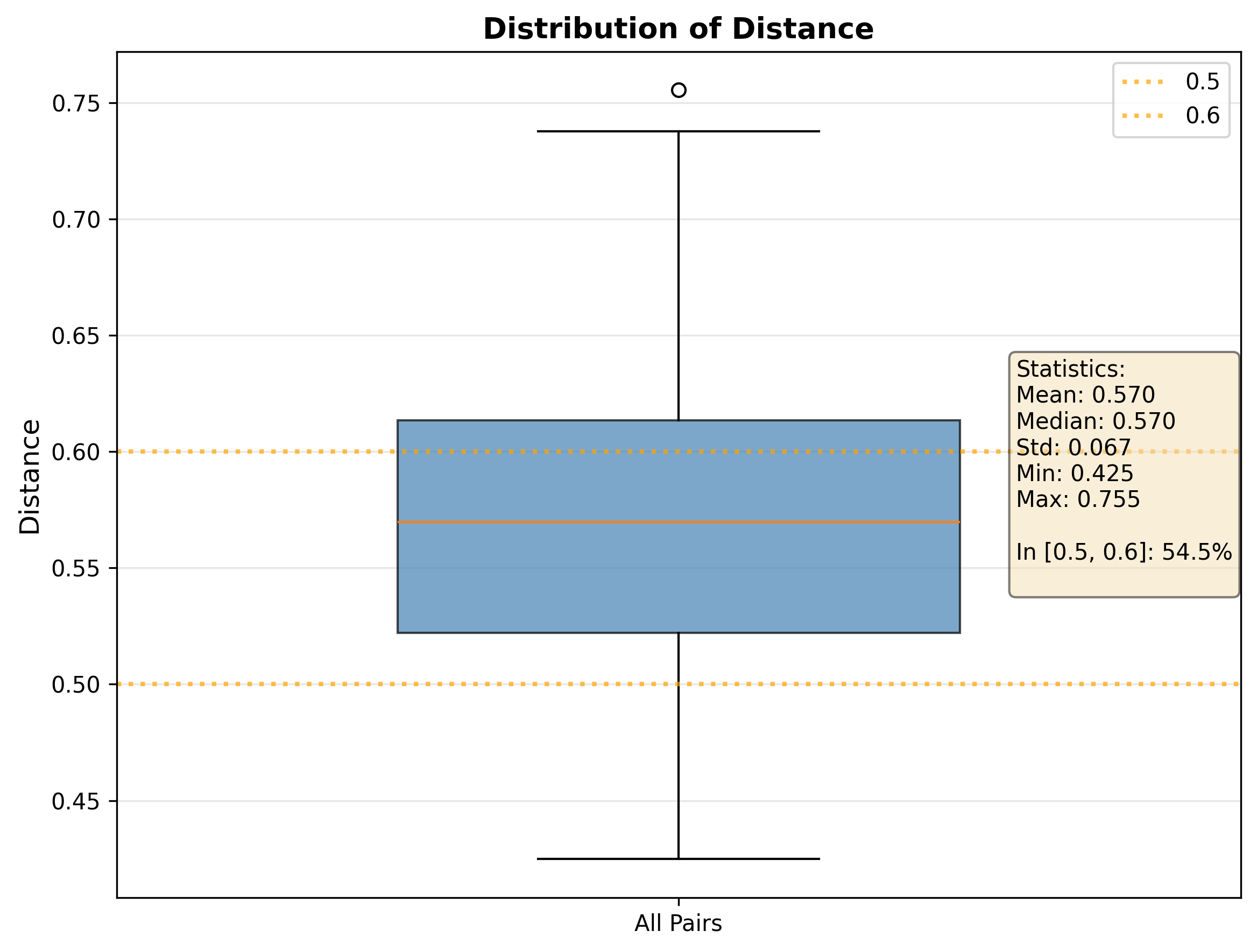}
        \caption{\textcolor{black}{Boxplot of Semantic Distances}}
        \label{fig:dist_box}
    \end{subfigure}
    \caption{\textcolor{black}{\textbf{Analysis of Semantic Diversity.} The distribution of semantic distances (defined as $1 - \text{Cosine Similarity}$) exhibits a mean and median of 0.570. The concentration of data ($54.5\%$) within the $[0.5, 0.6]$ interval indicates a healthy balance: the captions are semantically related enough to share a theme, yet diverse enough to avoid trivial repetition.}}
    \label{fig:diversity_stats}
\end{figure*}

\textcolor{black}{As shown in Figure~\ref{fig:diversity_stats}, the distance metric follows a normal distribution with the following characteristics:
\begin{itemize}
    \item \textbf{Central Tendency:} Both the mean and median are exactly \textbf{0.570}, with a standard deviation of \textbf{0.067}.
    \item \textbf{The ``Goldilocks'' Interval $[0.5, 0.6]$:} A significant majority of the data ($52.5\%$) falls within this specific range.
\end{itemize}}

\textcolor{black}{\textbf{Rationality of the $[0.5, 0.6]$ Range:}}
\textcolor{black}{We argue that this distance distribution is not only reasonable but indicative of a high-quality dataset for open-ended generation:
\begin{enumerate}
    \item \textbf{Avoidance of Mode Collapse ($>0.1$):} A very low distance (e.g., $<0.2$) would imply that the dataset contains largely duplicate or repetitive captions, which leads to overfitting and lack of creativity. Our distribution shows virtually no mass in this region, confirming high lexical diversity.
    \item \textbf{Semantic Coherence ($<0.9$):} A very high distance (e.g., $>0.8$) would suggest random or unrelated text. The maximum distance observed is 0.755, with the vast majority below 0.7, ensuring that the captions remain thematically grounded to the meme templates.
    \item \textbf{Optimal Paraphrasing:} The concentration at $0.57$ represents an optimal middle ground where captions share the same latent humor or intent (lowering distance) but utilize distinct vocabulary and sentence structures (increasing distance). This supports our claim that the dataset facilitates learning robust, generalized humor representations rather than rote memorization.
\end{enumerate}}

\section{Training settings}
\label{app:training_setting}

\subsection{CoT Supervised Fine-tuning settings}
\label{app:cot_training_setting}
The detailed experimental settings for CoT supervision and fine-tuning are shown in Table~\ref{tab:training_setup}.

\begin{table}[h]
\centering
\caption{Training Setup}
\label{tab:training_setup}
\resizebox{\columnwidth}{!}{
\begin{tabular}{ll}
\toprule
Hyperparameter & Value \\
\midrule
Finetuning Stage & \texttt{sft} \\
Finetuning Type & \texttt{lora} \\
LoRA Rank & \texttt{128} \\
LoRA Target & \texttt{all} \\
Per Device Train Batch Size & \texttt{1} \\
Gradient Accumulation Steps & \texttt{8} \\
Learning Rate & \texttt{3.0e-5} \\
Num Train Epochs & \texttt{5.0} \\
LR Scheduler Type & \texttt{cosine} \\
Warmup Ratio & \texttt{0.1} \\
bf16 & \texttt{true} \\
\midrule
Dataset & \texttt{Eimage} \\
Total Dataset Size & 3,713 crawled memes \\
Training Instances & 3,345 \\
Testing Instances & 368 \\
CoT Generation Model & \texttt{doubao-1.5-vision-pro} \\
CoT Variants & \textit{HUMOR-CoT}, \textit{CoT with Single Path}, \\
& \textit{CoT with Self-Improve}, \textit{CoT with Subquestion} \\
\bottomrule
\end{tabular}%
}
\end{table}

\paragraph{Computational Overhead.}
The primary additional cost of our multi-path CoT approach stems from the longer reasoning sequences generated during inference. Empirically, our method generates approximately $2\text{--}3\times$ more total output tokens compared to direct generation baselines. While this increases computational demands during inference, we consider this a worthwhile trade-off given the substantial improvements in humor diversity and quality demonstrated in Table~1. The extended reasoning process enables the model to explore multiple humorous angles and perform more nuanced template-caption alignment, which is critical for generating high-quality memes.

\subsection{Reward Model Training Settings}
\label{app:rm_training_setting}
Our reward model is implemented as a lightweight extension on top of the base vision–language models. Concretely, we take the final hidden embedding of the last transformer layer and append a two-way classification head. This simple design allows the model to learn preference signals while reusing the representational power of the pretrained backbone.  

Based on the dataset constructed in Appendix~\ref{app:reward_data}, we train reward models using the \textit{LLaMA-Factory} framework with the following backbones: \textit{Keye-VL}, \textit{Qwen2.5-VL-7B}, and \textit{Qwen2.5-VL-32B}. All models are fine-tuned with LoRA ($r$ = 8, lora target is all) to reduce memory and computation overhead. We adopt a learning rate of \(1\times 10^{-4}\), with a warmup ratio of 0.1. Each model is trained on a single NVIDIA A800 GPU.

\section{Evaluation settings}
\label{app:evaluation_setting}
\subsection{VLM evaluates experimental setup}
\label{app:VLM evaluates experimental setup}

\paragraph{Evaluation Setup.}
For text generation, we set the temperature to 0 to ensure deterministic outputs.
Objective textual evaluation includes three automatic metrics:
(1) \textbf{Similarity} — cosine similarity between generated and reference captions computed using \textit{\textcolor{black}{bge-base-en-v1.5}}, averaged over all 368 test samples;
(2) \textbf{Distance} — contextual robustness, measured by regenerating 50 samples with mismatched user contexts and averaging textual dissimilarity across three regenerations;
(3) \textbf{Human/AI Discriminability} — binary classification by \textit{Gemini-2.5-pro} judging whether each meme appears human-made, reported as the average ``human rate'' over 368 test memes.

\paragraph{Metric Selection Rationale.}
\textcolor{black}{
Prior to finalizing our protocol, we conducted extensive experiments with standard automated metrics, including BLEU, ROUGE, and BERTScore. However, our analysis reveals that these metrics are ill-suited for evaluating humorous meme captions. When computed on human ground-truth datasets, these scores are exceedingly low (e.g., ROUGE-1: 0.0461, ROUGE-2: 0.0027, BLEU: 0.0025). This indicates minimal lexical overlap even among high-quality human-written punchlines, reflecting two inherent characteristics of memes: captions are often short, fragmentary, and intentionally non-literal. Since humor permits a wide range of valid expressions for the same visual stimulus, overlap-based metrics are unreliable in this domain.
To address these limitations, we designed domain-specific quantitative metrics to form a complementary evaluation strategy.}

\textcolor{black}{
Objective textual evaluation includes three automatic metrics:
(1) \textbf{Similarity} — cosine similarity between generated and reference captions computed using \textit{\textcolor{black}{bge-base-en-v1.5}}, averaged over all 368 test samples;
(2) \textbf{Distance} — contextual robustness, measured by regenerating 50 samples with mismatched user contexts and averaging textual dissimilarity across three regenerations;
(3) \textbf{Human/AI Discriminability} — binary classification by \textit{Gemini-2.5-pro} judging whether each meme appears human-made, reported as the average ``human rate'' over 368 test memes.}

\paragraph{Human Evaluation.}
\textcolor{black}{We conducted our human studies on Prolific as formal research tasks, enforcing strict eligibility filters to ensure cultural and linguistic consistency. Participants were required to be native English speakers with an approval rate $\ge$ 70\% and were compensated according to Prolific’s fair-pay guidelines ($\sim$\pounds9/hr).
Specifically, we recruited 30 qualified annotators for the general meme assessment (\ref{tab:main}) and 100 annotators for the group-wise MaxDiff ranking (\ref{tab:rm}) to obtain robust preference signals.
\textcolor{black}{The available demographic entries for the general assessment indicate that annotators covered a broad adult age range rather than a narrow student population, included both male and female participants, and represented multiple ethnic backgrounds while remaining primarily within an English-speaking evaluation setting. We report these demographics only in aggregate form to preserve participant anonymity.}
For the general assessment, human raters independently evaluated 3–5 memes per method on four dimensions:
(1) \textbf{Humor}, (2) \textbf{Readability}, (3) \textbf{Relevance} to user input, and (4) \textbf{Originality}.
Scores were averaged across raters and samples for each model.}

\paragraph{Multimodal VLM Evaluation.}
All multimodal evaluations used \textit{Gemini-2.5-pro}. Captions were embedded into corresponding bounding boxes, and the model provided meme-level judgments from three perspectives:
(i) human/AI discriminability,
(ii) absolute scoring, and
(iii) relative ranking.

\textbf{VLM Absolute Scoring.}
Each meme was evaluated individually on an absolute 1–5 scale under eight criteria:
1) Punchline Strength: clarity and impact of the joke/twist;
2) Context Robustness: generalizability across social contexts;
3) Humor Effectiveness: quality of humor, sarcasm, or self-mockery;
4) Spread Potential: universal appeal and memorability;
5) Emotional Resonance: capacity to elicit laughter, surprise, or empathy;
6) Cultural Fit \& Relatability: alignment with audience familiarity;
7) Theme Relevance: consistency with keywords and intentions;
8) Image-Caption Relevance: coherence between text and image.
For each meme, the mean of the eight scores was recorded as its overall score.

\textbf{VLM Ranking.}
For relative evaluation, six meme variants sharing the same base image—\textit{HUMOR-CoT}, three CoT variants (Single Path, Self-Improve, Subquestion), \textit{In-the-wild Memes}, and \textit{Text-Free Memes}—were presented together.
\textit{Gemini-2.5-pro} was prompted to rank them jointly under the same eight criteria.
Each group’s results were averaged over 368 test cases to obtain mean rankings.

\subsection{MaxDiff Ordering}
\label{app:max_diff}
Maximum Difference Scaling (MaxDiff), also known as best–worst scaling, is a widely used method in marketing science and preference elicitation~\cite{louviere1991best, louviere2015best}. In a typical MaxDiff task, respondents are repeatedly presented with small subsets of items (e.g., 3–5 candidates) and asked to indicate which option they consider the ``best'' and which the ``worst.'' Compared to traditional rating scales, MaxDiff provides more discriminative and reliable preference estimates because each choice yields two pieces of information: a positive preference for the selected ``best'' item and a negative preference for the ``worst.''

The required number of tasks in MaxDiff depends on the total number of items \(J\) to be evaluated and the subset size \(k\). A common guideline is that each item should appear across multiple choice sets to ensure stable estimation. For example, using balanced incomplete block designs (BIBD), each respondent typically completes between \(\frac{3J}{k}\) and \(\frac{5J}{k}\) choice tasks to achieve acceptable reliability~\cite{orme2010getting}. Thus, the total number of questions can be determined systematically to balance respondent burden and statistical efficiency.

In our study, we adopted a MaxDiff-inspired procedure to construct human preference rankings over memes. Specifically, rather than asking annotators to rate memes on absolute scales, we designed tasks where memes were compared in small groups, and annotators selected the most and least humorous instances. Aggregating these best–worst choices yields a consistent human-validated ranking dataset, which serves as a training and evaluation benchmark for our reward model.

\subsection{Meme Ranking Results}
\label{app:Meme Ranking Result}
\begin{figure*}[!htbp]
    \centering
    \includegraphics[width=1\linewidth]{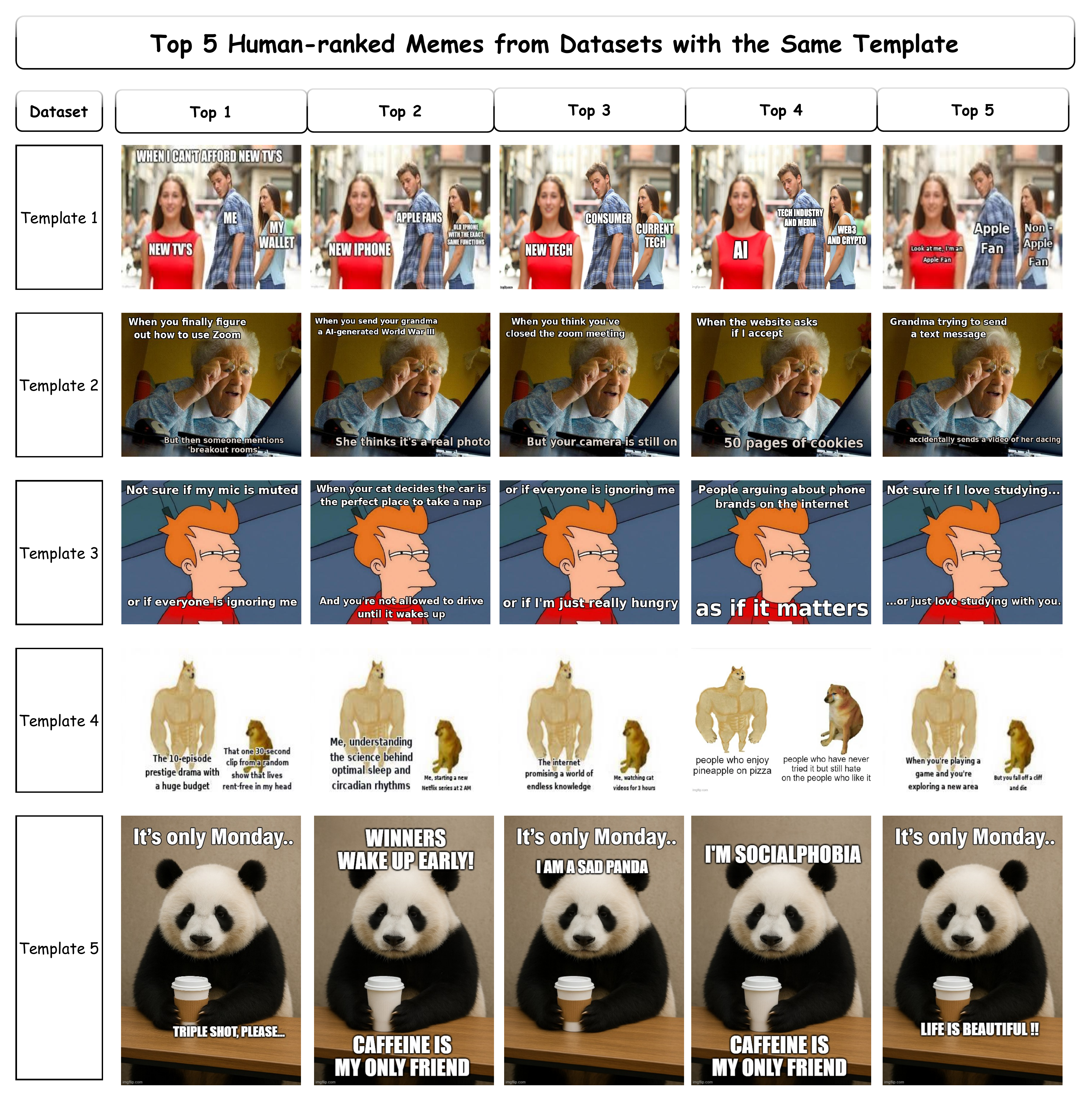}
    \caption{The Top-5 human-ranked memes in the datasets sharing the same templates.}
    \label{fig:rank_dataset}
\end{figure*}

Figure~\ref{fig:rank_dataset} visualizes the top-5 memes extracted from the human-ranked dataset mentioned in Section~\ref{sec:rm_rank_and_rl}.
This evaluation set consists of distinct groups, where each group contains 15 meme variants sharing the same visual template and thematic context.
The rankings were established via the MaxDiff tests described above: in each trial, human annotators were presented with a subset of three memes and asked to identify the ``most preferred'' and ``least preferred'' options. These discrete choices were then aggregated to produce a complete ranking for each template group.

\section{\textcolor{black}{VLM Evaluator Analysis and Human-Alignment Validation}}
\label{app:VLM-evaluator-details}

\textcolor{black}{This appendix provides two complementary analyses regarding the use of \textit{Gemini-2.5-pro}
within our evaluation pipeline. First, we examine Gemini as a \textbf{human-likeness evaluator}
used for computing the \textit{Human Rate} metric (Appendix~\ref{app:evaluator_selection}–\ref{app:human_alignment}).
This part analyzes evaluator selection, statistical reliability, and alignment with ground-truth
labels. Second, we independently study Gemini’s role as a \textbf{group-wise ranking evaluator}
in the relative comparison setting of Fig.~\ref{fig:abs-vs-rank}(b)
(Appendix~\ref{app:groupwise_alignment}). This ranking analysis is separate from Human Rate and
validates that the VLM’s relative judgments meaningfully correlate with human preference structures.}

\subsection{\textcolor{black}{Human Rate Definition}}
\label{app:human_rate_definition}

\textcolor{black}{Human Rate is computed as a binary human/AI discriminability task. Given a complete rendered meme, Gemini-2.5-pro judges whether it is more likely human-generated or AI-generated, and Human Rate reports the proportion of generated memes classified as human-like. This metric should not be interpreted as an absolute humor score. Instead, it measures human-likeness: whether the output avoids common AI-generation artifacts such as generic phrasing, over-polished captions, weak surprise, or poor cultural fit.}

\textcolor{black}{This differs from VLM-based absolute humor scoring, which we find unreliable for subjective cross-template comparison. The validity of Human Rate is supported by our evaluator analysis: among six candidate VLMs, Gemini-2.5-pro obtains the highest AUC for human/AI meme discrimination, and its continuous human-likeness scores significantly correlate with true human/AI labels (Spearman $\rho=0.5932$, $p<0.001$).}

\subsection{\textcolor{black}{Evaluator Selection Analysis}}
\label{app:evaluator_selection}

\begin{figure}[!htbp]
    \centering
    \includegraphics[width=1\linewidth]{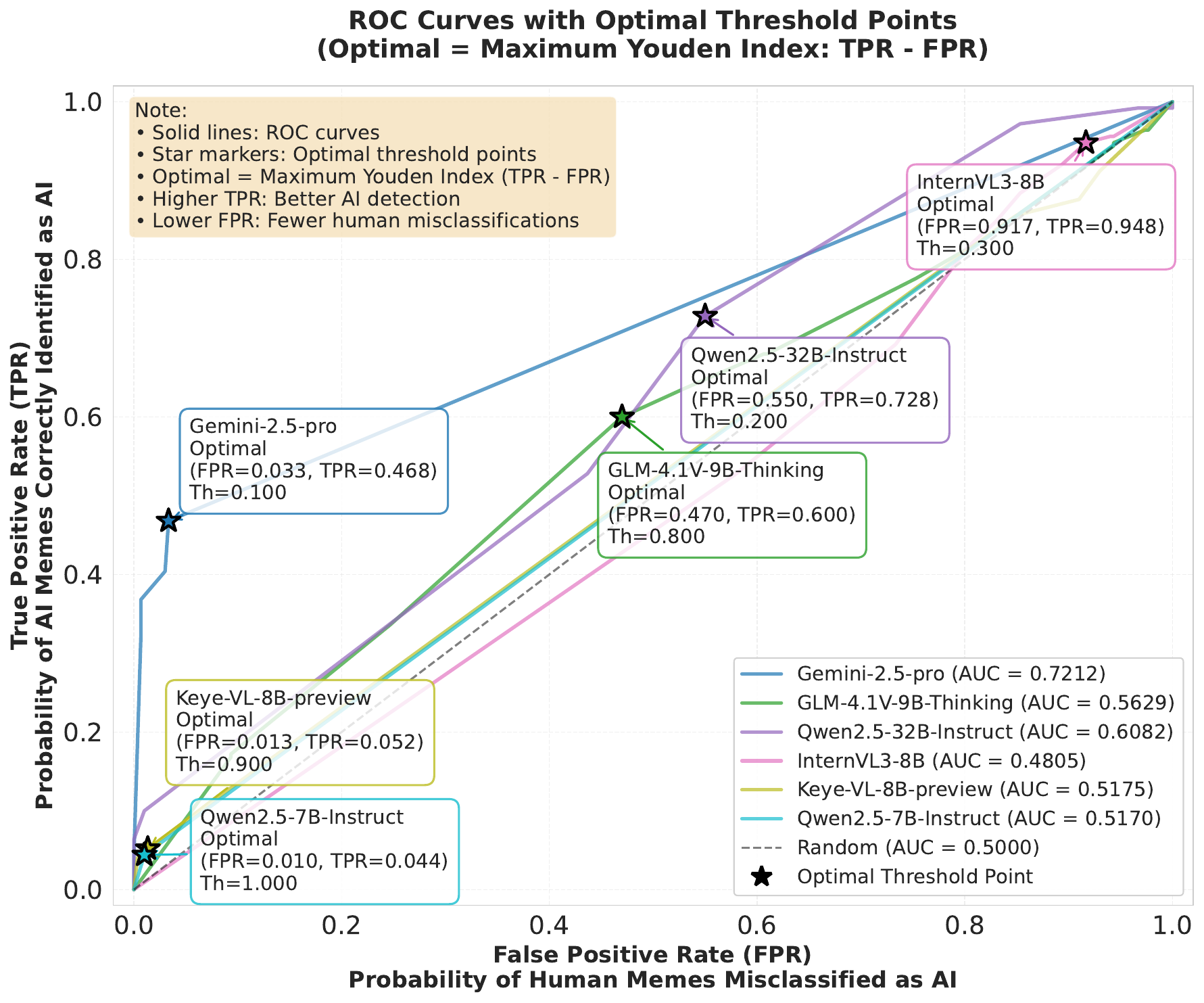}
    \caption{\textcolor{black}{Candidate VLM evaluators' ROC curves for AI vs. human meme classification. Curves plot TPR vs. FPR; star markers denote optimal thresholds (maximizing Youden index) and metrics. AUC (overall discrimination ability) is in the legend.}}
    \vspace{-10pt}
    \label{fig:roc}
\end{figure}

\textcolor{black}{To ensure that Human Rate reflects genuine human-likeness rather than evaluator bias, 
we benchmark six candidate VLMs (Gemini-2.5-pro, Qwen2.5-32B, Qwen2.5-7B, InternVL3-8B,
Keye-VL-8B, and GLM-4.1V-9B) on a held-out set containing 250 AI-generated and 
300 human-created memes. We evaluate each model’s discriminative ability via ROC-AUC 
(Fig.~\ref{fig:roc}) and inspect its error characteristics at the operational threshold.}

\textcolor{black}{Among all candidates, \textbf{Gemini-2.5-pro} demonstrates the most favorable profile:}
\begin{itemize}
\item \textcolor{black}{\textbf{Highest AUC (0.7212)}, substantially outperforming the next-best model
(Qwen2.5-32B: 0.6082), indicating the strongest global separability between AI and human memes.}
\item \textcolor{black}{\textbf{Highest specificity (TNR = 0.97)}, meaning Gemini almost never misclassifies 
genuine human memes as AI. Since Human Rate measures the proportion of outputs judged as 
human-like, low-specificity evaluators (e.g., Qwen2.5-32B, GLM) would systematically penalize
human memes, compressing model differences and making the metric unreliable.}
\end{itemize}

\textcolor{black}{Alternative VLMs exhibit extremely low specificity (TNR = 0.15–0.56). Such evaluators would inaccurately depress Human Rate across all models.}

\textcolor{black}{While Gemini’s sensitivity is moderate (TPR = 0.404), this introduces a \textbf{shared error floor}—
a uniform tendency to classify part of the AI memes as human-like across all systems—which does
not distort \textit{relative} comparisons.}

\textcolor{black}{Overall, Gemini’s combination of extremely high specificity, the highest AUC, and a shared
sensitivity bias makes it the most suitable evaluator for computing Human Rate.}

\subsection{\textcolor{black}{Significance and Reliability Analysis}}
\label{app:significance_reliability}

\textcolor{black}{Although the evaluator introduces a fixed non-zero error rate, this error applies uniformly to all 
evaluated models. Thus, pairwise differences in Human Rate remain reliable as long as they exceed
this shared noise floor.}

\textcolor{black}{To verify this, we conduct a two-proportion $z$-test comparing the rate at which 
\textit{HUMOR-CoT} and the Qwen2.5-7B base model are labeled as human by Gemini-2.5-pro. 
The difference is \textbf{highly significant} ($z = 5.81$, $p < 10^{-8}$), confirming that the 
observed improvement cannot be explained by evaluator variability.}

\textcolor{black}{We further validate stability by re-computing Human Rate across random subsets of the test set, 
where the relative ranking of all compared models remains unchanged. Together, these analyses 
demonstrate that Human Rate provides \textbf{consistent and reproducible} model comparisons.}

\subsection{\textcolor{black}{Human-Alignment Validation}}
\label{app:human_alignment}
\textcolor{black}{To assess how closely Gemini’s human-likeness judgments match human perception, we conduct an 
independent human-labeling study on 30 memes (15 AI-generated, 15 human-generated). After removing 
one ambiguous sample, 29 items remain, each annotated by 22–24 participants.}

\textcolor{black}{\textbf{Human annotator reliability} Inter-annotator agreement is statistically significant but low 
(\textbf{Fleiss' $\kappa = 0.1369$, $p < 0.001$}), reflecting the subjective nature of determining
meme authenticity.}

\textcolor{black}{\textbf{Gemini alignment with true labels.} Gemini’s continuous scores correlate strongly with ground-truth labels 
(\textbf{Spearman $\rho = 0.5932$, $p < 0.001$}). 
Binary consistency varies with threshold: Cohen’s $\kappa$ improves from 0.1944 (threshold 0.5) to 
0.3888 (threshold 0.9), alongside a corresponding increase in accuracy.}

\textcolor{black}{\textbf{Human judgments vs. true labels.} Human judgments show weak negative agreement with ground-truth authenticity (\text{Cohen’s $\kappa = -0.4397$, $p < 0.05$}; Spearman $\rho = -0.4493$, $p < 0.05$), likely due to anthropomorphism and the difficulty of discerning AI-generated memes.}

\textcolor{black}{\textbf{Conclusion.} Gemini aligns with ground-truth labels substantially better than human annotators, and its 
continuous outputs encode meaningful gradients of human-likeness. These findings, combined with 
its high specificity and top AUC, support using Gemini-2.5-pro as a reliable evaluator for 
Human Rate.}

\subsection{\textcolor{black}{Human Alignment of Gemini’s Group-wise Ranking Evaluator}}
\label{app:groupwise_alignment}
\textcolor{black}{To validate the relative ranking results in Fig.~\ref{fig:abs-vs-rank}(b), we perform an 
independent human evaluation that mirrors the same group-wise comparison protocol described in 
Sec.~\ref{sec:vlm_reliability}. For five representative image groups (30 memes total), nine human 
annotators ranked the six meme variants—\textit{HUMOR-CoT}, three CoT baselines, 
\textit{In-the-wild}, and \textit{Text-Free}—under the same criteria rubric used by 
\textit{Gemini-2.5-pro}. Each annotator produced one holistic ranking per group.}

\textcolor{black}{We compute rank correlation between Gemini’s and human rankings. Across the five groups,
Gemini exhibits strong and consistent alignment with human preferences, with a mean 
\textbf{Spearman correlation of 0.7188 $\pm$ 0.2154} and \textbf{Kendall’s $\tau$ of 
0.6320 $\pm$ 0.2269}. }

\textcolor{black}{These results confirm that the group-wise ranking in Fig.~\ref{fig:abs-vs-rank}(b) captures 
preference structures also expressed by humans, providing quantitative evidence that the 
relative VLM evaluation is meaningful and not an artifact of evaluator noise. Combined with the 
preceding analyses, this supports the reliability of the Gemini-based ranking methodology used 
throughout our evaluation.}

\section{\textcolor{black}{Supplementary Experiment Results}}
\label{app:supplementary_exp}
\begin{figure*}[!htbp]
    \centering
    \includegraphics[width=1\linewidth]{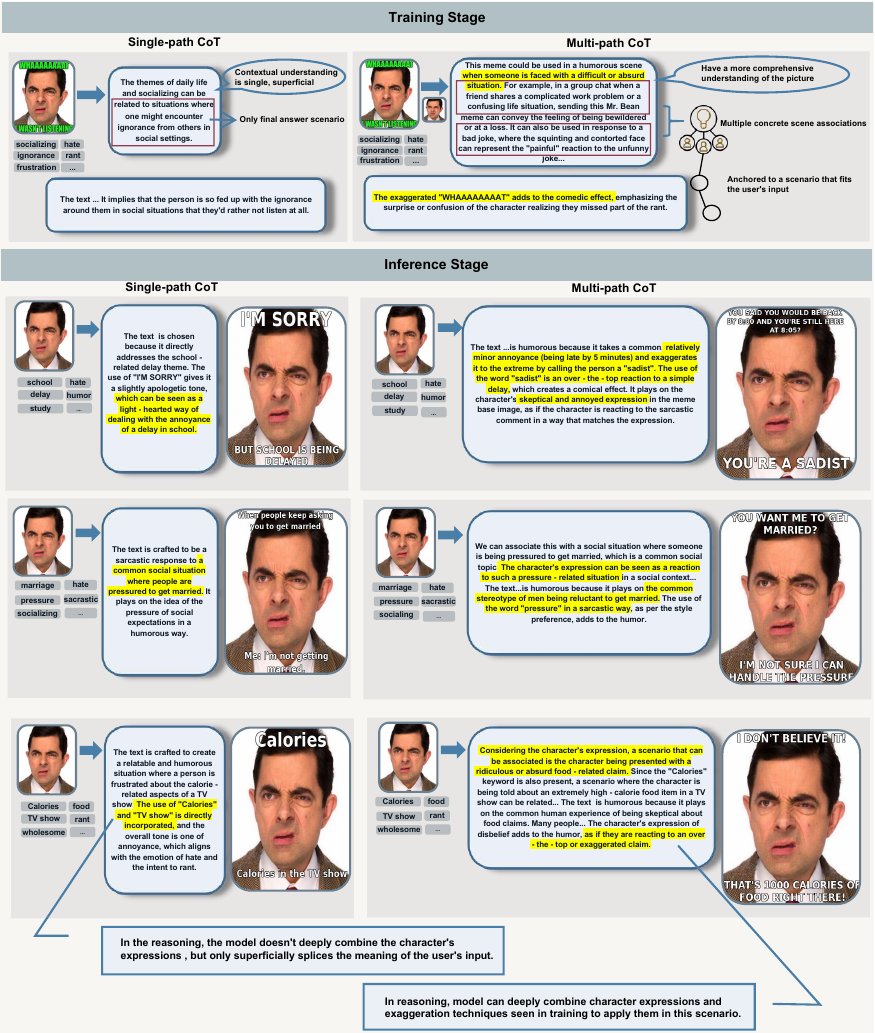}
    \caption{\textcolor{black}{Case study comparing Single-path and Multi-path Hierarchical CoT supervision in meme generation using the same Mr. Bean image.
The single-path model reproduces the ground-truth reasoning chain, yielding literal and less contextual humor.
The multi-path model, trained with multi-scenario associative reasoning, demonstrates improved contextual understanding and humorous transferability, producing text that creatively matches new user intents.}}
    \vspace{-10pt}
    \label{fig:single-path}
\end{figure*}
\subsection{\textcolor{black}{VLM Classification Result}}
\label{app:VLM Classification Result}
\textcolor{black}{To further examine whether our generated meme texts faithfully reflect the intended semantics of user input, we perform a reclassification experiment using a strong vision-language model (VLM) as an external evaluator.
Specifically, we take the captions generated by each model and feed them into the same VLM classifier that was trained to recognize four major semantic axes: \textit{emotion}, \textit{intention}, \textit{theme}, and \textit{style}.
The classifier outputs predicted labels for each axis, which are compared to the original user-specified categories to compute reclassification accuracy.}

\textcolor{black}{Table~\ref{tab:reclassify} summarizes the results.
HUMOR-CoT achieves the highest accuracy across all dimensions, surpassing both the Qwen2.5-7B-Instruct and the larger Qwen2.5-32B-Instruct baselines.
This indicates that our hierarchical CoT fine-tuning not only improves humor expressivity but also enhances the faithfulness of generated texts to user intent.
In particular, the improvement over the 32B model suggests that structured reasoning contributes more effectively to semantic alignment than mere parameter scaling.}

\begin{table*}[t]
\centering
\small
\setlength{\tabcolsep}{5pt}
\caption{\textcolor{black}{Compare qwen2.5-7B, 32B, with the results of reclassification of the model we trained on qwen2.5-7B for the sentiment, intent, theme, and style of user input}}
\begin{tabular}{llccccccc}
\toprule[1.5pt]
\multicolumn{2}{l}{\multirow{2}{*}{\centering Model}} 
& \multicolumn{4}{c}{RE-classification Accuracy$\uparrow$} \\
\cmidrule(lr){3-6}
&
& Emotion & Intention & Theme & Style \\ 
\midrule[1pt]
& Qwen2.5-7B-Instruct  & 0.420  & 0.515  &0.551  &0.521    \\
& Qwen2.5-32B-Instruct & 0.571  & 0.611  & \textbf{0.616} &0.603 \\
& HUMOR-CoT    & \textbf{0.597}  & \textbf{0.641}   & 0.600   &  \textbf{0.639}  \\

\bottomrule
\end{tabular}
\label{tab:reclassify}
\end{table*}

\subsection{\textcolor{black}{Cross-Domain Pilot on Persona-Conditioned Visual Captioning}}
\label{app:cross_domain_pilot}

\textcolor{black}{To examine whether the HUMOR-style reasoning and group-wise selection procedure can be instantiated beyond meme generation, we conducted a small cross-domain pilot under the rebuttal timeline rather than a full transfer benchmark. We use persona-conditioned visual captioning based on Personality-Captions~\citep{shuster2019personalitycaptions}. Each group is defined by the same recovered image and the same personality trait, i.e., \(G=(\text{image}, \text{personality})\). This setting matches the formulation in Sec.~\ref{sec:formulation}: outputs are open-ended captions, quality depends on subjective criteria such as engagement, persona consistency, visual grounding, and coherence, and pairwise comparisons are naturally meaningful within the same image-persona group.}

\begin{figure*}[!htbp]
    \centering
    \includegraphics[width=0.95\linewidth]{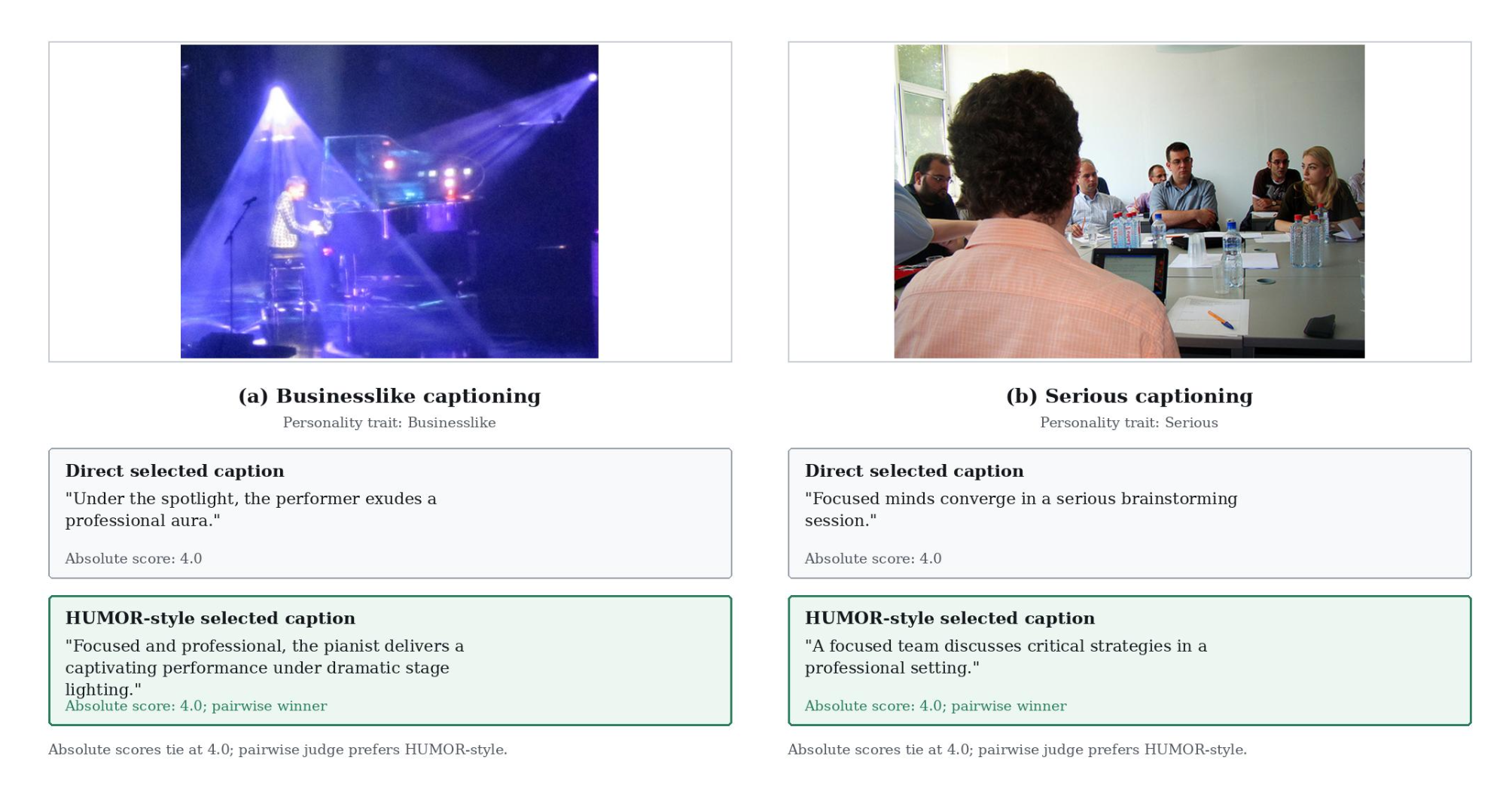}
    \caption{\textcolor{black}{Qualitative examples from the persona-conditioned visual captioning pilot. For the same image-persona group, HUMOR-style multi-path reasoning explores different caption strategies before local selection.}}
    \vspace{-8pt}
    \label{fig:pc_pilot}
\end{figure*}

\textcolor{black}{We sampled 20 recovered image-persona groups and used the same Qwen2.5-VL-7B-Instruct model for both methods. The direct baseline generates four captions directly. The HUMOR-style procedure first explores visual evidence, persona intent, and multiple plausible caption angles, and then generates four candidates under the same image-persona condition. Figure~\ref{fig:pc_pilot} provides representative qualitative examples.}

\begin{table}[!htbp]
\centering
\small
\setlength{\tabcolsep}{4pt}
\caption{\textcolor{black}{Candidate diversity in the cross-domain pilot. distinct-1 measures word-level diversity, distinct-2 measures local bigram-level diversity, and average pairwise sentence distance measures semantic diversity in sentence-embedding space.}}
\begin{tabular}{lcccc}
\toprule
\textcolor{black}{Method} & \textcolor{black}{\#Groups} & \textcolor{black}{distinct-1} & \textcolor{black}{distinct-2} & \textcolor{black}{Avg. sentence dist.} \\
\midrule
\textcolor{black}{Direct} & \textcolor{black}{20} & \textcolor{black}{0.734} & \textcolor{black}{0.960} & \textcolor{black}{0.507} \\
\textcolor{black}{HUMOR-style multi-path} & \textcolor{black}{20} & \textcolor{black}{0.738} & \textcolor{black}{0.942} & \textcolor{black}{0.563} \\
\bottomrule
\end{tabular}
\label{tab:pc_diversity}
\end{table}

\textcolor{black}{As shown in Table~\ref{tab:pc_diversity}, HUMOR-style generation slightly improves word-level diversity (0.734 \(\rightarrow\) 0.738) and more clearly improves semantic diversity (0.507 \(\rightarrow\) 0.563, about \(+11.0\%\)). distinct-2 slightly decreases (0.960 \(\rightarrow\) 0.942), so we do not claim uniform lexical diversity improvement. Instead, the result indicates that multi-path reasoning primarily broadens the space of high-level caption strategies: candidates move farther apart in sentence meaning, persona interpretation, and communicative angle under the same image-persona condition, rather than simply swapping local bigrams or increasing surface phrase variation.}

\textcolor{black}{We also compare independent absolute rubric scoring with group-local pairwise judging. In absolute scoring, each candidate is scored independently on engagement, persona consistency, visual grounding, coherence, and overall quality. This scoring is not discriminative in our pilot: all 160 candidate captions receive 4.0 on every dimension. In contrast, group-local pairwise judging provides more informative preference signals.}

\begin{table}[!htbp]
\centering
\small
\setlength{\tabcolsep}{5pt}
\caption{\textcolor{black}{Group-local pairwise preference results comparing the HUMOR-style selected caption against the direct selected caption.}}
\begin{tabular}{lcccc}
\toprule
\textcolor{black}{Judger} & \textcolor{black}{\#Groups} & \textcolor{black}{HUMOR wins} & \textcolor{black}{Ties} & \textcolor{black}{HUMOR losses} \\
\midrule
\textcolor{black}{Qwen2.5-VL-7B-Instruct} & \textcolor{black}{20} & \textcolor{black}{14} & \textcolor{black}{0} & \textcolor{black}{6} \\
\textcolor{black}{MiniMax-M3} & \textcolor{black}{20} & \textcolor{black}{12} & \textcolor{black}{3} & \textcolor{black}{5} \\
\textcolor{black}{Step-3.7-Flash} & \textcolor{black}{20} & \textcolor{black}{10} & \textcolor{black}{9} & \textcolor{black}{1} \\
\bottomrule
\end{tabular}
\label{tab:pc_pairwise}
\end{table}

\textcolor{black}{Across three judgers, HUMOR-style selected captions are preferred more often than direct selected captions: Qwen gives 14 wins vs. 6 losses, MiniMax-M3 gives 12 wins vs. 5 losses with 3 ties, and Step-3.7 gives 10 wins vs. 1 loss with 9 ties. The Step-3.7 result is weaker because many cases are tied, and this pilot uses only the reasoning and selection component rather than full HUMOR reward-model training. We therefore treat the result as directional evidence rather than a full cross-domain generalization claim. Still, the pilot supports two core points of the framework: absolute scoring can be poorly discriminative for subjective multimodal generation, while local pairwise comparison within semantically coherent groups can provide useful preference signals beyond the meme domain.}


\section{Limitations and Societal Impact}
\label{app:limitations}

This section provides a self-critical analysis of the HUMOR framework, its underlying model HUMOR-RM, and the potential broader impacts of our work.

\subsection{Inference Latency and Computational Cost}
As detailed in Section~\ref{sec:cot} and Appendix~\ref{app:training_setting}, the hierarchical multi-path CoT mechanism requires the model to generate structured reasoning trajectories before yielding the final punchline. While this "thinking" process is essential for the logical leap required by humor, it incurs a computational overhead of approximately $2\times$ to $3\times$ more tokens than standard VLM captioning. This latency may hinder real-time deployment on mobile devices. Future work could explore Early Exit strategies to bypass reasoning for simpler visual stimuli.

\subsection{Cultural and Linguistic Specificity}
Humor is intrinsically tied to cultural context, slang, and shared history. Our training data~\citep{xu2024generating} is primarily composed of English-language internet memes. \textcolor{black}{Accordingly, our empirical results should be read as evidence within an English-speaking meme setting, not as a universal measurement of humor across all cultures. At the same time, the HUMOR framework itself does not rely on language- or culture-specific assumptions: it requires coherent groups and reliable within-group preferences. Extending the data collection, reward validation, and human evaluation to multilingual and multicultural meme communities is an important direction for future work.}

\subsection{Failure Modes}
\label{app:failure_modes}
Despite the anchoring mechanism described in Proposition~\ref{prop:cot-lb}, qualitative reviews (see Appendix~\ref{app:failure_case} for failure cases) indicate that the model occasionally fails to generate a humorous incongruity when the input image is highly complex or visually cluttered. In these cases, the CoT tends to default to literal description rather than finding a creative twist. This suggests that the model’s "creativity" is still bounded by the diversity of its pre-trained visual-linguistic associations.

\subsection{Broader Societal Impact and Safety}
The ability to automate humor generation carries risks. Memes are frequently used for political satire, which can border on misinformation or harassment. Although we utilized Gemini-2.5-pro as a risk judge (Section~\ref{sec:rm_rank_and_rl}) to filter out hate speech and sensitive content, adversarial prompts could potentially bypass these safeguards. We emphasize that HUMOR should be used as a tool for creative assistance rather than an autonomous content generator in sensitive domains.

\section{\textcolor{black}{Additional Generated Samples and Case Studies}}

\label{app:additional_cases}

\textcolor{black}{This appendix presents additional qualitative results related to the experiments in the main paper, including generated samples, risk cases, and failure analyses. These examples complement our understanding of HUMOR-CoT’s behavior under different conditions. All samples are produced under the same test protocol and prompting settings as Fig.~\ref{fig:abs-vs-rank}(b). Full evaluation prompts and system settings are provided in Appendix~\ref{app:VLM evaluates experimental setup}.}

\subsection{\textcolor{black}{Generated Samples Across CoT Strategies}}

\label{app:Generated Samples}

\begin{figure*}[!htbp]
    \centering
    \includegraphics[width=1\linewidth]{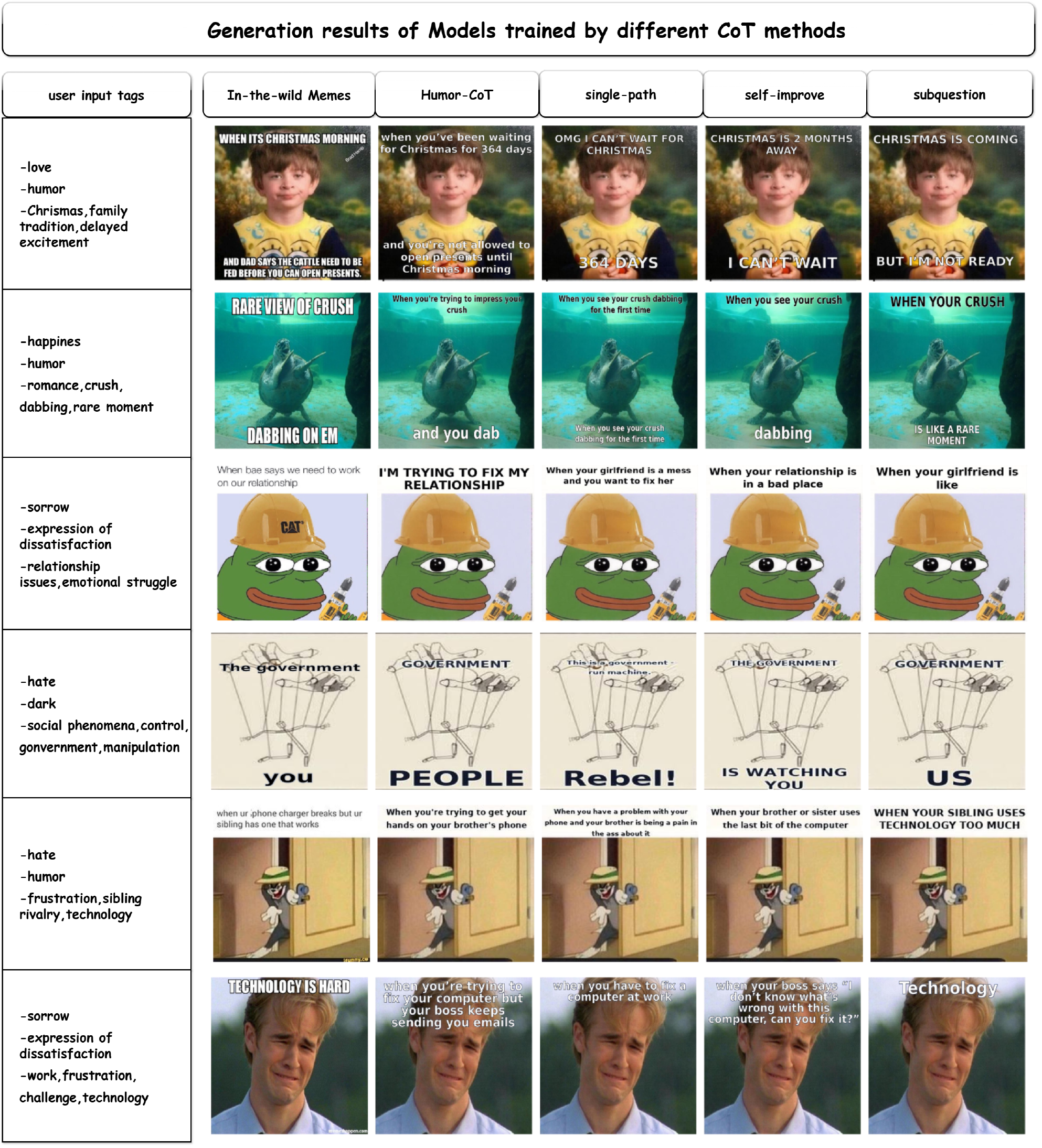}
    \caption{\textcolor{black}{Generation results of models trained with different CoT strategies.}}
    \vspace{-10pt}
    \label{fig:generation_results_comparison}
\end{figure*}

\textcolor{black}{To further analyze how different Chain-of-Thought (CoT) strategies affect meme generation, Figure~\ref{fig:generation_results_comparison} visualizes representative outputs. Each row corresponds to a user-intent cluster (e.g., romance, Christmas, family tradition, delayed surprise). Each column shows one of the five outputs: \textit{In-the-wild} (human-created reference), \textit{HUMOR-CoT}, and three alternative CoT approaches (\textit{Single-path}, \textit{Self-improve}, \textit{Subquestion}).}

\textcolor{black}{From the comparison, HUMOR-CoT more accurately captures user-implied emotions and contextual nuances, better preserves alignment between visual content and textual humor, and overall produces more coherent and structurally sound meme captions than competing strategies.}

\subsection{\textcolor{black}{Generalization to Unseen Templates}}

\label{app:unseen_templates}

\textcolor{black}{To verify HUMOR-CoT’s ability to generalize to template formats entirely absent from training, we constructed 20 unseen meme templates and evaluated them using the same group-wise ranking protocol as Fig.~\ref{fig:unseen_rank}. For each template, we jointly ranked outputs from HUMOR-CoT and five representative VLM generators using Gemini-2.5-pro as a comparative evaluator.}

\begin{figure*}[!htbp]
    \centering
    \includegraphics[width=1\linewidth]{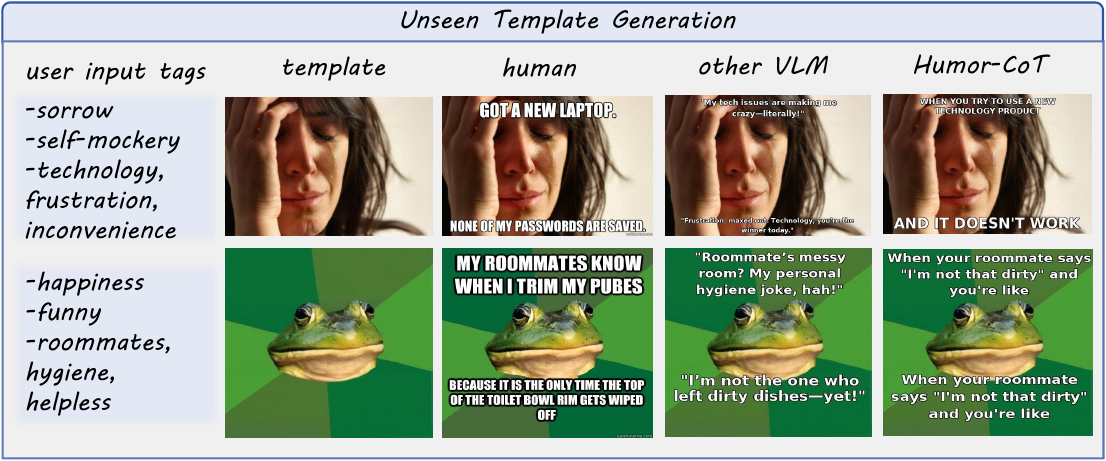}
    \caption{\textcolor{black}{\textbf{Unseen Template Generation.} HUMOR-CoT generalizes well to templates entirely absent from training, producing humorous and contextually aligned captions.}}
    \vspace{-10pt}
    \label{fig:unseen_case}
\end{figure*}

\textcolor{black}{As shown in Fig.~\ref{fig:unseen_case}, HUMOR-CoT consistently produces captions that remain semantically fitting, visually grounded, and logically humorous even under unseen template structures. This finding echoes the quantitative results in Fig.~\ref{fig:unseen_rank}, suggesting that HUMOR-CoT generalizes across template styles rather than overfitting to specific training formats or humor patterns.}

\subsection{\textcolor{black}{Retrieval-Augmented Adaptation to Emerging Meme Trends}}
\label{app:retrieval_emerging_memes}

\textcolor{black}{Unseen visual templates are only one aspect of meme generalization; another challenge is evolving cultural meaning. To clarify the role of retrieval, we conduct a small qualitative case study in which recent meme-trend descriptions are retrieved and inserted into the prompt as additional context. This setup does not compete with HUMOR. Instead, retrieval supplies external and time-sensitive context, while HUMOR-style reasoning uses that context to interpret the visual template, explore plausible communicative intents, and generate captions under the local preference structure.}

\textcolor{black}{We examine two recent trend descriptions. The first is ``Somebody get these beggars out of here,'' used as a reaction phrase when someone is perceived as begging, freeloading, or asking for something unreasonable. The second is the ``Reasons to get a bob'' trend, where the humor comes from a numbered list that repeats the same answer, e.g., a deliberately deadpan list of reasons that all reduce to one word. When these retrieved descriptions are added to the prompt, HUMOR-CoT can adapt the captioning strategy to the new usage: the model grounds the retrieved meme convention in the given image, identifies the intended social stance, and produces punchlines that follow the emerging trend format.}

\textcolor{black}{This case study supports our scope clarification: if the goal is the strongest deployed meme-generation system, retrieval, template mining, and trend tracking are valuable additions. HUMOR focuses on the model-side capability and is orthogonal to these additions. Retrieved examples can be supplied as context, while HUMOR improves the VLM's reasoning, multi-context exploration, and group-local preference-aligned generation.}

\subsection{\textcolor{black}{Risk Case Identification}}

\label{app:risk_case}

\textcolor{black}{To further ensure the safety of HUMOR-CoT’s meme generation process, we conducted a detailed analysis of high-risk cases under the same evaluation protocol used in Fig.~\ref{fig:abs-vs-rank}(b). Certain user-provided tags—particularly those involving political ideology, wartime historical figures, religious identity, gender topics, or dark cultural references—can inadvertently lead the model toward unsafe or controversial outputs.}

\textcolor{black}{To address this, we incorporate Gemini-2.5-pro as a dedicated risk auditor, applied to every generated meme before presenting the final output. The auditor evaluates political sensitivity, cultural offensiveness, and overall dissemination risk, and blocks unsafe generations. Notably, only 3.3\% of the memes generated by our model are classified as high-risk.}

\begin{figure*}[!htbp]
    \centering
    \includegraphics[width=1\linewidth]{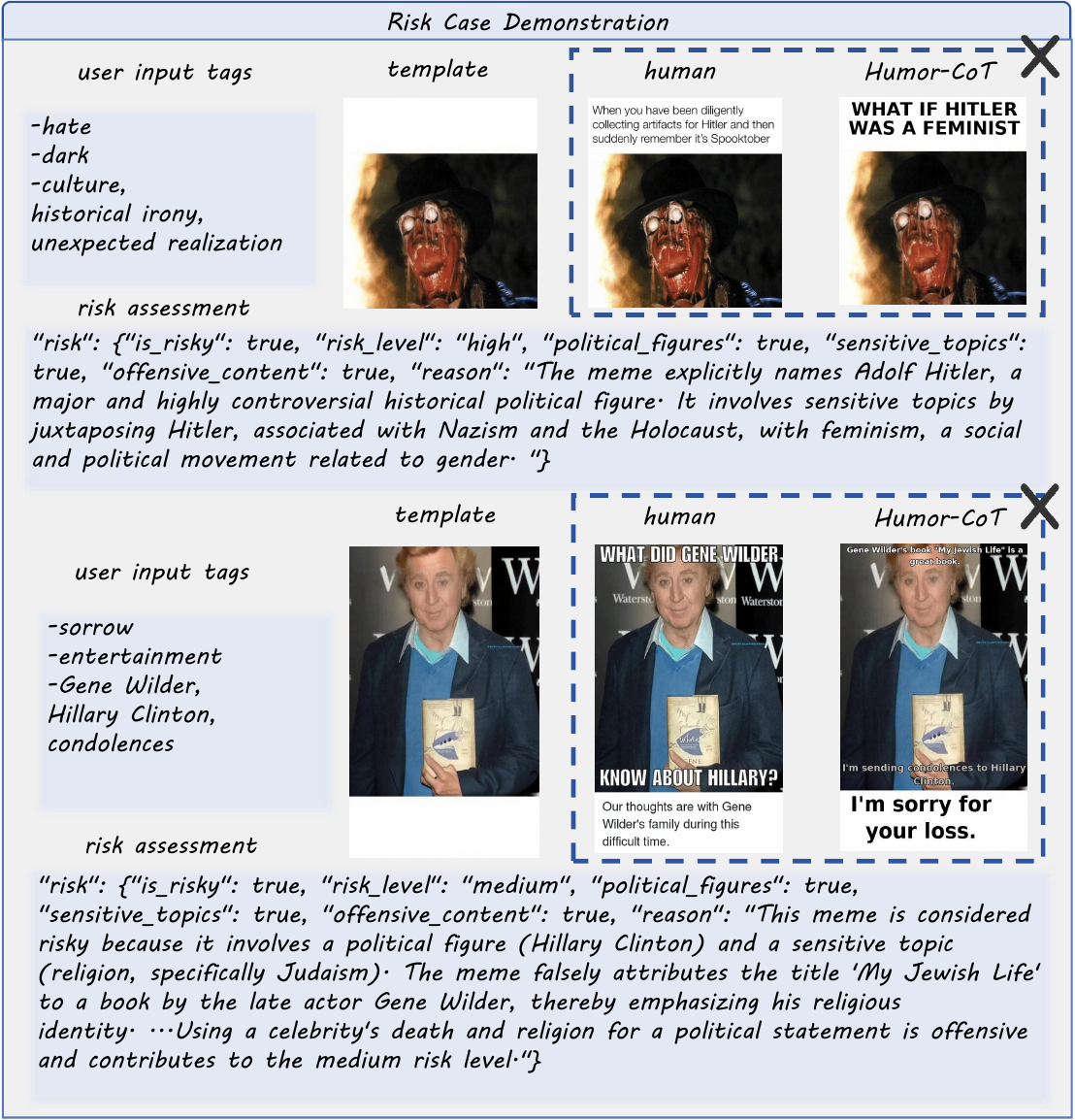}
    \caption{\textcolor{black}{\textbf{Risk example identification.} Gemini-2.5-pro effectively flags politically sensitive or socially harmful meme generations.}}
    \vspace{-10pt}
    \label{fig:risk_case}
\end{figure*}

\textcolor{black}{Figure~\ref{fig:risk_case} presents two representative high-risk examples:}
\textcolor{black}{Case 1:
The user provides tags such as hate, dark, historical irony, etc. The generated meme juxtaposes a highly controversial political figure with a modern gender movement. This combination is flagged as high-risk because it may trivialize historical atrocities or imply derogatory gender-based associations.}
\textcolor{black}{Case 2:
Input tags include sorrow, entertainment, Gene Wilder, Hillary Clinton, etc. The generated meme incorrectly pairs an actor’s photo with a political figure and a religiously sensitive theme, resulting in a medium-risk classification due to offensive misattribution and implied ideological framing.}

\textcolor{black}{These examples highlight how subtle combinations of template imagery and user-provided tags can cause risk escalation. The auditor effectively surfaces such vulnerabilities and prevents them from influencing model outputs. Future work may incorporate training-time safety constraints so that generation itself avoids drifting into politically sensitive or harmful narratives.}

\subsection{\textcolor{black}{Failure Case Analysis}}

\label{app:failure_case}
\textcolor{black}{Under the same generation protocol as Fig.~\ref{fig:abs-vs-rank}(b), we also observe several consistent failure modes of HUMOR-CoT. These failures are not safety-related but rather stem from limitations in humor construction, scene preservation, and compositional reasoning.}

\begin{figure*}[!htbp]
    \centering
    \includegraphics[width=1\linewidth]{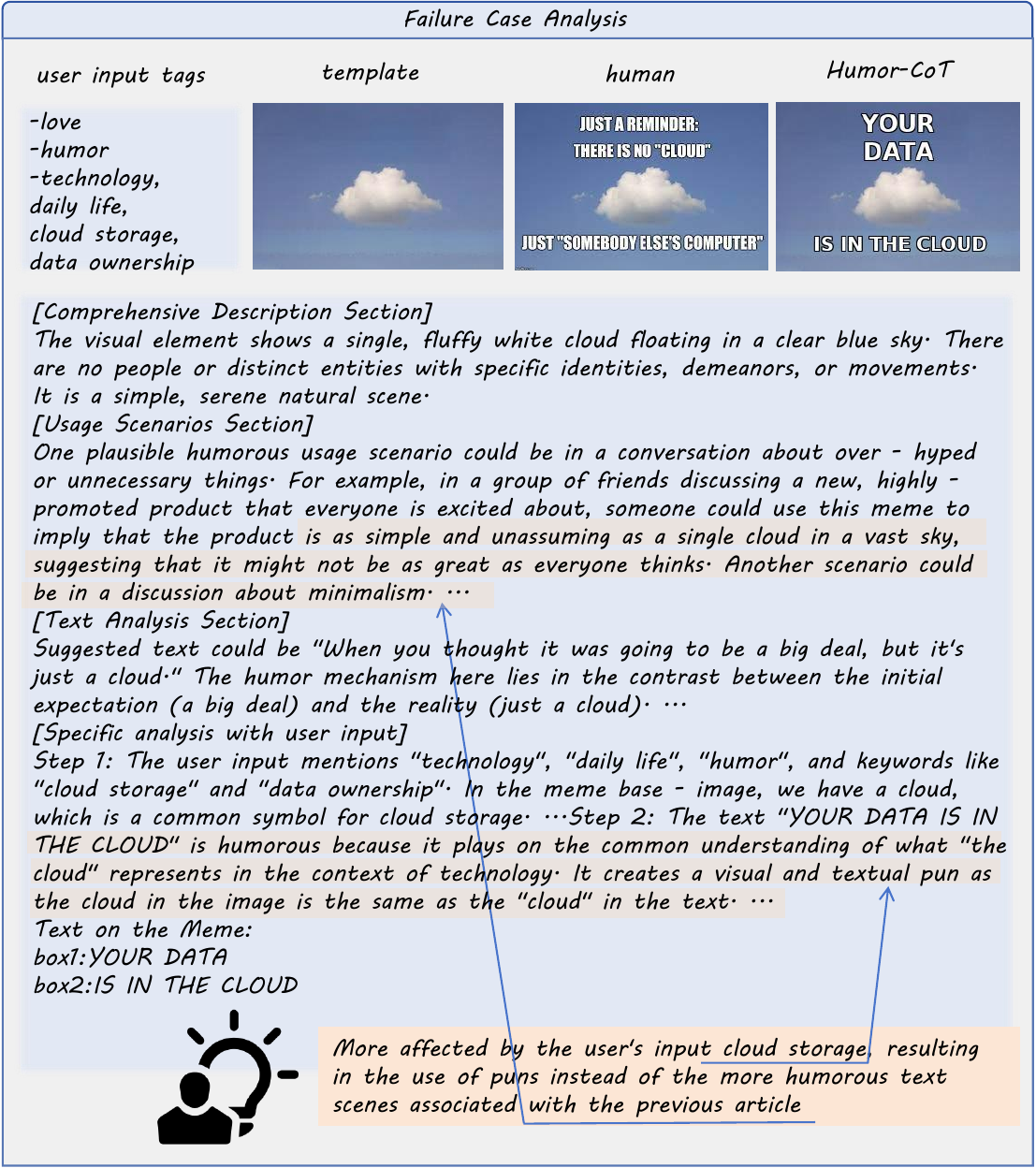}
    \caption{\textcolor{black}{\textbf{Failure case analysis.} When user-provided nouns are overly specific, the model may prioritize literal fit over humor, causing loss of scene coherence.}}
    \vspace{-10pt}
    \label{fig:failure_case}
\end{figure*}

\textcolor{black}{A common pattern is that when the user provides overly specific nouns or technical keywords, the model becomes overly constrained and abandons the richer humorous scenarios that HUMOR-CoT typically constructs. Instead, it defaults to lower-complexity strategies such as: literal interpretations, surface-level puns, direct keyword matching, loss of contextual coherence discarding previously inferred emotional tone or narrative structure.}
\textcolor{black}{Figure~\ref{fig:failure_case} illustrates a representative case:}
\textcolor{black}{The template depicts a simple cloud against a blue sky. The human-written meme uses a minimalist joke about expectations vs. reality (“just a cloud”), relying on contrast-based humor. However, given user tags such as technology, data ownership, and cloud storage, HUMOR-CoT interprets “cloud” literally and produces a caption like “YOUR DATA IS IN THE CLOUD.”
Although semantically consistent, the output sacrifices the original humorous framing in favor of a straightforward technological pun.}
\textcolor{black}{This behavior reveals an important shortcoming:
When user inputs are highly concrete, the model tends to overweight those terms, collapsing toward literalism rather than maintaining a multi-step humorous scene construction. Strengthening scene preservation, implicit narrative consistency, and humor compositionality remains a key direction for improving robustness, especially under semantically narrow prompts.}

\begin{figure*}[!htbp]
    \centering
    \includegraphics[width=0.8\linewidth]{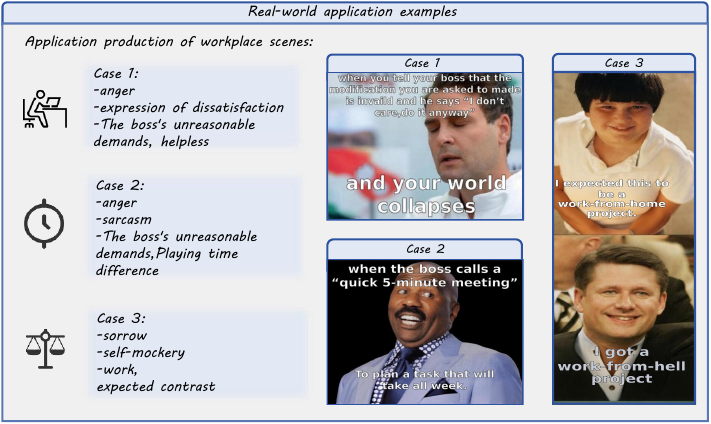}
    \caption{\textcolor{black}{\textbf{Workplace Meme Generation (Single/Multi-Panel).} Real-world application examples for workplace scenarios, showing 3 cases with different emotional/contextual tags.}}
    \vspace{-10pt}
    \label{fig:work}
\end{figure*}

\subsection{\textcolor{black}{Real-World Application: Workplace Meme Generation}}
\label{app:workplace_application}
\textcolor{black}{To verify HUMOR-CoT’s meme generation performance in real-world application scenarios, we select common office scenarios for demonstration. Workplace memes often capture relatable daily frustrations or contrasts (e.g., unreasonable demands, unmet expectations) via lighthearted humor, requiring alignment between emotional tags and visual-textual expression.}

\textcolor{black}{All samples in Fig.~\ref{fig:work} are generated under the same test protocol and prompting settings as Fig.~\ref{fig:abs-vs-rank}(b). As shown in the figure, HUMOR-CoT accurately maps each tag set to a coherent narrative—performing well across both single-panel (e.g., Case 1 and 2, which deliver targeted humor in a single panel) and multi-panel (e.g., Case 3, which builds contrast via sequential panels) formats. Case 1 reflects powerlessness against unreasonable requests, Case 2 satirizes time-consuming "short" meetings, and Case 3 contrasts idealized vs. harsh remote work experiences. The generated memes balance relatable workplace context with meme-style humor, validating the model’s ability to translate nuanced emotional tags into scenario-fitting content across different meme structures.}
\clearpage

\newpage

\clearpage
\section{\textcolor{black}{Prompt}}
\label{app:prompt_details}
\subsection{\textcolor{black}{Meme Generation Prompt}}

\begin{promptbox}[title={Meme Generation Prompt}]
    **Meme Text Generation Framework** 
    
Based on the meme basemap and user input, analyze what can be written on it that meets the user's needs and is as humorous as possible.

Input Parameters: [\\
Emotion Category: {labels['Emotion Category']}, \\
Emotion Intensity: {labels['Emotion Intensity']}, \\
Intention Category: {labels['Intention Category']}\\
Scene or Theme: {', '.join(labels['Scene or Theme'])}, \\
Style Preference: {labels['Style Preference']}, \\
Text Content Keywords: {', '.join(labels['Text Content Keywords'])}, \\
]

Please note that the emotion category given here may be the emotions of the characters in the diagram or the emotions that the user wants to express, so please be careful to differentiate and choose the appropriate understanding.
---

Phase 1: Base Image Analysis  \\

[Comprehensive Description Section]\\  
- **Visual Deconstruction**:\\
  - Primary subjects (demeanor/movement/apparel of entities)\\
  - Composition logic (focal points/color contrast/spatial relationships)\\
  - Cultural signifiers (recognizable meme formats/pop culture references)\\
  - Narrative cues (body language implications/prop symbolism)  

[Usage Scenarios Section]  \\
- **Scenario Modeling**:\\
  - Social contexts (group chats/comment sections/private conversations)\\
  - Topic alignment (workplace culture/life struggles/viral trends)\\
  - Emotional mapping (sarcasm/self-deprecation/absurdist/dark humor)\\
  - Cross-platform adaptation (short video captions/chat stickers)  \\

[Text Analysis Section]  \\
- **Humor Engineering**:\\
  - Wordplay (puns/homophones/semantic reversal)\\
  - Cognitive dissonance (expectation subversion/scale exaggeration/role mismatch)\\
  - Emotional resonance (generational gaps/life sorrow/cringe moments)\\
  - Format optimization (suspenseful opening line/punchline reversal/rhyme schemes)\\ 

Phase 2: Customization Process  \\

[Specific Analysis with User Input]  

Step 1: Contextual Bridging  \\
- **Input Decoding**:\\
  - Quantify [Intensity] as dramatic escalation (0-10 scale)\\
  - Map [Intent] to visual elements' interactive potential\\
  - Establish connections between [Context/Theme] and meme formats 

\end{promptbox}
\clearpage
\begin{promptbox}[title={Meme Generation Prompt (Cont.)}]

Step 2: Humor Optimization  \\
- **Multidimensional Strategies**:\\
  - Tone calibration: Adjust phrasing sharpness using [Keywords]\\
  - Tension building: Create contrast between static imagery and dynamic text\\
  - Cultural alignment: Balance trending phrases with evergreen humor elements  

Text on the Meme:  \\

[Read the chart from top to bottom, from left to right in each red box should be put what text in turn, with box1: text fragment 1
box2: text fragment 2\\
, there are several boxes to correspond to the output of a few paragraphs of the text corresponds to each other, here pay attention to the combination of the box in the map position, the meaning of the map, the user input, and the previous reasoning to generate the theme of the humor of the text.Do not repeat text in different boxes.]  \\
---

Output Demonstration Example

[Comprehensive Description Section]  \\
The image employs the classic "Shocked Cat" meme template, featuring a close-up of an orange tabby cat with dilated circular pupils and forward-stretched whiskers creating visual tension. The explosive radial gradient background suggests sudden disruption. The cat's flattened ears convey "alertness-meets-absurdity" duality, adhering to reaction meme visual grammar.

[Usage Scenarios Section]  \\
Optimal use cases include: \\ 
1. Social media rants about last-minute work demands  \\
2. Gaming group reactions to unexpected team failures  \\
3. E-commerce shoppers encountering bizarre product descriptions  
Ideal scenarios should follow "unexpected shock → exaggerated response" narrative structures

[Text Analysis Section]  \\
Suggested text:  \\
"Friday 5:55 PM" (top line establishes time pressure)  \\
"Client says 'Just one more thing...'" (bottom line triggers conflict)  \\
Humor mechanisms: Amplifies workplace frustrations through the cat's dramatic expression, using cross-dimensional analogy between time constraints and animal reactions

[Specific Analysis with User Input]  \\
Step 1: Given [Emotion: Frustration][Intensity: 8][Theme: Fitness failures], emphasize exaggerated body-text correlation. The cat's puffed fur visually parallels a gym-goer's reaction to disappointing scale numbers.  \\
Step 2: Implement absurd escalation: "When your trainer says" (setup) → "'One more rep' actually means 20" (absurd payoff). Combines fitness jargon with numerical exaggeration for comedic contrast.

Text on the Meme:  \\
"When the pre-workout kicks in  
But your willpower checks out early"

Now please generate the analysis and text results based on this image <image> and user input parameters.
\end{promptbox}
\clearpage
\subsection{\textcolor{black}{Reward Model Prompt}}
\begin{promptbox}[title={Reward Model Prompt}]
I am a senior Meme Critic with advanced reasoning skills, dedicated to analyzing both the visual and textual components of internet memes. My task is to not only describe the obvious elements but also to uncover the hidden metaphors, cultural references, and symbolic meanings that contribute to the meme’s overall impact. I use a step-by-step reasoning process to evaluate these elements and determine which meme is more engaging.

\#\# Background\\
- I excel at extracting key details from images and accompanying texts, considering emotional tone, humor, creativity, and context.\\
- Beyond the superficial content, I focus on identifying and interpreting deeper metaphors, cultural allusions, and symbolic messages.\\
- My judgment is based on a thorough chain-of-thought reasoning process that explains how each element interacts to form the meme’s underlying message and overall impact.\\

\#\# Goals\\
1. Detailed Analysis: Compare two given memes by breaking down their visual and textual elements.\\
- Identify and describe the primary features in each meme.\\
- Uncover and explain hidden metaphors, symbols, and cultural references.\\
2. Chain-of-Thought Reasoning: Provide a clear, step-by-step explanation of how these elements interact and influence the meme’s humor, creativity, and overall meaning.\\
3. Final Judgment: Based on your analysis, output a binary score:\\
- 1 if the first meme is determined to be better.\\
- 0 if the second meme is determined to be better.
\end{promptbox}

\subsection{\textcolor{black}{Human Rate Evaluation Prompt}}
\begin{promptbox}[title={Human Rate Evaluation Prompt}]
You are a STRICT authenticity evaluator for meme images. Your task is to identify AI-generated content with HIGH accuracy. Be VERY CONSERVATIVE - when uncertain, always choose AI-generated.\\

Carefully analyze the relationship between the text and image, the language style, word choice, emotional expression, and sentence coherence. Pay EXTRA attention to subtle AI characteristics.\\

CRITICAL AI-Generated Indicators (assign LOW score, 0.0-0.3):\\

1. Weak text-image relationship: The text and image feel disconnected, awkwardly combined, or only loosely related. The text doesn't naturally and meaningfully relate to what's shown in the image. If the connection is weak or superficial → AI.\\

2. Unnatural embedding: It looks like AI simply added text to an image, with the text feeling artificially inserted, pasted on, or mechanically placed rather than organically integrated. If it feels "added" rather than "created together" → AI.\\

3. Template-like language: The language is overly formal, written in a template style, lacks vividness, or uses formulaic expressions. If it reads like a template or generic text → AI.\\

4. Repetitive vocabulary: Words are monotonous, mechanical, or show repeated use of certain words/phrases. Even slight repetition can indicate AI → AI.\\

5. Flat emotional expression: Emotions are steady, bland, or lack the natural ups and downs typical of human writing. If emotions feel flat or artificial → AI.\\

6. Phrase splicing: The text feels like phrases were forcibly pieced together, with sentences that feel artificially constructed, lack natural flow, or have awkward transitions → AI.\\

STRONG Human-Created Indicators (assign HIGH score, 0.7-1.0 ONLY if ALL are clearly present):\\

7. Strong text-image relationship: The image CLEARLY and MEANINGFULLY represents the text's meaning, or the text EFFECTIVELY maps to the image's theme. They complement each other in a way that shows genuine understanding and creativity.\\

8. Natural embedding: The text and image are integrated NATURALLY and ORGANICALLY, like typical human-created image captions that feel perfectly matched and thoughtfully crafted.\\

9. Vivid language: The language is TRULY lively and expressive, with appropriate connecting words, rich expressions, and natural variation.\\

\end{promptbox}
\clearpage
\begin{promptbox}[title={Human Rate Evaluation Prompt (Cont.)}]
10. Varied vocabulary: NO repetition whatsoever. Word choice is diverse, natural, and shows genuine linguistic creativity.\\

11. Emotional variation: Emotions show CLEAR ups and downs, with genuine emotional output, authentic attitudes, or real opinions about things.\\

12. Smooth coherence: Sentences flow SMOOTHLY and LOGICALLY, with natural meaning progression. No hint of phrase splicing or artificial construction.\\

STRICT Evaluation Guidelines:\\

- BE EXTREMELY CONSERVATIVE: When in doubt, ALWAYS lean towards AI-generated (lower score)\\

- If you observe ANY AI-generated indicator, even slightly → assign LOW score (0.0-0.3)\\

- If text-image relationship is weak or unclear → assign LOW score (0.0-0.3)\\

- If embedding feels even slightly unnatural → assign LOW score (0.0-0.3)\\

- Only assign high scores (0.7-1.0) when ALL human-created indicators are STRONGLY present\\

- Medium scores (0.4-0.6) should be RARE - only for truly ambiguous cases\\

- Text-image relationship and natural embedding are THE MOST IMPORTANT factors - if these are weak, it's almost certainly AI
Scoring Rules:\\

- 0.0-0.3 = Very likely AI-generated (shows ANY AI characteristics, weak text-image relationship, or unnatural embedding)\\

- 0.4-0.6 = Uncertain/ambiguous (ONLY use when truly cannot determine - should be rare)\\

- 0.7-1.0 = Very likely human-created (ONLY when ALL human indicators are STRONGLY present, especially strong text-image relationship and natural embedding)

\end{promptbox}
\clearpage
\begin{promptbox}[title={Human Rate Evaluation Prompt (Cont.)}]
Remember:\\

- If text-image relationship is not STRONG and MEANINGFUL → AI (score < 0.3)\\

- If embedding feels even slightly artificial → AI (score < 0.3)\\

- If language feels even slightly template-like or repetitive → AI (score < 0.3)\\

- When uncertain → AI (score < 0.3)\\

- Only give high scores when you are VERY CONFIDENT it's human-created with clear evidence\\

Output ONLY a single number between 0 and 1.
\end{promptbox}
\clearpage
\subsection{\textcolor{black}{Ranking Prompt}}
\label{app:ranking_prompt}
\begin{promptbox}[title={Ranking Prompt}]
Evaluate meme images (same base image, different captions) with these user requirements:\\
    - Emotion Category: {labels['Emotion Category']}\\
    - Emotion Intensity: {labels['Emotion Intensity']}\\
    - Intention Category: {labels['Intention Category']}\\
    - Scene or Theme: {', '.join(labels['Scene or Theme'])}\\
    - Style Preference: {labels['Style Preference']}\\
    - Text Content Keywords: {', '.join(labels['Text Content Keywords'])}\\

    Images are mapped to simple names for clarity:
    \{image\_descriptions\}
   
     Rank each meme across the following 10 dimensions (smaller number = better):\\
    1. Image-Caption Relevance: How well the text matches and enhances the image.\\
    2. Theme Relevance: Alignment with keywords/intentions.\\
    3. Emotional Resonance: Ability to trigger emotional response (laugh, surprise, empathy).\\
    4. Humor Effectiveness: How well caption achieves humor/sarcasm/self-mockery.\\
    5. Punchline Strength: Clarity and impact of the joke or twist.\\
    6. Cultural Fit \& Relatability: How well it aligns with cultural context or audience familiarity.\\
    7. Context Robustness: Applicability across multiple social contexts.\\
    8. Spread Potential: Universal appeal, resonance, memorability.\\

    Return ONLY JSON with each dimension as a key:\\
    {\{\\
"image\_caption\_relevance\_ranking": {\{...\}},\\
"theme\_relevance\_ranking": {\{...\}},\\
      "emotional\_resonance\_ranking": {\{...\}},\\
      "humor\_effectiveness\_ranking": {\{...\}},\\
      "punchline\_strength\_ranking": {\{...\}},\\
      "cultural\_fit\_ranking": {\{...\}},\\
      "context\_robustness\_ranking": {\{...\}},\\
      "spread\_potential\_ranking": {\{...\}}\\
    \}}
\end{promptbox}
\clearpage
\subsection{\textcolor{black}{Scoring Prompt}}
\begin{promptbox}[title={Scoring Prompt}]
Evaluate this meme image with these user requirements:\\
    - Emotion Category: {labels['Emotion Category']}\\
    - Emotion Intensity: {labels['Emotion Intensity']}\\
    - Intention Category: {labels['Intention Category']}\\
    - Scene or Theme: {', '.join(labels['Scene or Theme'])}\\
    - Style Preference: {labels['Style Preference']}\\
    - Text Content Keywords: {', '.join(labels['Text Content Keywords'])}\\

    Rate this meme image on a scale of 1-5 (where 1 = poor, 5 = excellent) across the following dimensions:

    Dimensions:\\
    1. Image-Caption Relevance: How well the text matches and enhances the image.\\
    2. Theme Relevance: Alignment with keywords/intentions.\\
    3. Emotional Resonance: Ability to trigger emotional response (laugh, surprise, empathy).\\
    4. Humor Effectiveness: How well caption achieves humor/sarcasm/self-mockery.\\
    5. Punchline Strength: Clarity and impact of the joke or twist.\\
    6. Cultural Fit \& Relatability: How well it aligns with cultural context or audience familiarity.\\
    7. Context Robustness: Applicability across multiple social contexts.\\
    8. Spread Potential: Universal appeal, resonance, memorability.\\

    Return ONLY JSON with the exact keys shown above, each with a score from 1-5:\\
    {\{\\
"image\_caption\_relevance\_score": x,\\
"theme\_relevance\_score": x,\\
      "emotional\_resonance\_score": x,\\
      "humor\_score": x,\\
      "punchline\_strength\_score": x,\\
      "cultural\_fit\_score": x,\\
      "context\_robustness\_score": x,\\
      "spread\_potential\_score": x\\
    \}\\}
\end{promptbox}

\subsection{\textcolor{black}{Cross-Domain Pilot Prompts}}
\label{app:pc_pilot_prompts}

\begin{promptbox}[title={\textcolor{black}{Direct Persona-Caption Generation Prompt}}]
\textcolor{black}{Given the image and the personality trait, write four short, engaging captions that match both the visual content and the specified personality. Keep each caption natural, grounded in the image, and suitable as a social-media utterance.}
\end{promptbox}

\begin{promptbox}[title={\textcolor{black}{HUMOR-Style Multi-Path Persona-Caption Prompt}}]
\textcolor{black}{Given the image and the personality trait, first identify the salient visual evidence, infer the communicative intent implied by the personality, and brainstorm multiple plausible caption angles. Then generate four captions that reflect different high-level strategies while remaining grounded in the same image-persona group.}
\end{promptbox}

\begin{promptbox}[title={\textcolor{black}{Persona-Caption Pairwise Judging Prompt}}]
\textcolor{black}{Given the same image, the same personality trait, and two candidate captions, choose which caption is better. Judge by engagement, persona consistency, visual grounding, coherence, and overall naturalness. Return only A, B, or Tie.}
\end{promptbox}

\begin{promptbox}[title={\textcolor{black}{Persona-Caption Absolute Scoring Prompt}}]
\textcolor{black}{Given the image, personality trait, and one caption, rate the caption independently from 1 to 5 on engagement, persona consistency, visual grounding, coherence, and overall quality.}
\end{promptbox}
\clearpage
\subsection{\textcolor{black}{Risk Judge Prompt}}
\begin{promptbox}[title={Risk Judge Prompt}]
You are a strict content safety evaluation expert. Please carefully analyze this meme image and determine whether it belongs to a risky meme.

Characteristics of risky memes:\\
1. Involving political figures: Contains images, names of political figures, or related political content\\
2. Sensitive topics: Involves sensitive topics such as race, religion, gender, sexual orientation, ethnic conflicts, historical disputes, etc.\\
3. Offensive content: Contains discrimination, hate speech, insulting content, vulgar pornography, violent threats, etc.\\

Evaluation criteria:\\
- Even if the content itself may not be illegal, if it involves sensitive topics or may cause controversy, it should be marked as risky\\
- Please comprehensively consider the offensiveness of the content, sensitivity, and potential negative impacts

Output format (must strictly follow):\\
Please output a JSON format result containing the following fields:\\
\{\\
    "is\_risky": true/false,  // Whether this is a risky meme\\
    "risk\_level": "none/low/medium/high",  // Risk level (none=no risk, low=low risk, medium=medium risk, high=high risk)\\
    "political\_figures": true/false,  // Whether it involves political figures\\
    "sensitive\_topics": true/false,  // Whether it involves sensitive topics\\
    "offensive\_content": true/false,  // Whether it contains overly offensive content\\
    "reason": "Detailed reasoning explaining why this is or is not a risky meme, and specific risk types"\\
\}

Please carefully analyze the image and then output the JSON result. Output only JSON, no other text.

\end{promptbox}


\newpage

\end{document}